\documentclass[pdflatex,sn-mathphys]{sn-jnl}

\usepackage{lineno}
\PassOptionsToPackage{colorlinks}{hyperref}
\PassOptionsToPackage{hyphens}{url}
\pdfoutput=1
\usepackage{lmodern}
\usepackage{amsmath,amssymb}
\usepackage{iftex}
\usepackage{caption}
\usepackage{siunitx}
\ifPDFTeX
  \usepackage[T1]{fontenc}
  \usepackage[utf8]{inputenc}
  \usepackage{textcomp} 
\else 
  \usepackage{unicode-math}
  \defaultfontfeatures{Scale=MatchLowercase}
  \defaultfontfeatures[\rmfamily]{Ligatures=TeX,Scale=1}
\fi
\usepackage{numprint}
\usepackage{parskip}
\usepackage{array}
\newcolumntype{C}{>{\centering\arraybackslash} m{2.6cm} }  

\usepackage{xcolor}
\definecolor{BBBlue}{HTML}{26196C}
\usepackage{xurl} 
\AtBeginDocument{%
  \hypersetup{
    citecolor=BBBlue,
    linkcolor=BBBlue,   
    urlcolor=BBBlue}}
\usepackage{graphicx}
\makeatletter
\def\maxwidth{\ifdim\Gin@nat@width>\linewidth\linewidth\else\Gin@nat@width\fi}
\def\maxheight{\ifdim\Gin@nat@height>\textheight\textheight\else\Gin@nat@height\fi}
\makeatother
\setkeys{Gin}{width=\maxwidth,height=\maxheight,keepaspectratio}
\makeatletter
\def\fps@figure{htbp}
\makeatother
\makeatletter
\AtBeginDocument{%
\renewcommand*\figurename{Figure}
\renewcommand*\tablename{Table}
}
\AtBeginDocument{%
\renewcommand*\listfigurename{List of Figures}
\renewcommand*\listtablename{List of Tables}
}
\newcounter{pandoccrossref@subfigures@footnote@counter}
\newenvironment{pandoccrossrefsubfigures}{%
\setcounter{pandoccrossref@subfigures@footnote@counter}{0}
\begin{figure}\centering%
\gdef\global@pandoccrossref@subfigures@footnotes{}%
\DeclareRobustCommand{\footnote}[1]{\footnotemark%
\stepcounter{pandoccrossref@subfigures@footnote@counter}%
\ifx\global@pandoccrossref@subfigures@footnotes\empty%
\gdef\global@pandoccrossref@subfigures@footnotes{{##1}}%
\else%
\g@addto@macro\global@pandoccrossref@subfigures@footnotes{, {##1}}%
\fi}}%
{\end{figure}%
\addtocounter{footnote}{-\value{pandoccrossref@subfigures@footnote@counter}}
\@for\f:=\global@pandoccrossref@subfigures@footnotes\do{\stepcounter{footnote}\footnotetext{\f}}%
\gdef\global@pandoccrossref@subfigures@footnotes{}}
\@ifpackageloaded{float}{}{\usepackage{float}}
\floatstyle{ruled}
\@ifundefined{c@chapter}{\newfloat{codelisting}{h}{lop}}{\newfloat{codelisting}{h}{lop}[chapter]}
\floatname{codelisting}{Listing}

\makeatother

\usepackage{tabularx}
\usepackage[export]{adjustbox}
\usepackage{subcaption}

\jyear{2021}%

\theoremstyle{thmstyleone}%

\theoremstyle{thmstyletwo}%

\theoremstyle{thmstylethree}%

\begin{document}

\title[Deep Learning for Estimating Image Memorability]{Embracing New Techniques in Deep Learning for Estimating Image Memorability}

\author*[1]{\fnm{Coen D.} \sur{Needell}}\email{coeneedell@gmail.com}

\author[2]{\fnm{Wilma A.} \sur{Bainbridge}}\email{wima@uchicago.edu}

\affil*[1]{\orgdiv{Department of Psychology}, \orgname{University of Chicago}, \orgaddress{\street{5848 S. University Ave}, \city{Chicago}, \postcode{60637}, \state{IL}, \country{United States of America}}}

\abstract{Various work has suggested that the memorability of an image is
consistent across people, and thus can be treated as an intrinsic
property of an image. Using computer vision models, we can make specific
predictions about what people will remember or forget. While older work
has used now-outdated deep learning architectures rooted in shallow visual processing to predict image memorability,
innovations in the field have given us new techniques to apply to this
problem. Here, we propose and evaluate five alternative deep learning
models which exploit developments in the field from the last
five years, largely the introduction of residual neural networks, which are intended to allow the model to use semantic
information in the memorability estimation process. These new models were tested against the prior state of the art with a combined dataset built to optimize both within-category and across-category predictions.
Our findings suggest that the key prior memorability network had overstated its generalizability and was overfit on its training set. Our new models outperform this prior model, leading us
to conclude that Residual Networks outperform simpler convolutional
neural networks in memorability regression. We make our new state-of-the-art model readily available to the research community, allowing
memory researchers to make predictions about memorability on a
wider range of images.}

\maketitle

\hypertarget{background}{%
\section{Introduction}\label{background}}

A person's ability to remember an image is thought of as something that varies
from person to person. Some people have ``good'' memories, while others forget
images easily. While this is certainly true, recent research has found a surprising consistency across individuals in what they remember and forget, driven by the images themselves; images can be intrinsically memorable and forgettable across observers. Because of this high consistency in memory performance, each image can be assigned a memorability score between 0 and 1
 \citep{bainbridgeMemorabilityHowWhat2019, IsolaWhatMakes2011}. These ``memorability scores''
are determined experimentally by testing how well a large group of
people remember an image. This is an expensive and time-consuming
process, especially at scale. There are also still big open questions about what determines the memorability of an image. In this paper, we propose a new method for
evaluating the memorability of images, and reveal an important contribution of semantic information in memory performance.

Memorability scores that can be assigned to an image are conventionally generated using a memory test. A
participant is given a series of images and is asked whether or not they
have seen them before. By aggregating how human participants score on
each image, researchers can generate an estimate for the ground truth
memorability of that image. Memorability scores generated from a subset
of the participants are a good estimator of the memorability scores from
the complement of that subset (measured by a high Spearman rank
correlation between them) \citep{Bainbridge_Isola_Oliva_2013, IsolaParikhTorralbaOliva2011}.
The question then is how do these memorability scores arise, and why are
they so consistent across people?

As soon as one views an image, high level visual and memory areas in the brain show a sensitivity to the memorability of that image, regardless of task or whether participants consciously remember that image or not  \citep{bainbridgeDissociatingNeuralMarkers2018, bainbridgeMemorabilityStimulusdrivenPerceptual2017, Mohsenzadeh2019}. This sensitivity may even occur in the brains of non-human primates, suggesting that memorability is evoked by images themselves, during late perceptual processes \citep{jaeglePopulationResponseMagnitude2019}. A current hypothesis is that memorability could reflect how perceptual inputs are prioritized for memory, given that not all perceived information can be remembered. Specifically, memorable items may be those that we visit first during retrieval in the larger search space of memories, and memorable items are reinstated earlier in the brain than forgettable ones \citep{xieMemorabilityWordsArbitrary2020}. However, the stimulus features that cause an image to be memorable are still under active exploration. Memorability shows low correspondences to low-level visual features like colors and contrasts, and is also not predictable by comprehensive sets of human-labeled features like attractiveness or interest. Computational models promise to be the next step in understanding what causes a given image to be remembered \citep{isolaWhatMakesPhotograph2014}.

Given that memorability is intrinsic to an image, then this implies that we should be
able to derive memorability directly from an image. The earliest
work in applying computer vision to memorability used hand-engineered
Histograms of Oriented Gradients (HOG), and dense Scale Invariant
Feature Transforms
(dense-SIFT) \citep{khoslaModifyingMemorabilityFace2013}. While these
methods can be very effective, they are hand-engineered, meaning that they need to be
tuned by a human, instead of being determined algorithmically, and thus require assumptions about the relevant visual features. As end-to-end trained vision techniques like convolutional
neural networks (CNNs) became available, the problems associated with HOGs and
dense-SIFT models could be avoided, giving us better performing estimators
at the cost of interpretability.

In recent years, other methods for estimating memorability have been proposed using
convolutional neural network regressions. The convolutional neural
network is a deep learning technique that scans over an image in small
regions, and optimizes the process of analyzing the regions, rather than
optimizing over every data dimension (every pixel and channel). This
generally yields good performance while lowering training costs and
preventing vanishing gradients. The most well-known of these is
MemNet \citep{khoslaUnderstandingPredictingImage2015}, although it does
have a few challengers which have employed attention
models \citep{fajtlAMNetMemorabilityEstimation2018}, image captioning
features \citep{squalli-houssainiDeepLearningPredicting2018}, and transfer learning \citep{basavarajuObjectMemorabilityPrediction2019}. MemNet's
memorability estimations had a rank correlation of 0.57 with held-out
ground truth scores from its training dataset and is the most commonly
used neural network regression for this purpose. However, some roadblocks have 
emerged for researchers trying to use MemNet. MemNet was built in Caffe, a deep 
learning framework 
which has been defunct since shortly after MemNet's publication, and its online demo is no longer functional. Importantly, MemNet uses techniques which are no longer preferred for
computer vision regression. Techniques like residual neural networks,
semantic segmentation, and new optimization methods have
been developed and become widespread in the meantime, and have improved the ability of these networks to derive richer, semantic information from an image.

Here, we present a new state-of-the-art model for predicting
image memorability based on recent advances in computer vision and our
understanding of memorability itself. Using these new techniques, as
well as an improved combination dataset, this new model can be more
generalizable to the image-space. Further, these new advances reveal an important role of semantic information in image memorability, and have spurred the creation of a new dataset for examining the role of different features in image memorability.

\section{Methods}

First, we will reproduce MemNet on a modern framework, and explore its
statistical properties in a controlled environment. Then, we will outline
deep learning methods that utilize  new techniques for the same
problem, and show that their use results in large performance
increases. The techniques we will discuss are based around residual
neural networks (ResNets), a specialization of convolutional neural
networks. Next, we will explore these new models, and show that the
addition of a residual neural network gives the model access to higher
level conceptual features. Finally, we will show that our model makes successful predictions of memory across a diverse range of image sets, making it a useful tool for the research community.

\hypertarget{data-sources}{%
\subsection{Data Sources}\label{data-sources}}

The original MemNet was trained on a dataset called LaMem
 \citep{khoslaUnderstandingPredictingImage2015}, the Large Scale
Memorability Dataset. LaMem contains 58,741 images, all labeled with the
average memorability for each image. If one splits all of the human
participants into two groups, and takes the Spearman rank correlation
between each group's average memory performance for a given image, the correlation will be 0.67,
suggesting that people are indeed highly consistent in the
images they remember and forget. The images in LaMem are a compilation
of a few more well-known datasets, the MIR Flickr Image Set
 \citep{huiskesMIRFlickrRetrieval2008}, the AVA (Aesthetic Visual
Analysis image set)  \citep{murrayAVALargescaleDatabase2012}, the
Affective Image Set  \citep{machajdikAffectiveImageClassification2010}, 
two saliency image sets \citep{juddLearningPredictWhere2009},
 \citep{ramanathanEyeFixationDatabase2010}, the Scene Understanding (SUN) dataset
 \citep{xiaoSUNDatabaseLargescale2010}, an image popularity dataset
constructed using Flickr  \citep{khoslaWhatMakesImage2014}, an anomaly
detection dataset  \citep{salehObjectCentricAnomalyDetection2013}, and a
dataset designed for attribute-based object description
 \citep{farhadiDescribingObjectsTheir2009}. These datasets were chosen because they have other metadata associated with them, like
popularity on Flickr, or eye tracking data, or aesthetic ratings. All
together, LaMem has the benefit of being relatively naturalistic
and diverse. The dataset was compiled with the intention of looking for
correlations between memorability and these other data points. LaMem is a
well suited dataset for this task because of its size and relative
quality.

Our reimplementation of MemNet, as well as all of our models, were trained on a
mixture of LaMem and MemCat
 \citep{goetschalckxMemCatNewCategorybased2019}. MemCat is a smaller, but
higher quality dataset of images and memorability scores.
MemCat was designed to help understand how memorability changes across and within categories of images. It has 10,000
images, and its inter-human rank correlation is 0.78. This value
generally increases with the amount of samples the dataset has per
image, and it scales with the certainty of each value. Taken together,
this implies that the memorability scores in MemCat are better
estimations of human performance. 

\begin{figure}
    \centering
    Example LaMem Images
    \begin{tabular}{C C C C}
       \includegraphics[width=\textwidth,height=2cm]{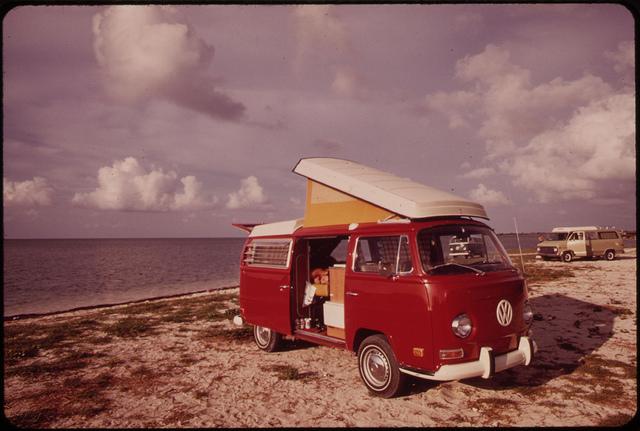} & \includegraphics[width=\textwidth,height=2cm]{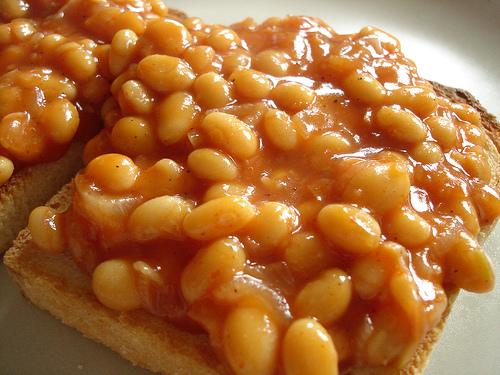} & \includegraphics[width=\textwidth,height=2cm]{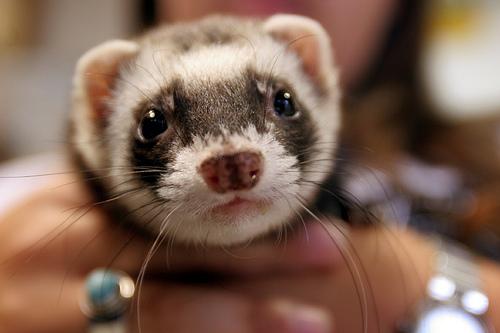} & \includegraphics[width=\textwidth,height=2cm]{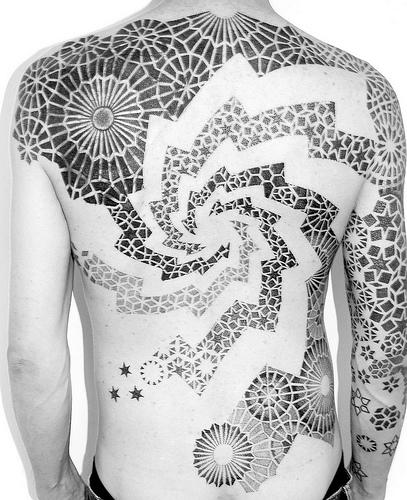} \\
       & & & \\
    \end{tabular}
    Example MemCat Images
    \begin{tabular}{C C C C}
       \includegraphics[width=\textwidth,height=2cm]{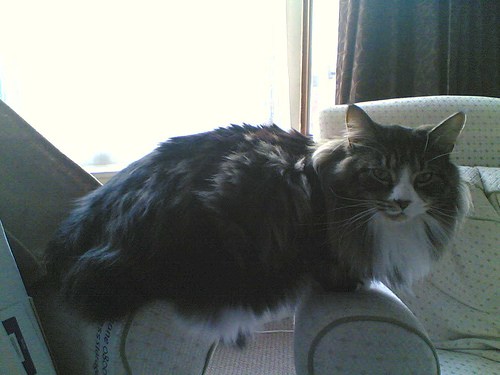} \linebreak cat & \includegraphics[width=\textwidth,height=2cm]{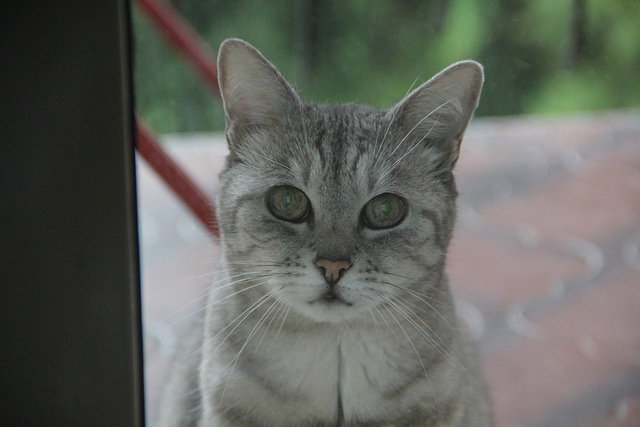} \linebreak cat & \includegraphics[width=\textwidth,height=2cm]{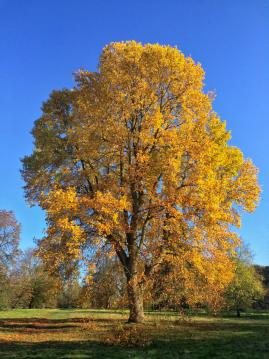} \linebreak forest (broadleaf) & \includegraphics[width=\textwidth,height=2cm]{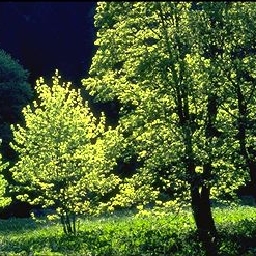} \linebreak forest (broadleaf) \\
    \end{tabular}
    \caption{Sample images from our combined dataset. The first row is from LaMem \citep{khoslaUnderstandingPredictingImage2015}, and the second row is from MemCat \citep{goetschalckxMemCatNewCategorybased2019}. The specific category is written underneath each MemCat image.}
    \label{fig:samp}
\end{figure}

These two datasets have some key features that can influence the
behavior of our deep learning models. LaMem was compiled from a group of
other datasets, which in turn were compiled to study secondary
metadata of that image, like popularity on Flickr, aesthetic rating,
etc. Considering the data sources for LaMem, we can notice that they
are not a general sampling of naturalistic images. Flickr, for
example, is an online community focused around sharing interesting
images, thus, datasets from Flickr have a selection bias toward less
naturalistic images. Among
the other data sources, we have images that are designed to trigger eye
movement, images that are designed to be anomalous, images that are
designed to be affective, and so on. For example, many of the images from Flickr
 are intended as artistic renderings, and not
necessarily true reflections of reality. On the other hand, LaMem is
very generalizable across image types. LaMem contains images of scenes,
objects, people, and even a few images of text and
iconography (fig. \ref{fig:samp}). The vast majority of LaMem's images are scenes, but the
fact that it contains images of other types will improve the
generalizability of the model across those types.

MemCat, on the other hand, was designed to describe several categories
of object-focused images. Each image fits neatly into a category,
like "cat" or "snowscape" and as such, every
image is an object image, where the objects are largely unoccluded, and
generally presented in context, and serve as the focus of each
image (fig. \ref{fig:samp}). This
contrasts with LaMem whose images are in a less natural context, and
less focused on presenting a single object per image. However, what
MemCat sacrifices in type-generalizability, it earns back in
within-object generalizability; there are roughly a hundred images of
deciduous forests and a hundred more of coniferous forests.

The final dataset used in the current study is a combination of these two datasets,  with a mixture of objects, scenes, and miscellaneous
images. The trends and features found in the larger but lower certainty
dataset (LaMem) get reinforced by the smaller, more precise dataset (MemCat), resulting in a final model that is powerful and generalizable on
multiple axes.

\hypertarget{models}{%
\subsection{Architectures}\label{models}}

We will compare three deep learning architectures, two of which are novel, that serve as the
backbone for five models. First we will examine the MemNet architecture as downloaded in its original CaffeModel form. We also re-implemented its architecture using the more modern PyTorch \citep{teamTorchTensorsDynamic} framework.
Second, we will describe our new ResMem architecture, an architecture that
uses residual neural networks as a secondary feature, and used here to 
create two models, called ResMem and ResMemRetrain. The third architecture we implement is M3M, which
introduces a tertiary feature, semantic segmentation.
Our new architectures also employ relatively simplified preprocessing steps, which avoid the assumption present in prior models that memorability is preserved across cropped and normalized versions of images.

\hypertarget{memnet}{%
\subsubsection{MemNet}\label{memnet}}

We tested two implementations of MemNet, one using its original Caffe model, and one re-implementing the architecture of MemNet using more modern approaches and best practices in deep learning. First, we tested MemNet's
architecture in its original Caffe implementation \citep{khoslaUnderstandingPredictingImage2015}. The main
innovation in MemNet was the use of convolutional neural networks (CNNs) for
feature decomposition. Convolutional neural networks were first created
in the 1980s \citep{fukushimaNeocognitronSelforganizingNeural1980}, but
did not see widespread use until the 2000s when GPUs became widespread,
and a GPU implementation was written for
CNNs \citep{chellapillaHighPerformanceConvolutional}. Then, the final
piece needed to use CNNs at the modern level was to implement
backpropagation \citep{ciresanDeepBigSimple2010}. Backpropagation speeds
up neural network training by taking the derivative of the network's
output with respect to each of the parameters of the model, and uses
that to decide which parameters to try in the next step. This innovation
was shortly followed up by
AlexNet \citep{krizhevskyImageNetClassificationDeep2012}, a model that
used these techniques for image classification.

\begin{figure}
\hypertarget{fig:mndiag}{%
\centering
\includegraphics{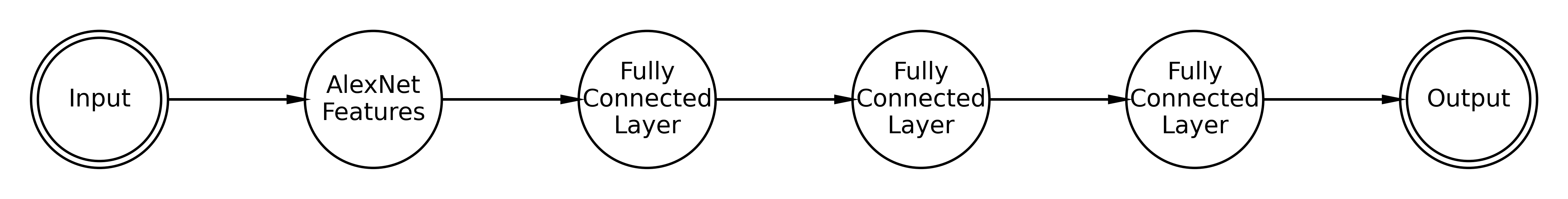}
\caption{A skeleton diagram showing MemNet's architecture. It is characterized
by a deep CNN followed by three fully connected layers.}\label{fig:mndiag}
}
\end{figure}

MemNet is designed after AlexNet (fig. \ref{fig:mndiag}), which consisted of 5
convolutional layers (some of which are pooled) followed by 3 fully
connected layers.  MemNet differs from AlexNet in the number of filters and channels in each
layer and the number of neurons in the fully connected layers,  but the overall architecture is the same.
The intention in building MemNet was to leverage AlexNet's unparalleled
efficiency (for the time) at extracting features from images.

The Caffe model of MemNet made publicly available could not be run from the online model alone  \citep{khoslaUnderstandingPredictingImage2015}.
To allow the model to function, we implemented additional steps based on discussions with prior users of the network. First, the input images were offset by a preset image, which is purported to be an image mean.
Second, the images were subjected to a Ten Crop Transformation, where the images were cropped five times to a certain size, once with the crop aligned to each corner, and once aligned to the center of the image.
The images were then mirrored, and the process was repeated to yield ten crops in total.
These ten crops have their memorability scores estimated by the neural network, and their estimations are averaged to get one score. 
These two transformations are common, and were individually considered good practices in 2015, but are two separate approaches to the same problem.
After this, the score was transformed using a standard scaling method. 
In the original MemNet model, the parameters for this scaling were hard-coded, and were not part of the training process.
This hard-coded scaling is not considered best practice, and for this reason, we also constructed a model using a modern framework, employing best practices, while using an identical architecture to MemNet.

For this second implementation of MemNet, we reverse-engineered it using PyTorch\citep{teamTorchTensorsDynamic}, a popular framework for designing neural networks in Python. We chose PyTorch because it gives researchers a lot of direct control over the architecture of their models, and we believe that it will be well-maintained for the foreseeable future. The original Caffe version of MemNet was pretrained on only the LaMem dataset, but here we trained the PyTorch implementation of MemNet on our broader mixed dataset. This allows us to directly compare it to our new model. Further information about the process of hyperparameter tuning can be seen in the model training section of the Supplemental Information.

\hypertarget{resmem}{%
\subsubsection{ResMem}\label{resmem}}

Historically, the next big advancement in using neural networks for
computer vision was residual neural networks. ResNets are
built by taking convolutional layers and connecting them across levels
with a ``residual connection.'' The design builds on constructs observed
in pyramidal cells in the cerebral cortex. In reality, this is
implemented as adding the previous layer
to the convolution at each step \citep{heDeepResidualLearning2015}.

\begin{figure}
\hypertarget{fig:resmemdiag}{%
\centering
\includegraphics[width=\textwidth,height=5cm]{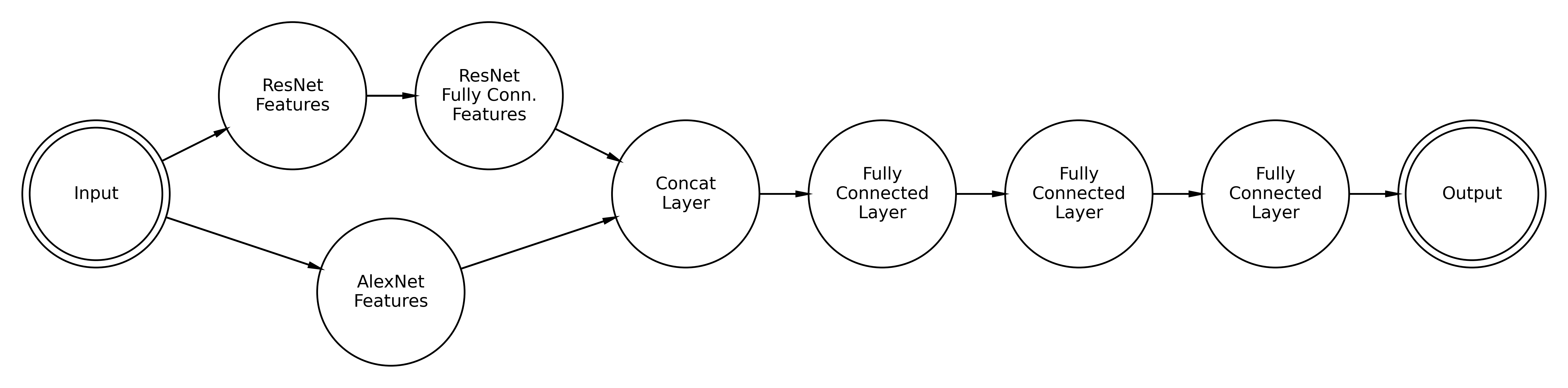}
\caption{A skeleton diagram for ResMem. The input image is run through a 
deep residual neural network, and an AlexNet feature architecture, and the
two features are combined, then run through a fully connected deep neural network.
}\label{fig:resmemdiag}
}
\end{figure}

The advent of ResNets allowed for convolutional neural networks to be
built deeper, with more layers, without the gradient vanishing. We utilized a pre-trained ResNet from the \texttt{torchvision} model
zoo \citep{teamTorchTensorsDynamic}, specifically, the whole model and not just
the convolutional features, and added it as an input feature for our
regression step. The model we used is called ``ResNet-152'' and was
originally trained for image classification using ImageNet, an industry
standard dataset for image classification \citep{deng2009imagenet}. 
ImageNet is a dataset of roughly 14 million images, and for the ResNet-152 training process they are classified into 1000 categories that are then predicted by ResNet-150.
The process of using a model that was trained on a different but similar problem is called transfer learning, and is a staple technique of deep learning.
This model consists of 152 layers of residual neural networks, followed by a short
fully connected section that produces a 1000-length feature vector.
Since the ResNet was previously trained on a semantic classification
problem, this means that the ResNet features we are using are a semantic
representation of the input image. We combine this ResNet feature with AlexNet
features to create ResMem (fig. \ref{fig:resmemdiag}).

We can use this structure to make two models. In one, the optimization algorithm does not have access to the ResNet-152 parameters during training.
This means that the intermediate ResNet features are categorical data as
the original was trained on ImageNet \citep{deng2009imagenet}, a standard image classification dataset. In the
other, which we call ResMemRetrain, we allow the ResNet component
of the model to retrain for the memorability task. This allows the
model to specialize the ResNet features for this task, but the model loses
the assumption that the ResNet features are semantically in line with
ImageNet. By allowing ResNet-152 to retrain,
the model gains specificity at the expense of a much longer training time. Further information about the hyperparameter tuning of these models can be seen in the Supplemental Information.

\hypertarget{m3m}{%
\subsubsection{M3M}\label{m3m}}

\begin{figure}
\hypertarget{fig:tripdiag}{%
\centering
\includegraphics{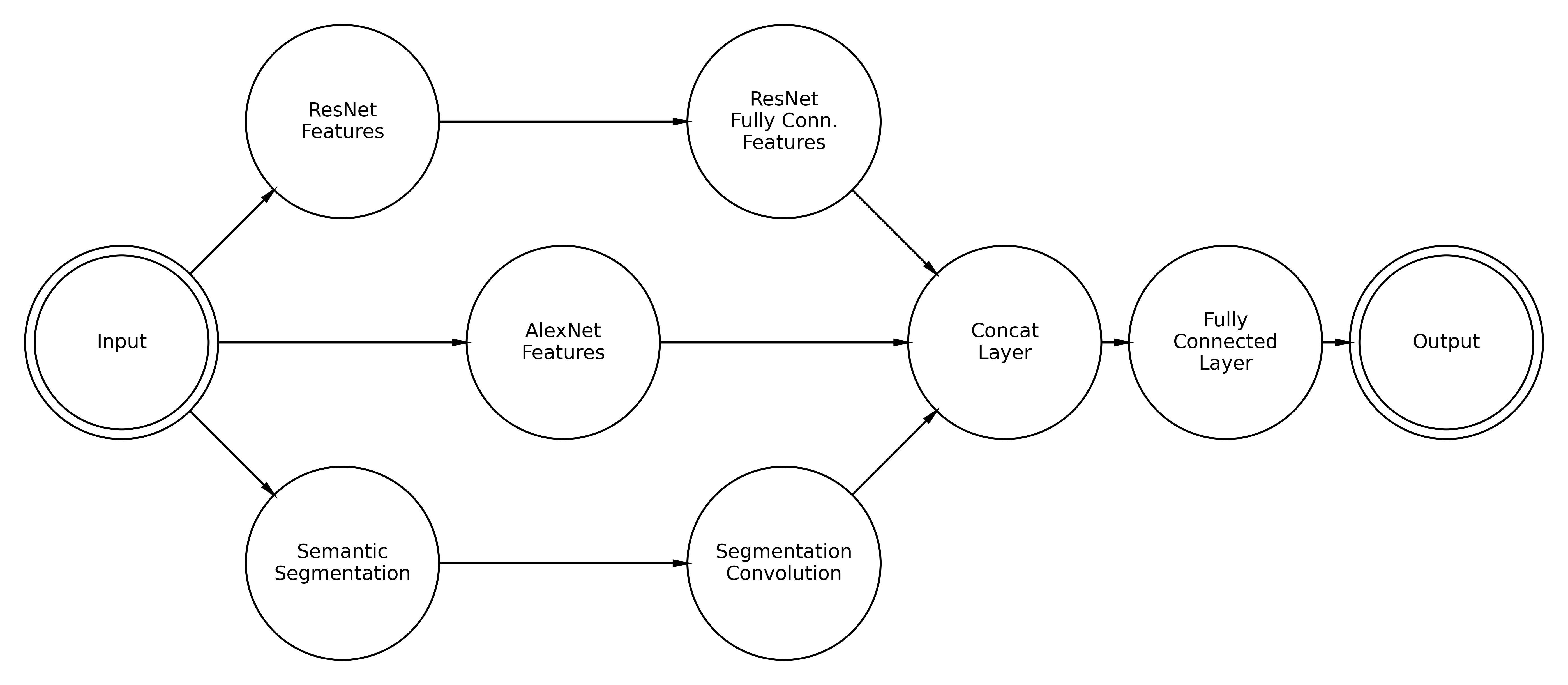}
\caption{A skeleton diagram for M3M. In addition to the ResNet and CNN, a Semantic Segmentation feature
is also sent to the fully connected deep neural network. Since semantic segmentation produces very high
dimensional outputs (on the order of one million dimensions), this feature must also be passed through
a convolutional neural network to down-scale it.
}\label{fig:tripdiag}
}
\end{figure}

Finally, we added a third feature to our model based on
Semantic Segmentation (fig. \ref{fig:tripdiag}). Semantic segmentation is a common computer vision
task where a model takes an image and classifies the object represented in each pixel.
The segmentation model we are using is called fcn\_ResNet-50, and also comes
from the PyTorch model zoo \citep{teamTorchvisionImageVideo}.
This new model was built without any retraining of the two
pretrained models, as semantic segmentation is a very high memory usage
task, and would overload standard graphics cards.

The hypothesis is that by adding another semantic feature to the model,
the overall accuracy of the model will improve. Where the classification
model ResNet-152 returns a 1000-vector with estimated probabilities for
each category, fcn\_ResNet-50 returns an array the same size as the
input image, but instead of having three channels (Red, green, and blue)
it has 21 channels, with estimated probabilities for each category for
each pixel. This is considerably more semantic information, and also
contains coupled spatial-semantic information.

Of important note, Semantic Segmentation outputs a data array
with more channels than a normal image, so it cannot be reshaped
directly into a standard linear layer. To solve this, we utilize a small
convolutional neural network to process the segmentation into more
simple convolutional features. 

\hypertarget{feature-vis}{%
\subsection{Network Feature Analysis}\label{feature-vis}}

Each layer of a convolutional neural network or residual neural
network performs feature extraction. They extract features in the
output of the previous layer, and the first layer extracts features from
the input image. The ``features'' are patterns that appear in the data,
and when optimized, the model picks out the patterns that are most
useful in solving the problem. Because these features resist precise
definition, it is difficult to tell what the model is sensitive to;
they can only be analyzed \emph{post facto}. However, the consensus is that early
layers are more perceptual and the later layers are more
conceptual  \citep{jozwikDeepConvolutionalNeural2018}. 
While fully mapping out the relationships in a deep neural network of this size is impossible, performing what feature analysis we can gives preliminary insight into the types of features important for predicting the memorability of an image.

We can visualize the important features in the model
by optimizing the inputs to maximally trigger a
certain feature in the model. By running this process on a sampling of layers and
features, we can get a sense of the important neural network features in our model. When we apply a qualitative label to each feature, we can
also see what images trigger which features, and get a better idea of how
those features contribute to memorability in combination.

\section{Results}

\begin{table}[h]
  \begin{tabular}{ l c c c}
    Model & MSE Loss $\downarrow$ & Spearman Rank Correlation $\uparrow$ & Approx. Train Time \\
    \hline
   MemNet & 0.012 & 0.55 & 2500s \\ 
   ResMem & 0.009 & 0.66 & 8900s \\  
   ResMemRetrain & 0.008 & 0.67 & 24000s \\
   M3M & 0.009 & 0.68 & 24600s 
  \end{tabular}
  \caption{A summary of results across models. Training time is for 20 epochs on an NVIDIA 1080TI running CUDA natively on Arch Linux.  MSE loss is the average square distance between an estimation and the ground truth. 
  Spearman rank correlation is a measure of how well preserved the order of scores is.
  The MemNet shown in this table is the model created with PyTorch.
  Train time is generally correlated with inference time, but ResMemRetrain is an exception, it takes the same amount of time to estimate scores as ResMem.}\label{tab:results}
\end{table}

\subsection{MemNet}

We tested the publicly available version of MemNet, which yielded a Spearman rank correlation between predictions and human memory performance
of 0.565. This is within expected limits of the 0.57 reported in the paper 
 \citep{khoslaUnderstandingPredictingImage2015}. Beyond the rank correlation we 
can also examine other statistical properties of the predictions.

\begin{figure}
\centering
\begin{subfigure}[b]{0.49\textwidth}
\centering
\caption{MemNet in Caffe}
\includegraphics{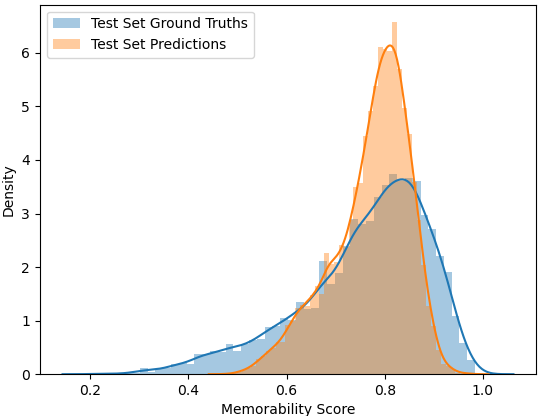}
\label{fig:cafmndist}
\end{subfigure}
\hfill
\begin{subfigure}[b]{0.49\textwidth}
\centering
\caption{MemNet in PyTorch}
\includegraphics{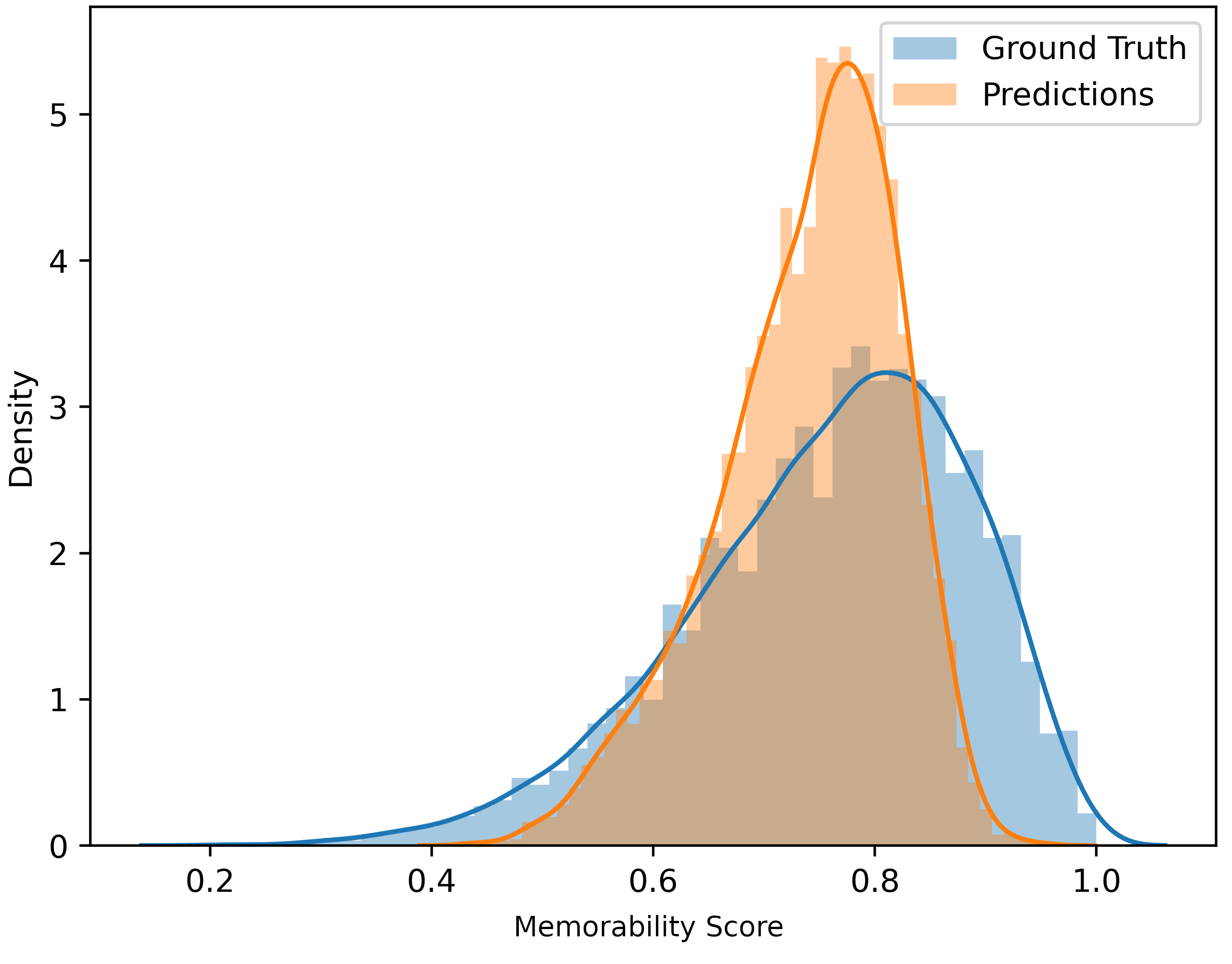}
\label{fig:mndist}
\end{subfigure}
\centering

\begin{subfigure}[b]{0.49\textwidth}
  \centering
  \caption{ResMem}
  \includegraphics{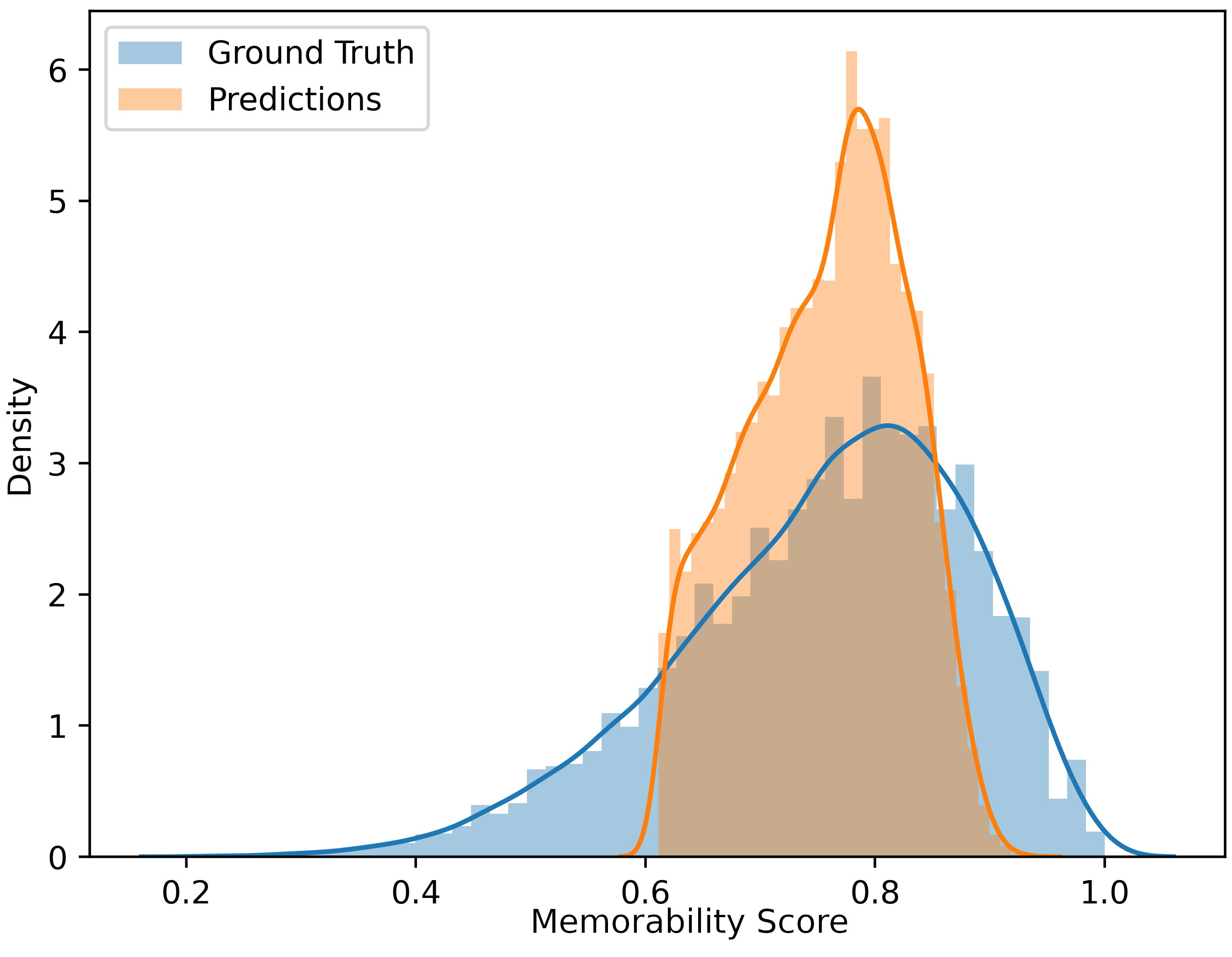}
  \label{fig:rmdist}
\end{subfigure}
\hfill
\begin{subfigure}[b]{0.49\textwidth}
  \hypertarget{fig:rmrdist}{%
  \centering
  \caption{ResMemRetrain}
  \includegraphics[width=\textwidth,height=9cm]{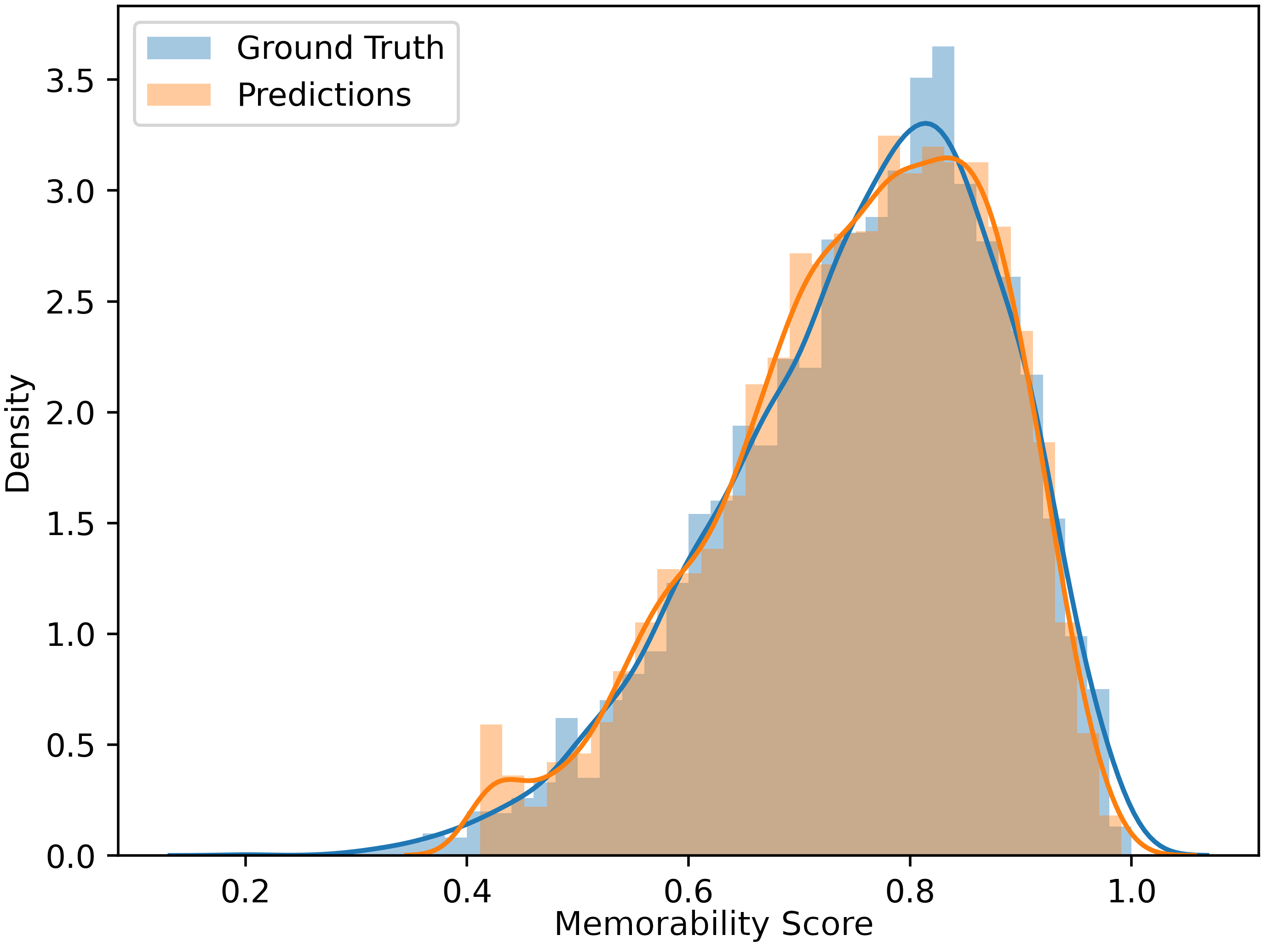}
  \label{fig:rmrdist}
  }
\end{subfigure}
\centering
  
\begin{subfigure}[b]{0.49\textwidth}
  \centering
  \caption{M3M}
  \includegraphics[width=\textwidth,height=9cm]{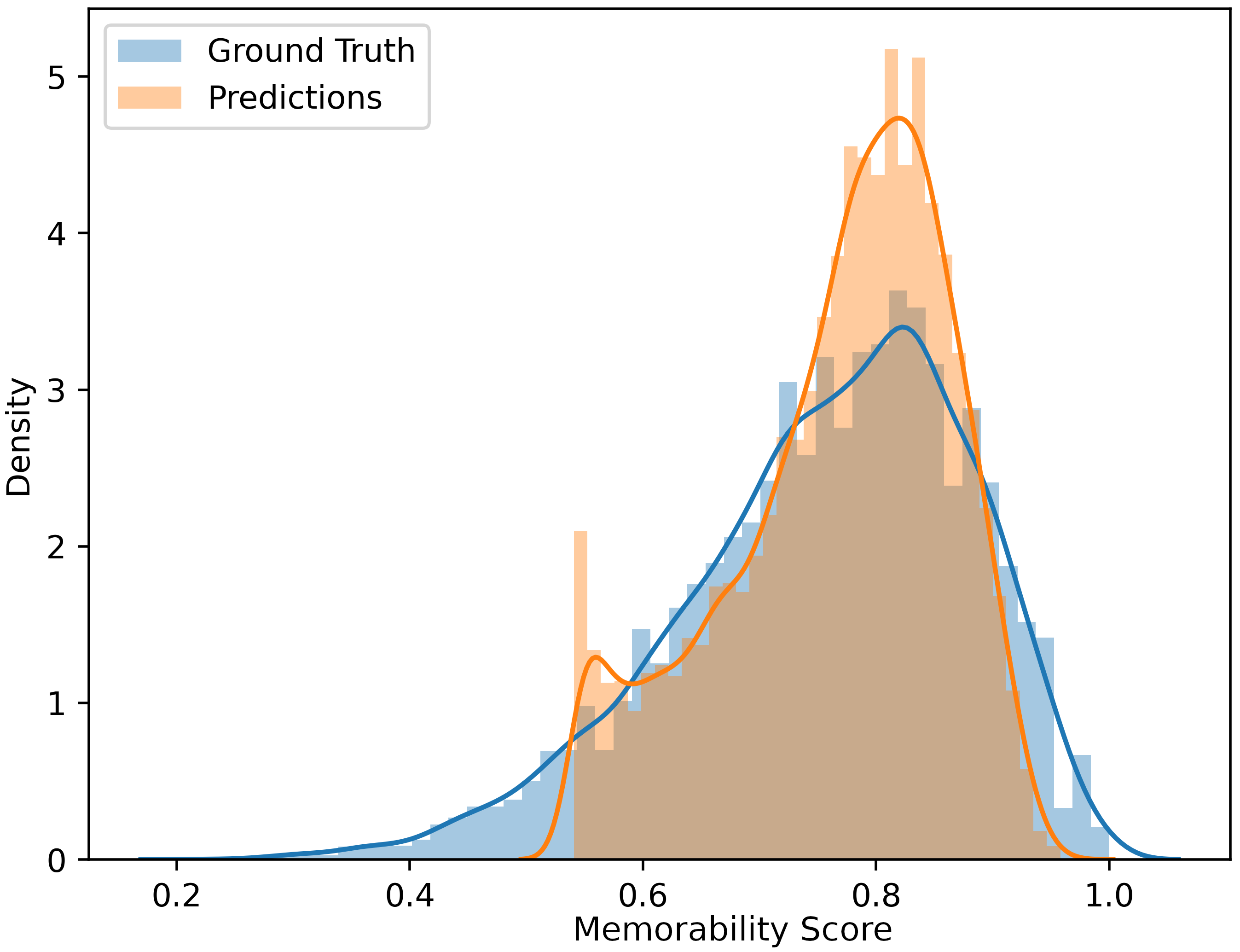}
  \label{fig:tripdist}
\end{subfigure}

\caption{Plots showing the distribution of model predictions compared with ground truths, for our five models MemNet in Caffe (a), MemNet in PyTorch (b), ResMem (c), ResMemRetrain (d), and M3M (e). Model predictions are shown
in orange, and human ground truths are shown in blue. These plots are created by estimating
the Probability Density Function of the memorability scores, which range from 0 to 1. These plots are drawn such that the area under the curve between two x-values is the probability of randomly drawing an x-value in that range.}

\end{figure}

Specifically, we can plot the distribution of ground truth memorability scores and compare them to 
the distribution of estimations (fig. \ref{fig:cafmndist}). An optimal model would produce the same distribution as the ground truths, and these visualizations can give us a deeper understanding of the model's statistical behavior. 
Note that passing this test does not show that the model is correct, since the distributions could be the same and have mismatched scores, giving a high loss and a low rank correlation.
Failing the test, however, is sufficient to show that the model under-performs.
A common issue with complex estimation models is regression towards the mean. 
This is when a model optimizes loss by predicting the mean value in the input data with some small variation.
If that small variation is correlated with the ground truth, then the model will achieve a good Spearman rank correlation, but a mediocre loss. The distribution of MemNet predictions shows a tendency for this regression towards the mean.

 When trained on the combination dataset, with early stopping determined by rank correlation
on a validation set, the PyTorch implementation of MemNet has a rank correlation 0.55 on a
held out test set, and an MSE Loss of 0.012 (Table \ref{tab:results}). This implies that, on
average, the memorability scores are accurate to 0.11. The reverse engineered MemNet's prediction distribution is fairly close to ground truth. 
However, the distribution is dense around the mean, and has a hard cutoff on the high
end (fig.~\ref{fig:mndist}).

\subsection{ResMem}

The ResMem model using the pre-existing ResNet-152 parameters had a rank correlation of 0.66 and a loss of 0.009 on the test set, which corresponds to accuracy within 0.092 on average. By looking at the prediction distribution for ResMem (fig. \ref{fig:rmdist}), 
we see that the shape is roughly correct for most of
the range, but there are sharp cutoffs at the high and low end. We can infer that the model is highly accurate within
the 0.6-0.9 range, although it struggles outside this region.

For ResMemRetrain,  the intermediate
features from ResNet were re-optimized to estimate memorability,
rather than to classify image category. While we can no longer interpret the
activation of those classification neurons as corresponding with
ImageNet categories, the model gained significant predictive power. ResMemRetrain is
the version of ResMem that has been published on the Brain Bridge Lab
\href{https://brainbridgelab.uchicago.edu/resmem}{website (https://brainbridgelab.uchicago.edu/resmem)} and on the Python Package Index (PyPI, \href{https://pypi.org/}{https://pypi.org/})
as \texttt{resmem}. This model achieves a rank
correlation of 0.67, similar to its counterpart, and achieves an MSE loss of 0.008. It is, on average, accurate to within 0.08 for
all images.  Compared to the others, ResMemRetrain has a distribution of predictions much closer to ground truth (fig. \ref{fig:rmrdist}). While prediction success is cut-off at the low end at 0.411, overall this is much closer to
the ground truth distribution. The reduced performance for low memorability images is likely due to the fact that only a few images have low memorability
scores. The lowest score in the dataset is 0.2, and the lowest score that
ResMemRetrain will assign is 0.411. We anticipate this is not an important 
issue since only 0.6\% of images in the dataset have a memorability
score below 0.411.

\subsection{M3M}

Testing with M3M yielded mixed results. While adding the semantic
segmentation features to the model does increase rank correlations to 0.68, this increase is very small compared to the heavy increase
in complexity. Additionally, MSE loss on the test set was 0.009, which is not meaningfully different from the MSE loss on ResMem and ResMemRetrain. This demonstrates that the semantic segmentation information
does not contribute to memorability any more than the higher level semantic 
data from ResMem does. While this is a null result, it is a null result
that can tell us about the state of memorability: the
exact location of semantic objects in the image does not matter as much
as the simple existence of the semantic objects.

The distribution plot for M3M displays similar behavior to 
ResMem (fig. \ref{fig:tripdist}). This also
supports our earlier conclusion that M3M's performance increases in
terms of rank correlation and loss are not worth the added costs. One
training run of M3M took 30 hours to train on an Nvidia 1080TI. For
comparison, ResMemRetrain takes 15 hours to run through the same amount
of training steps, and ResMem without retraining and MemNet take 80
minutes.

\hypertarget{model-analysis}{%
\subsection{Model Evaluation}\label{model-analysis}}

\begin{figure}
    \centering
    \[\arraycolsep=6pt
    \newcolumntype{L}{>{\centering\arraybackslash} m{2.2cm} }
    \newcolumntype{K}{>{\centering\arraybackslash} m{2.2cm} }
    \begin{array}{L K K K} 
    \textbf{LaMem Low Loss} &\includegraphics[height=2.5cm]{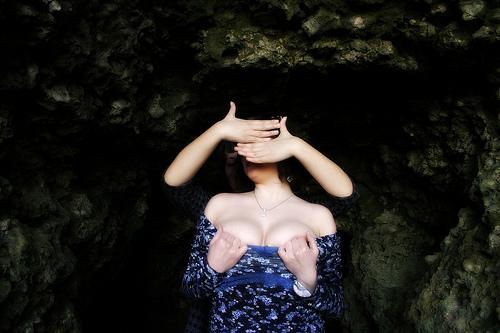} & \includegraphics[height=2.5cm]{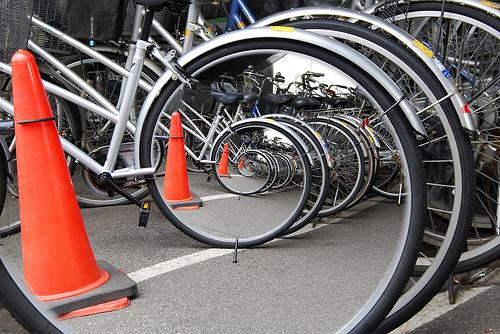} & \includegraphics[height=2.5cm]{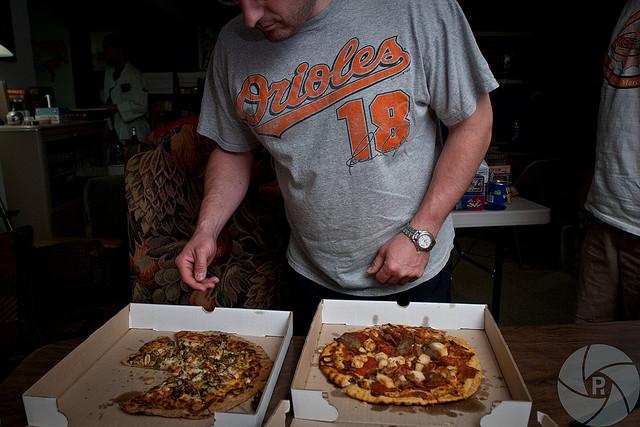} \\
    {\small Memorability:} & 0.769, 0.769 &  0.809, 0.809 &  0.827, 0.827 \\
    {\small SE Loss:} & \num{8.31e-12} & \num{2.22e-12} & \num{3.71e-13} \\ 
    \end{array}
    \] 
    \[\arraycolsep=6pt
    \newcolumntype{L}{>{\centering\arraybackslash} m{2.2cm} }
    \newcolumntype{K}{>{\centering\arraybackslash} m{2.2cm} }
    \begin{array}{L K K K} 
    \textbf{LaMem High Loss} &\includegraphics[height=2.5cm]{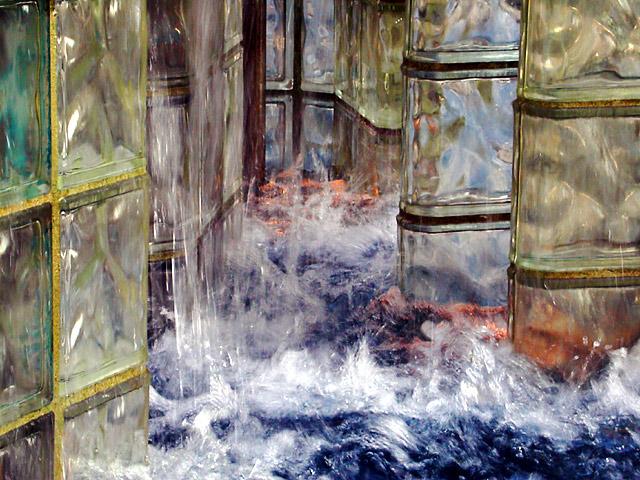} & \includegraphics[height=2.5cm]{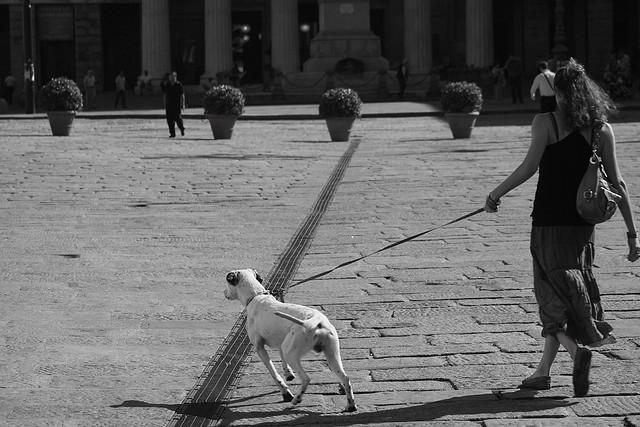} & \includegraphics[height=2.5cm]{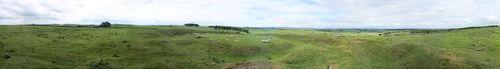} \\
    {\small Memorability:} & 0.314, 0.748 & 0.429, 0.849 & 0.805, 0.424 \\
    {\small SE Loss:} & 0.188 & 0.177 & 0.145
    \end{array}
    \] 
    \[\arraycolsep=6pt
    \newcolumntype{L}{>{\centering\arraybackslash} m{2.2cm} }
    \newcolumntype{K}{>{\centering\arraybackslash} m{2.2cm} }
    \begin{array}{L K K K} 
    \textbf{MemCat Low Loss} &\includegraphics[height=2.5cm]{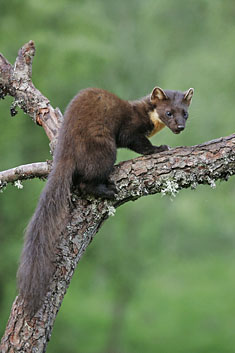} & \includegraphics[height=2.5cm]{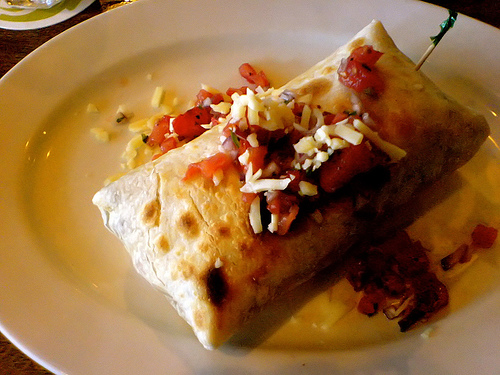} & \includegraphics[height=2.5cm]{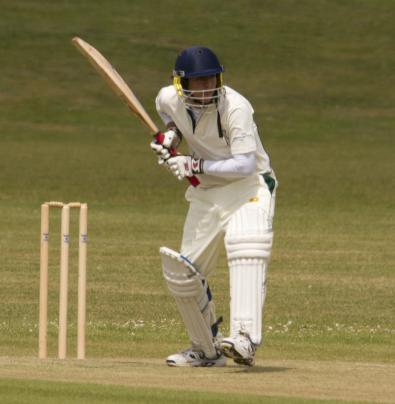} \\
    {\small Memorability:} & 0.802, 0.802 & 0.838, 0.838 & 0.667, 0.667  \\
    {\small SE Loss:} & \num{3.79e-13}& \num{1.47e-11} & \num{3.99e-11}
    \end{array}
    \] \
    \[\arraycolsep=6pt
    \newcolumntype{L}{>{\centering\arraybackslash} m{2.2cm} }
    \newcolumntype{K}{>{\centering\arraybackslash} m{2.2cm} }
    \begin{array}{L K K K} 
    \textbf{MemCat High Loss} &\includegraphics[height=2.5cm]{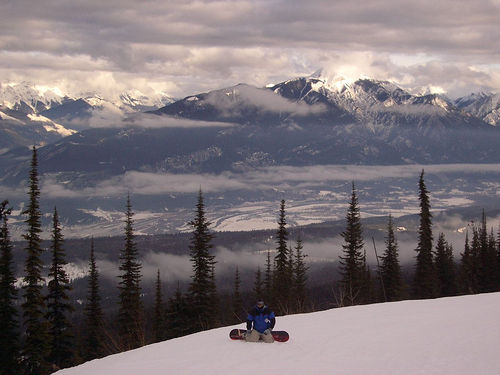} & \includegraphics[height=2.5cm]{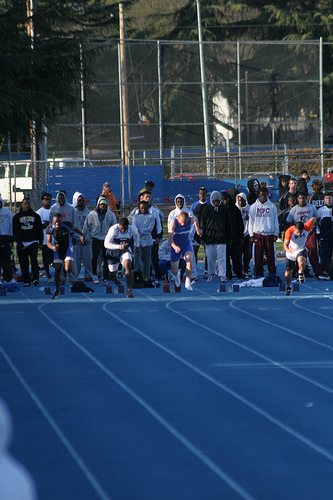} & \includegraphics[height=2.5cm]{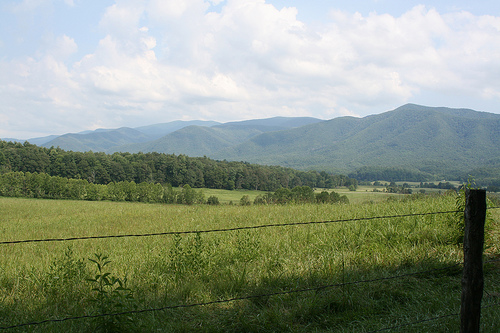} \\
    {\small Memorability:} & 0.407, 0.815 & 0.429, 0.767 & 0.761, 0.432 \\
    {\small SE Loss:} & 0.167 & 0.115 & 0.108 
    \end{array}
    \]
    \caption{Highest and lowest estimation loss images from each source dataset. Memorability scores appear below each image, with the first value being the ground truth, and the second value being the estimation by ResMemRetrain. Mean Squared Error Loss appears below that, where lower numbers indicate a more accurate estimation.}
    \label{fig:loss_producing_images}
\end{figure}

Overall our new models outperform MemNet, while still maintaining reasonable train times  (Table \ref{tab:results}). Figure \ref{fig:loss_producing_images} shows the images for which our model makes the most successful predictions, and which ones have the largest prediction error.
Adding a Residual Neural Network step to the model improves predictions by a considerable margin. 
Since ResMem employs both conceptual and perceptual features in its estimation process instead of just perceptual features, this implies that both types of features contribute to memorability.
While the improvement on the general performance metrics shows that our multi-feature models outperform MemNet, it can also be useful to examine the other statistical properties of the model's estimations.

Given the improvement in the model made by the addition of the semantic features from ResNet, one hypothesis could be that image category information alone is sufficient for making successful predictions of memorability. To test this question, we used a ResNet pretrained for image category classification, and conducted a simple linear regression on the estimated category membership for predicting memorability. This model achieves a Spearman rank correlation of 0.477, while using AlexNet instead yields a rank correlation of 0.425. Thus, while just the image category alone can predict memorability to some degree, this basic model under-performs all models reported here that are designed specifically for predicting memorability.

Another interesting question is how well ResMemRetrain can predict the memorability within individual categories; are there some image categories that are better predicted than others? To test this, we evaluated the accuracy of ResMemRetrain across the categories in MemCat. This is presented with the caveat that MemCat is not large enough to have a sizable sample in every category in both the training and the test set, so the figures presented are calculated on the entire image set, and includes images present in both the training and test set. Thus, the performance measures should be considered as relative measures to compare across categories, rather than absolute measures of performance. The top 5 and bottom 5 subcategories defined by mean squared prediction error are presented in Table \ref{ref:subcat_rank} and all superordinate MemCat categories are presented in Table \ref{ref:supercat_rank}. In general, our model is best able to predict the memorability of landscape images in comparison to images of animals in terms of rank correlation. The full table of all subcategories is available on \href{https://osf.io/qf5ry/}{OSF}.

\begin{table}[h!]
\centering
    \begin{tabular}{l | c c c c c c}
   Category & MSE & Rank Corr & $\hat{y}$ Mean & $y$ Mean & $\hat{y}$ SD & $y$ SD \\ 
   \hline
champagne & 0.0018 & 0.6891 & 0.8755 & 0.8676 & 0.0563 & 0.0540\\ 
tennis & 0.0018 & 0.8830 & 0.7731 & 0.7749 & 0.0854 & 0.0858\\ 
zucchini & 0.0020 & 0.6575 & 0.9098 & 0.8975 & 0.0602 & 0.0594\\ 
hotdog & 0.0020 & 0.5391 & 0.8824 & 0.8954 & 0.0743 & 0.0710\\ 
cocktail & 0.0023 & 0.7175 & 0.8624 & 0.8467 & 0.0569 & 0.0522\\ 
\vdots & \vdots & \vdots & \vdots & \vdots & \vdots & \vdots \\
mountain & 0.0088 & 0.4904 & 0.5069 & 0.5073 & 0.0844 & 0.1140\\ 
forest (broadleaf) & 0.0090 & 0.6619 & 0.5323 & 0.4994 & 0.0904 & 0.1234\\ 
snowboarding & 0.0092 & 0.4526 & 0.7156 & 0.7135 & 0.0831 & 0.0861\\ 
zebra & 0.0109 & 0.4076 & 0.7473 & 0.8260 & 0.0694 & 0.0604\\ 
giraffe & 0.0131 & 0.4502 & 0.7407 & 0.8306 & 0.0721 & 0.0645\\ 
    \end{tabular}
    \caption{Memorability and performance statistics on MemCat broken down by MemCat subcategory. Tests were performed with the release version of ResMemRetrain available in the \texttt{resmem} package. $\hat{y}$ indicates the set of memorability score predictions and $y$ indicates the set of ground truth memorability scores. MSE is Mean Square Error and Rank Corr is Spearman's Rank Correlation.}
    \label{ref:subcat_rank}
\end{table}
\begin{table}[h!]
    \centering
    \nprounddigits{3}
    \begin{tabular}{l | c c c c c c}
   Category & MSE & Rank Corr & $\hat{y}$ Mean & $y$ Mean & $\hat{y}$ SD & $y$ SD \\ 
   \hline
   food & 0.0031 & 0.6991 & 0.8513 & 0.8532 & 0.0774 & 0.0771\\ 
vehicle & 0.0043 & 0.7409 & 0.7548 & 0.7572 & 0.0949 & 0.0942\\ 
sports & 0.0046 & 0.6862 & 0.7690 & 0.7809 & 0.0815 & 0.0873\\ 
animal & 0.0052 & 0.6761 & 0.8085 & 0.8028 & 0.0991 & 0.0934\\ 
landscape & 0.0067 & 0.8071 & 0.5812 & 0.5981 & 0.1087 & 0.1327\\ 
    \end{tabular}
    \caption{Memorability and performance statistics on MemCat broken down by MemCat superordinate category, following the same format as Table \ref{ref:subcat_rank}.}
    \label{ref:supercat_rank}
\end{table}

\pagebreak

\subsection{Network Feature Analysis}
\begin{figure}
    \centering
    \[\arraycolsep=0.5pt\def\arraystretch{2.2}
    \newcolumntype{K}{>{\centering\arraybackslash} m{2.1cm} }
    \newcolumntype{L}{>{\centering\arraybackslash} m{1cm} }
    \begin{array}{L K K K K K}
         Layer 13 & 
         \includegraphics[height=2.3cm]{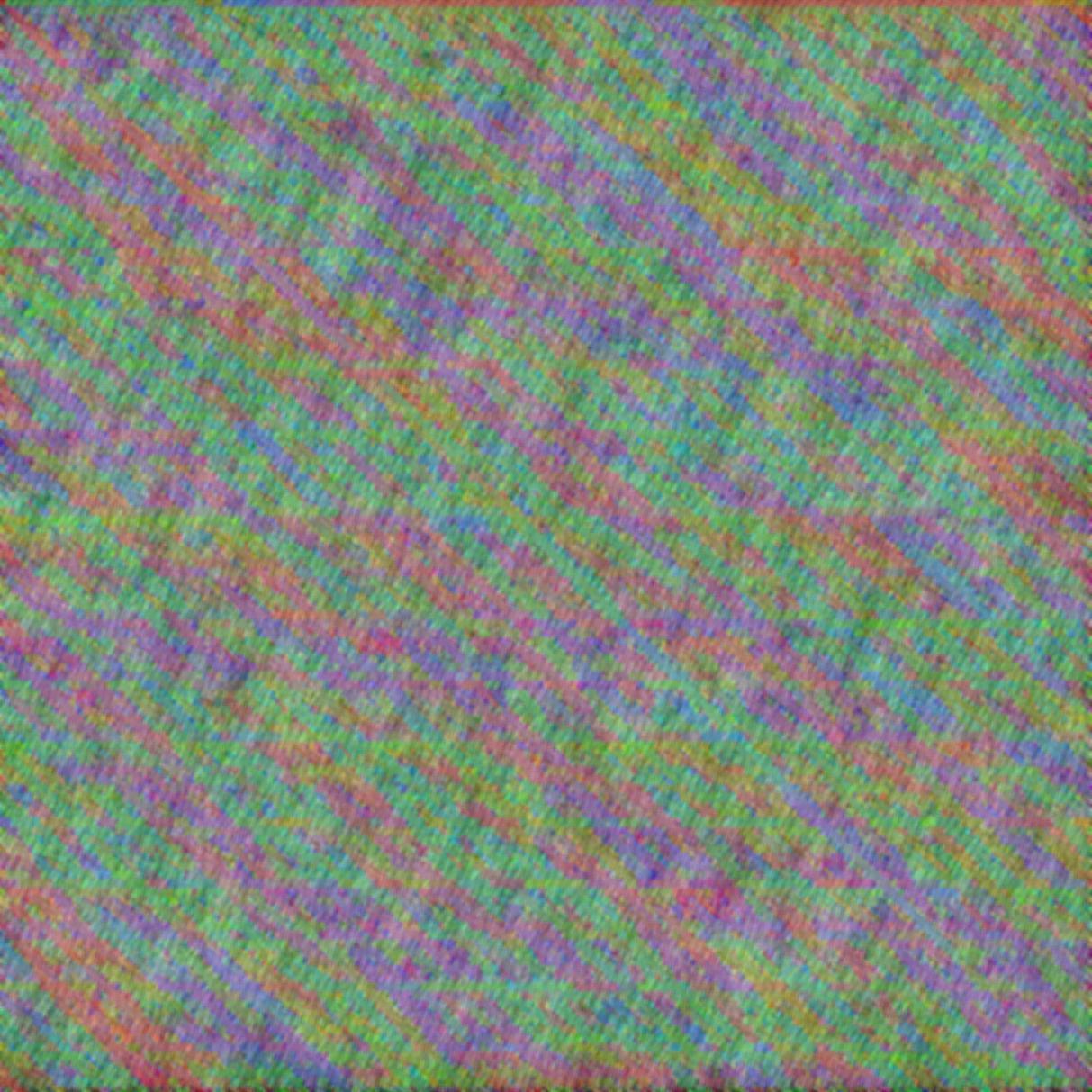} &
         \includegraphics[height=2.3cm]{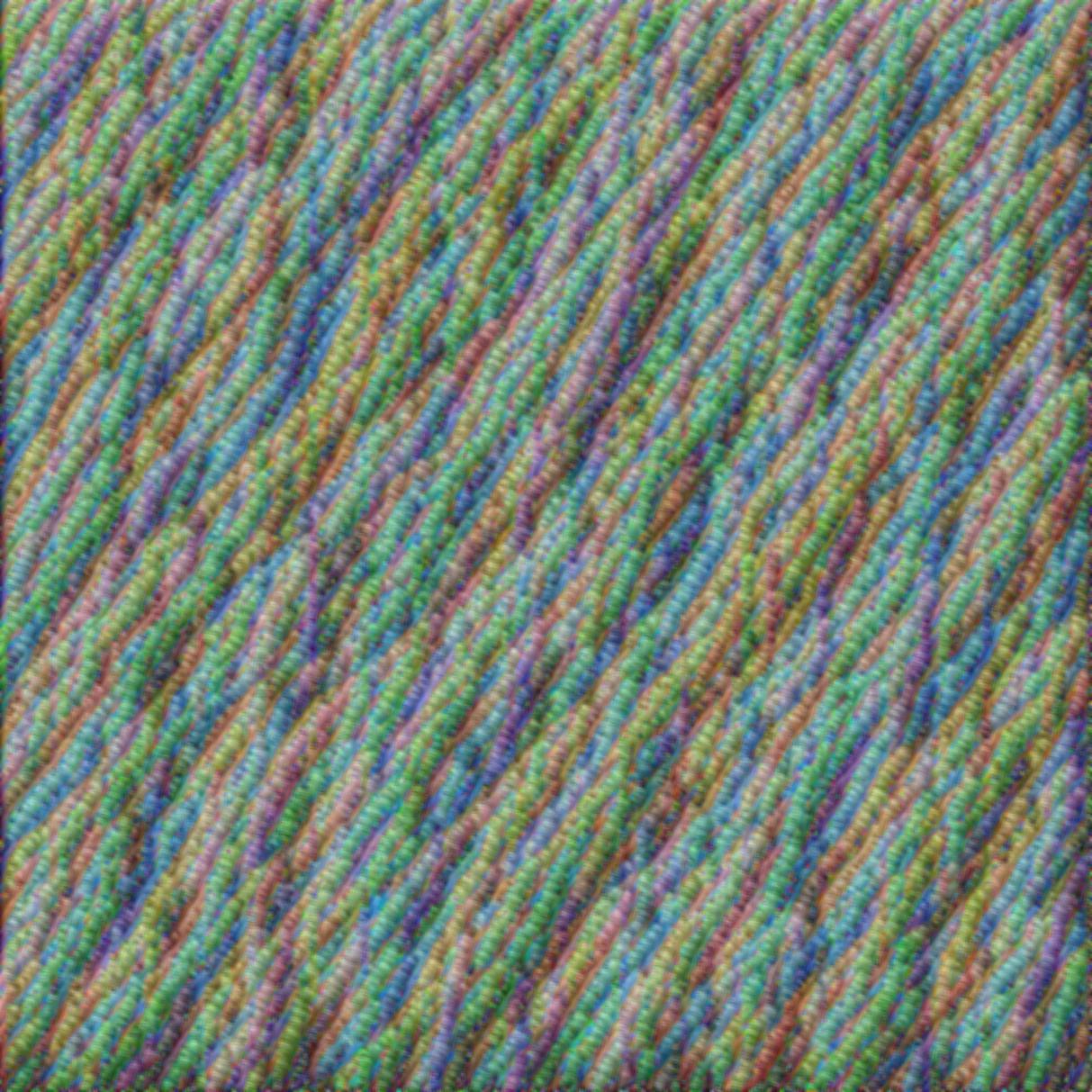} &
         \includegraphics[height=2.3cm]{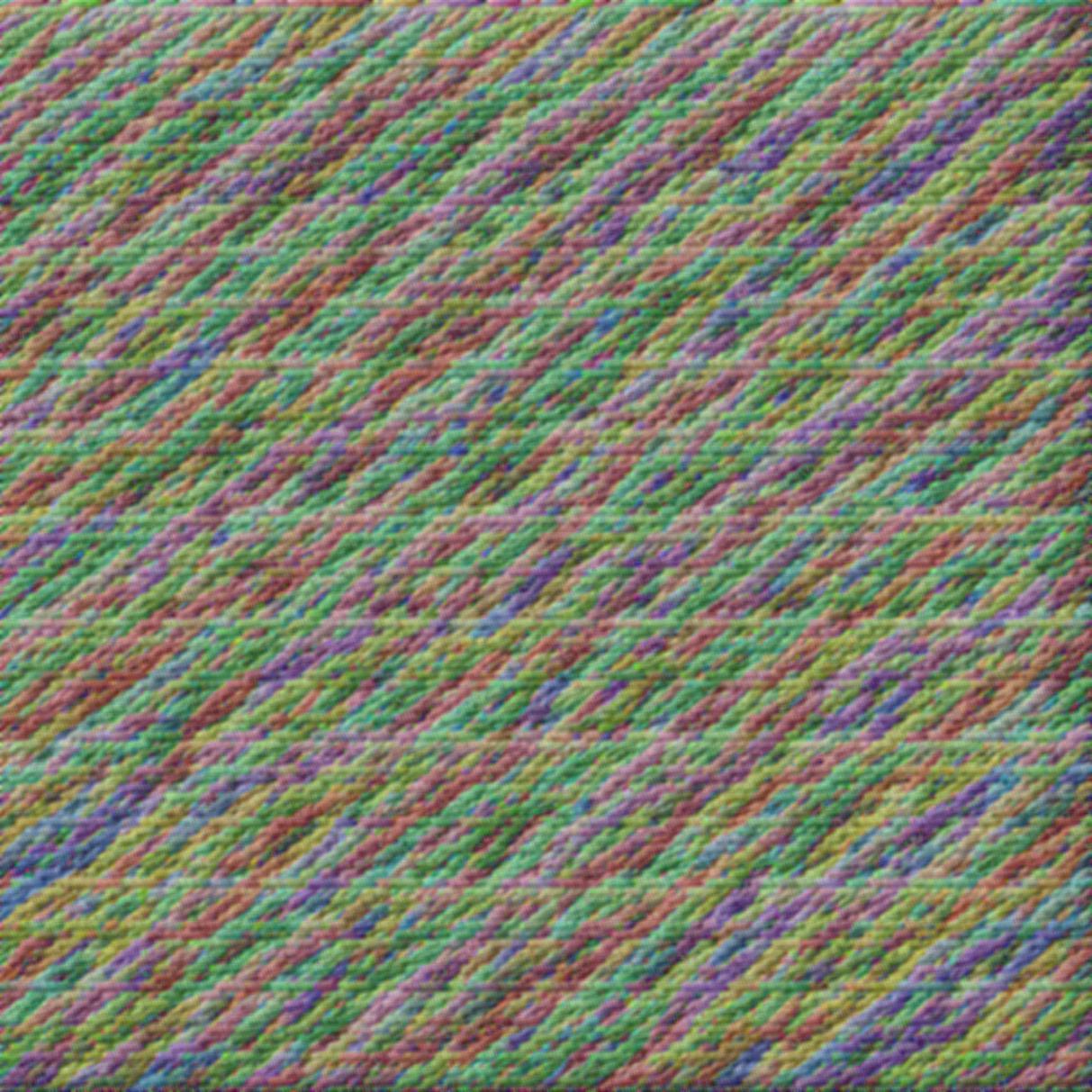} &
         \includegraphics[height=2.3cm]{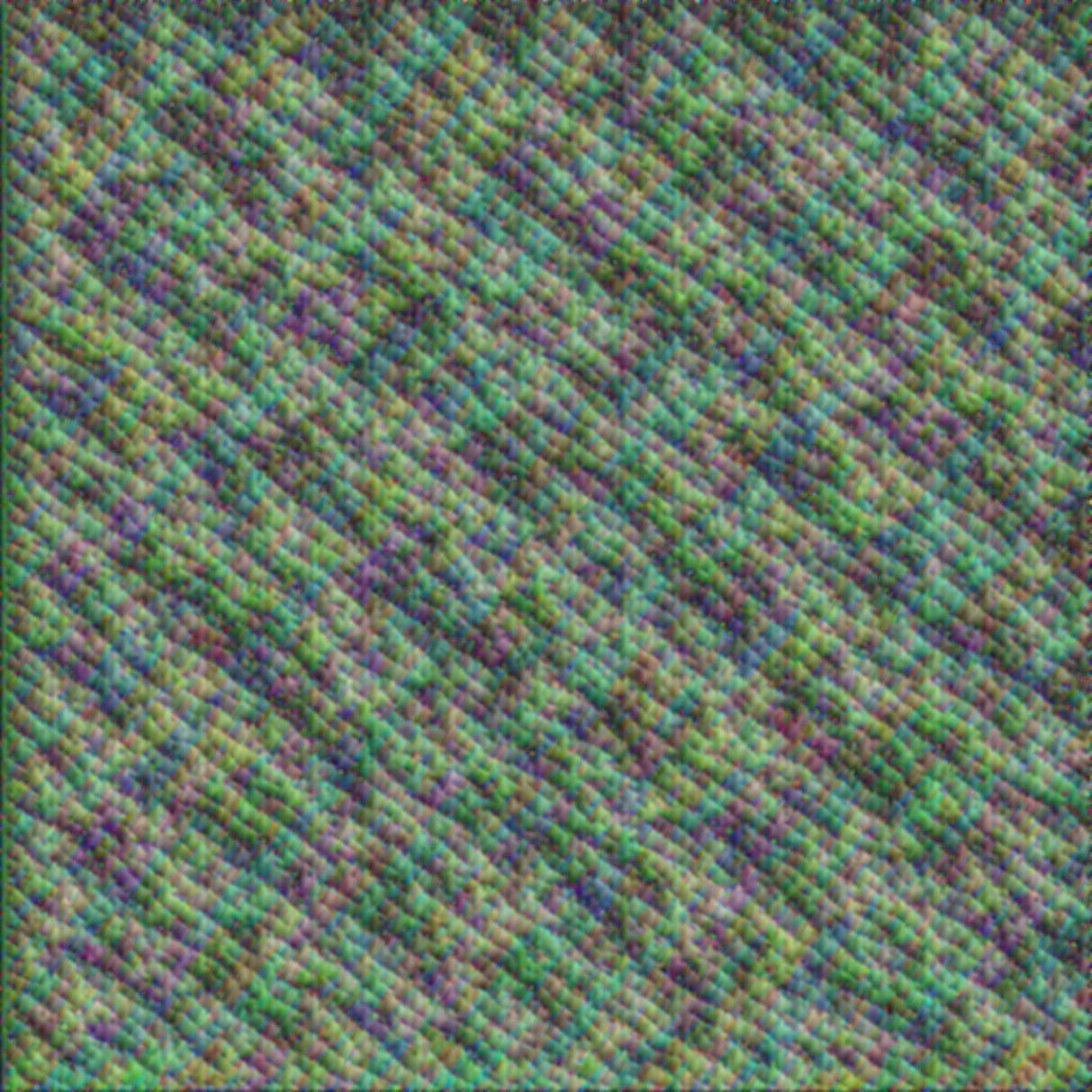} &
         \includegraphics[height=2.3cm]{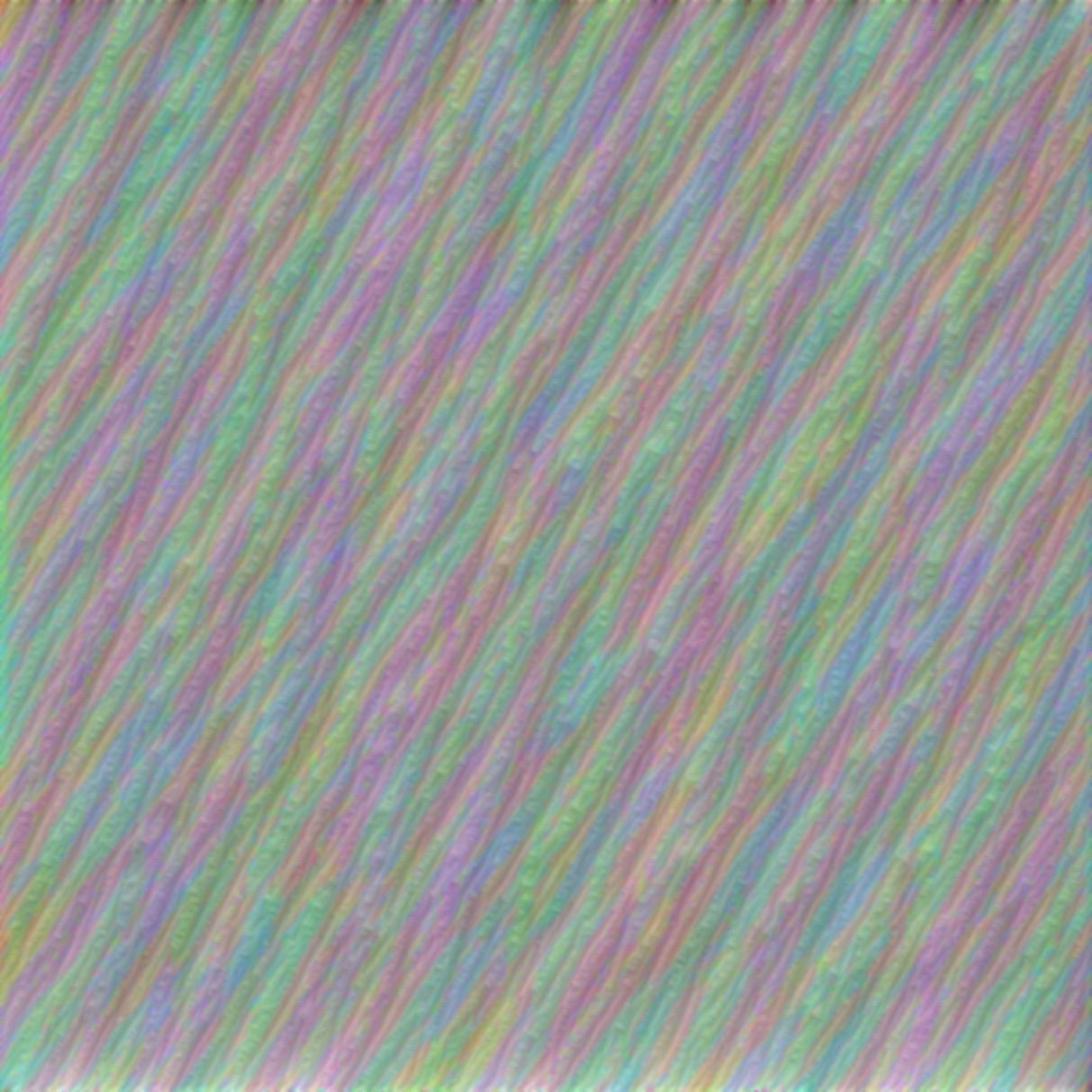} \\
         Layer 35 & 
         \includegraphics[height=2.3cm]{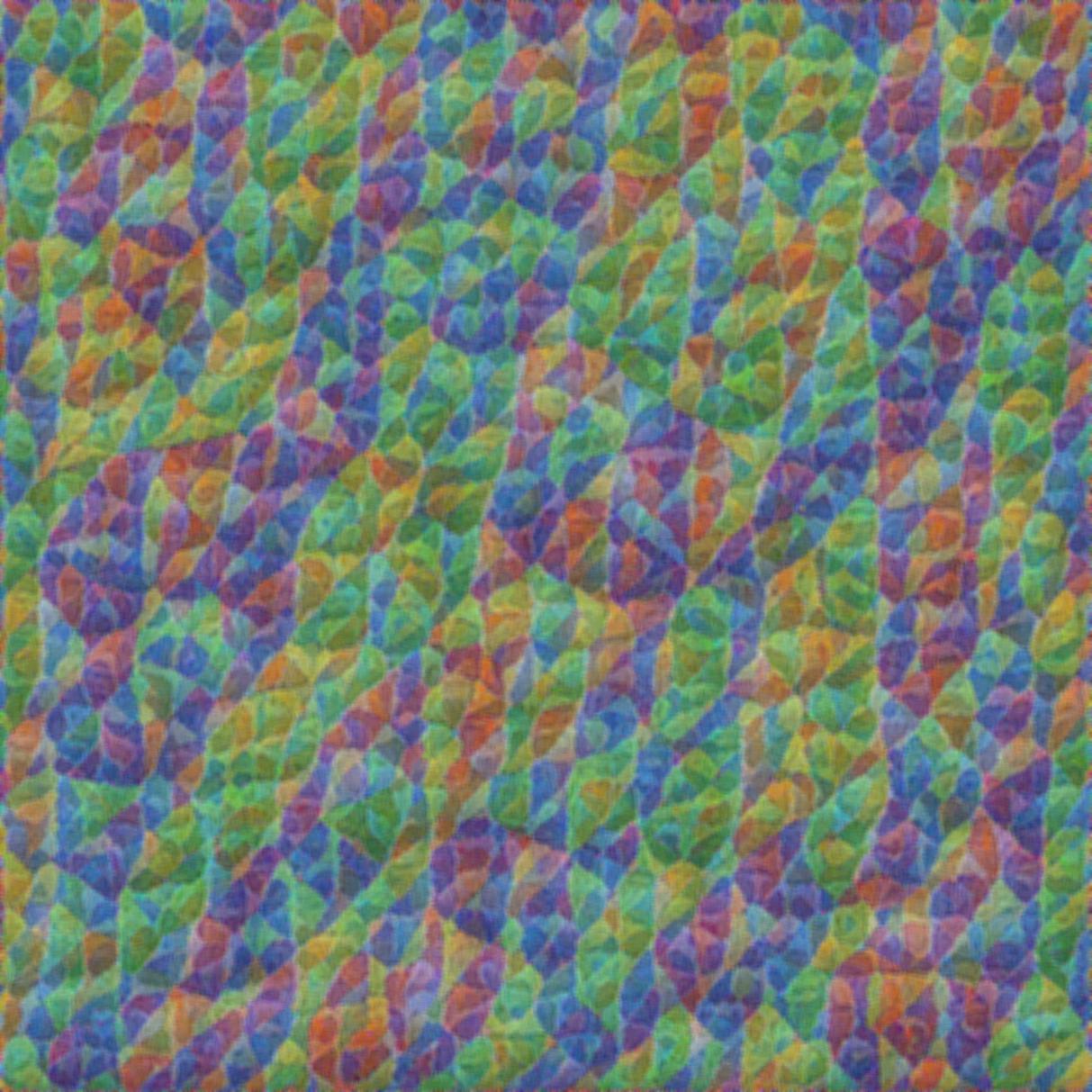} &
         \includegraphics[height=2.3cm]{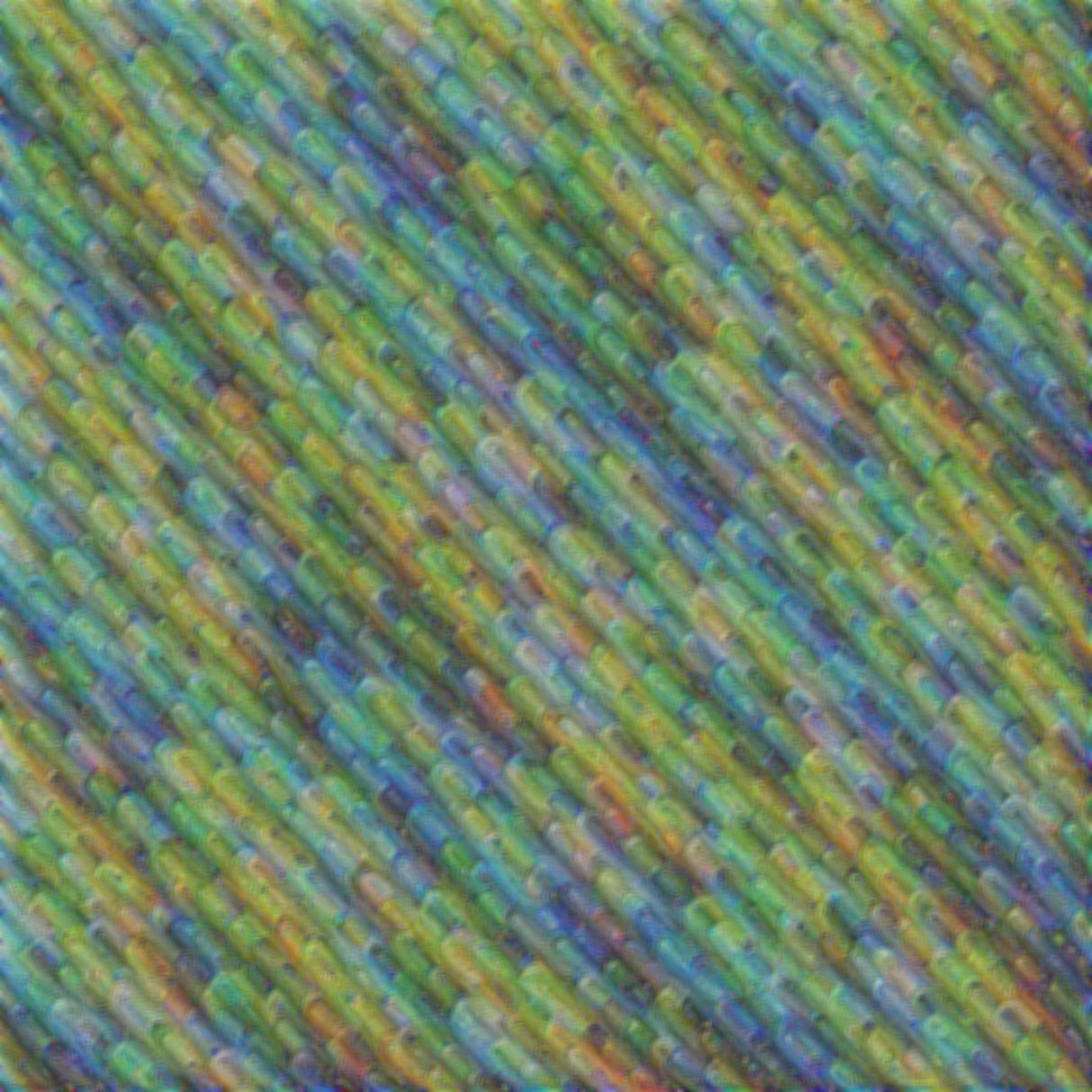} &
         \includegraphics[height=2.3cm]{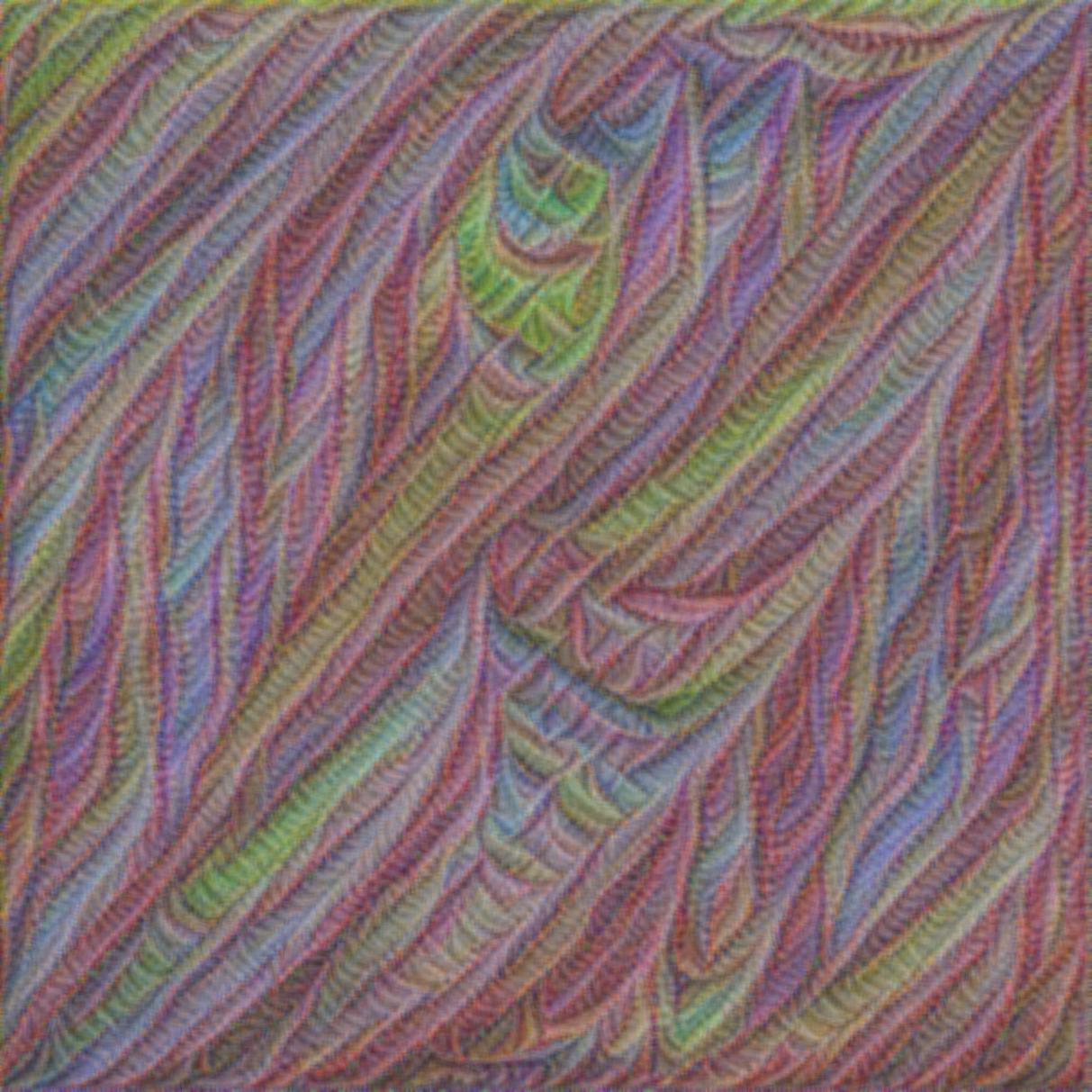} &
         \includegraphics[height=2.3cm]{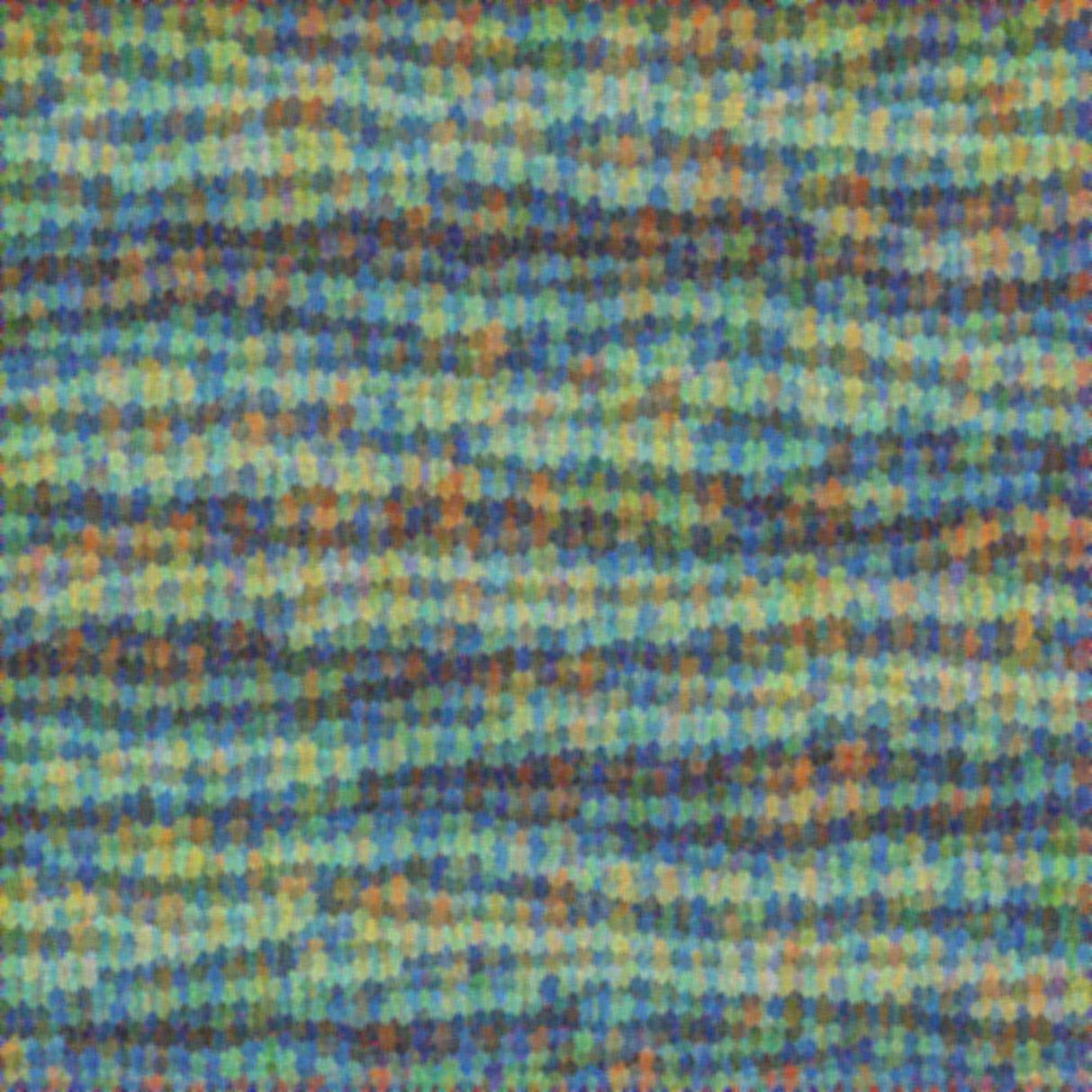} &
         \includegraphics[height=2.3cm]{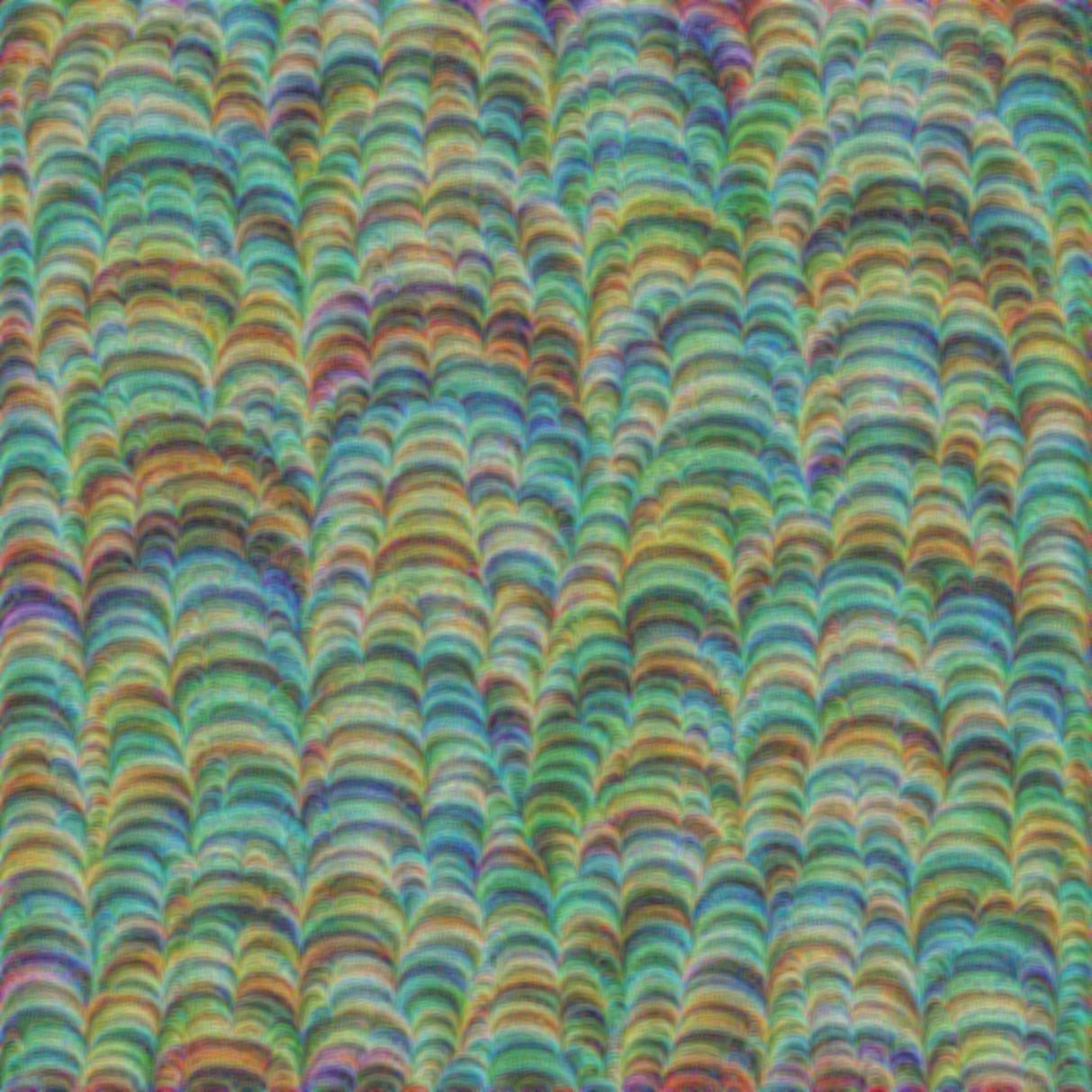} \\
         Layer 81 & 
         \includegraphics[height=2.3cm]{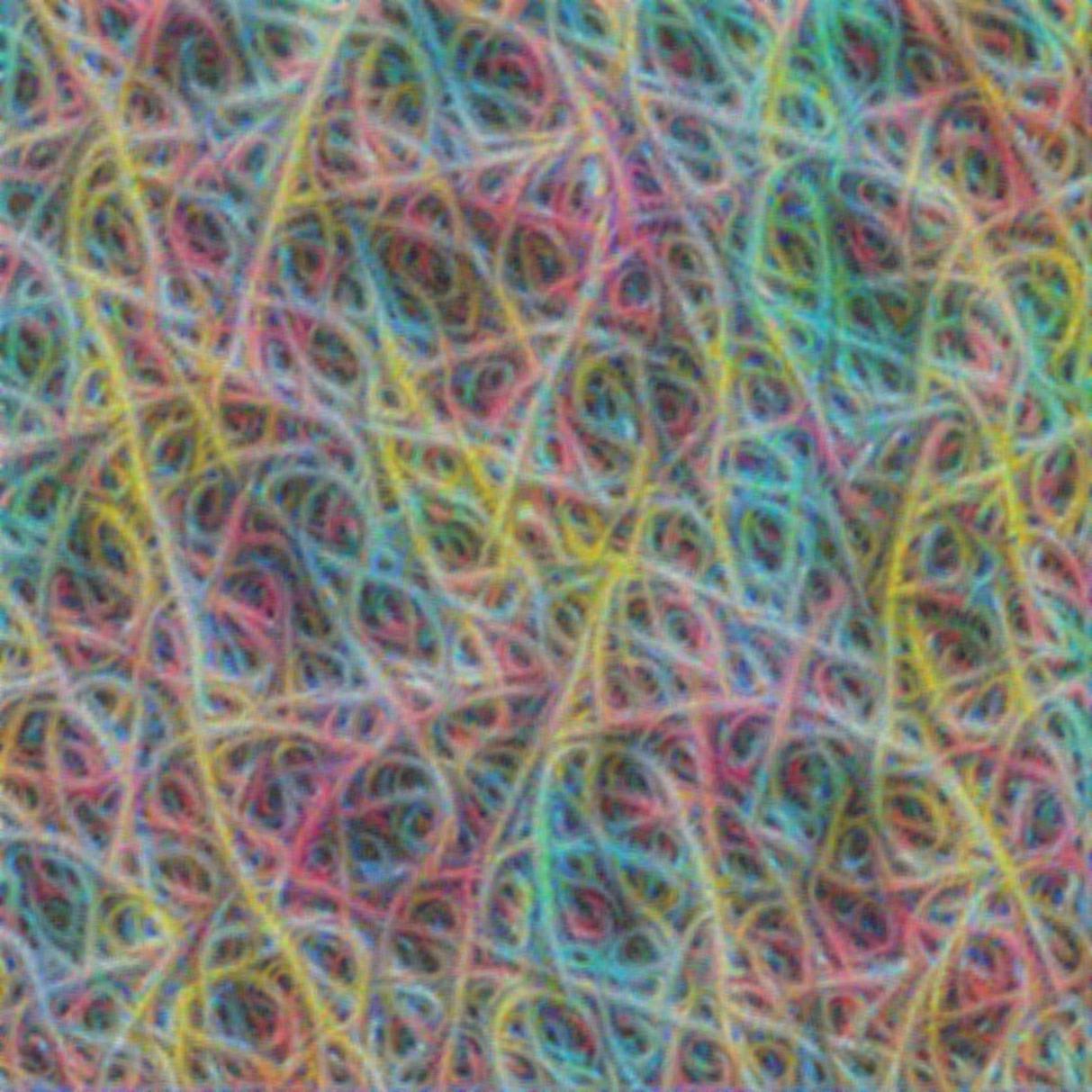} &
         \includegraphics[height=2.3cm]{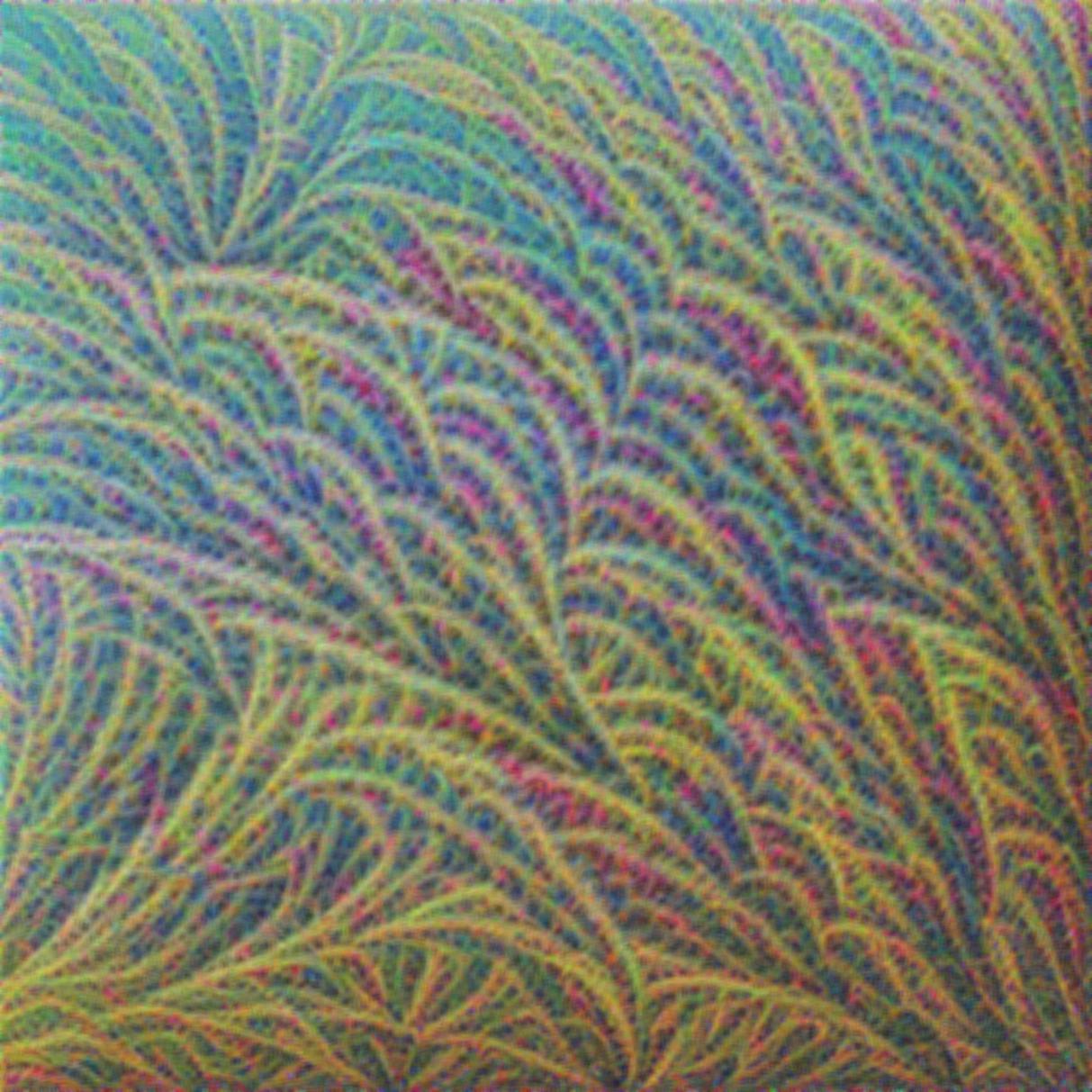} &
         \includegraphics[height=2.3cm]{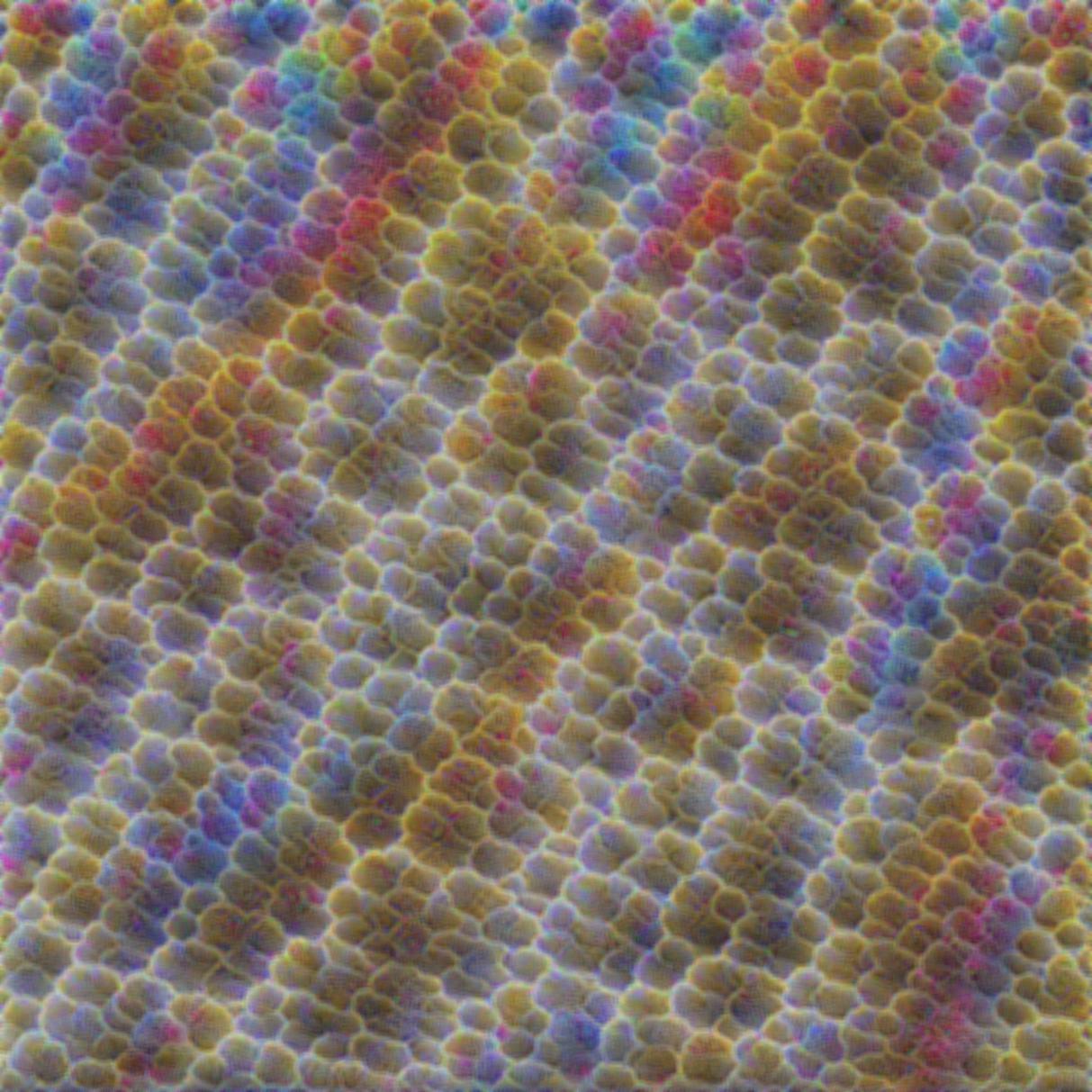} &
         \includegraphics[height=2.3cm]{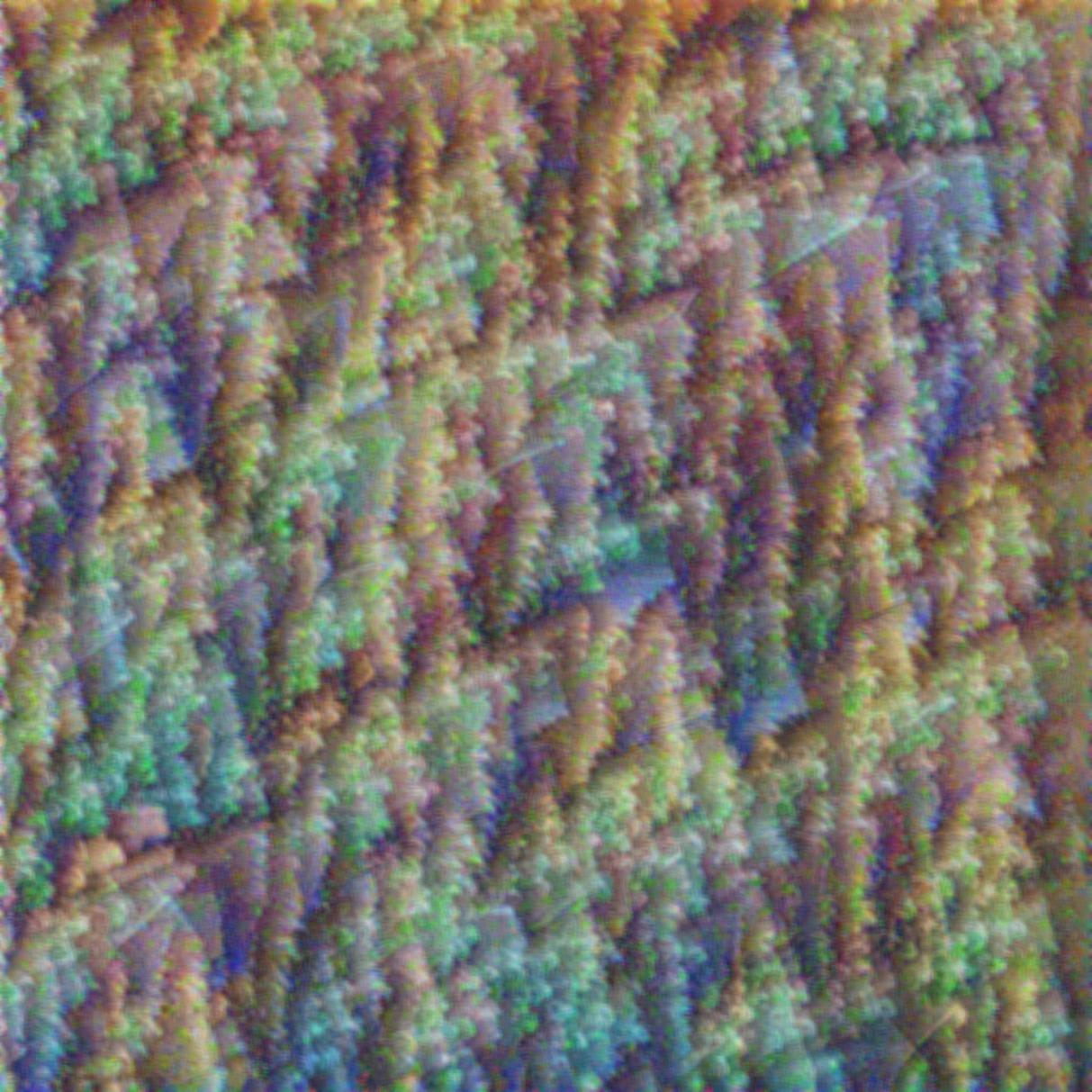} &
         \includegraphics[height=2.3cm]{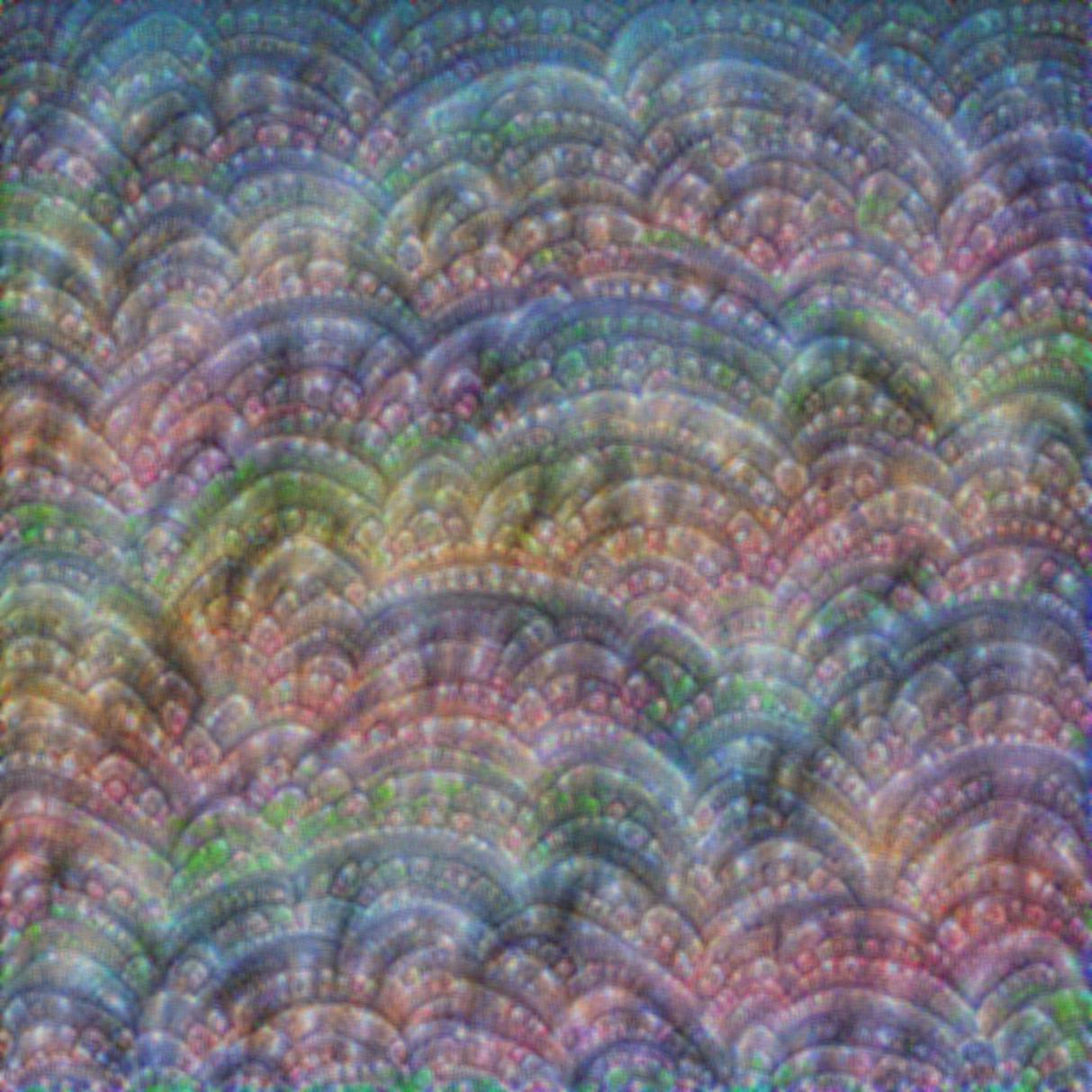} \\
         Layers 96-126 & 
         \includegraphics[height=2.3cm]{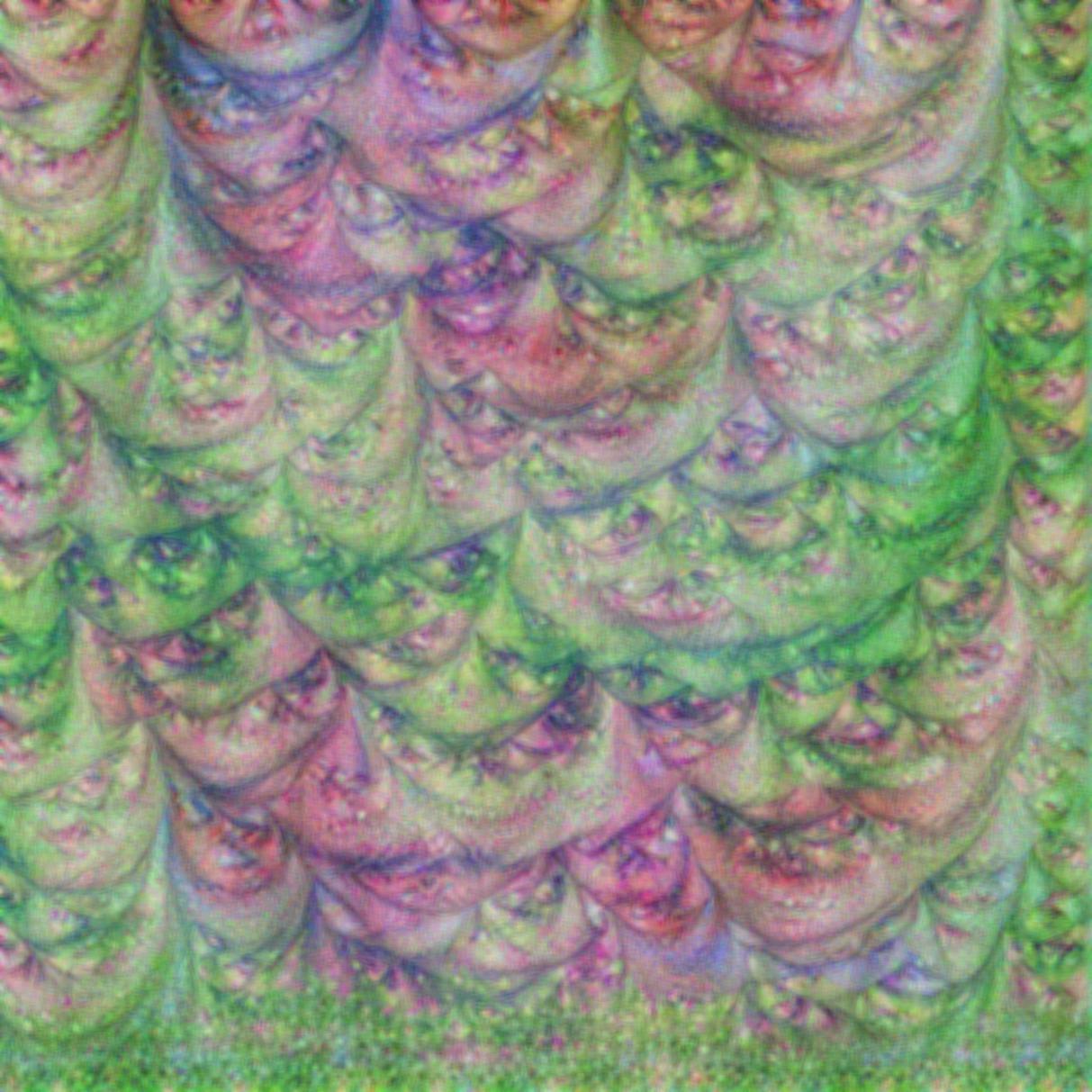} &
         \includegraphics[height=2.3cm]{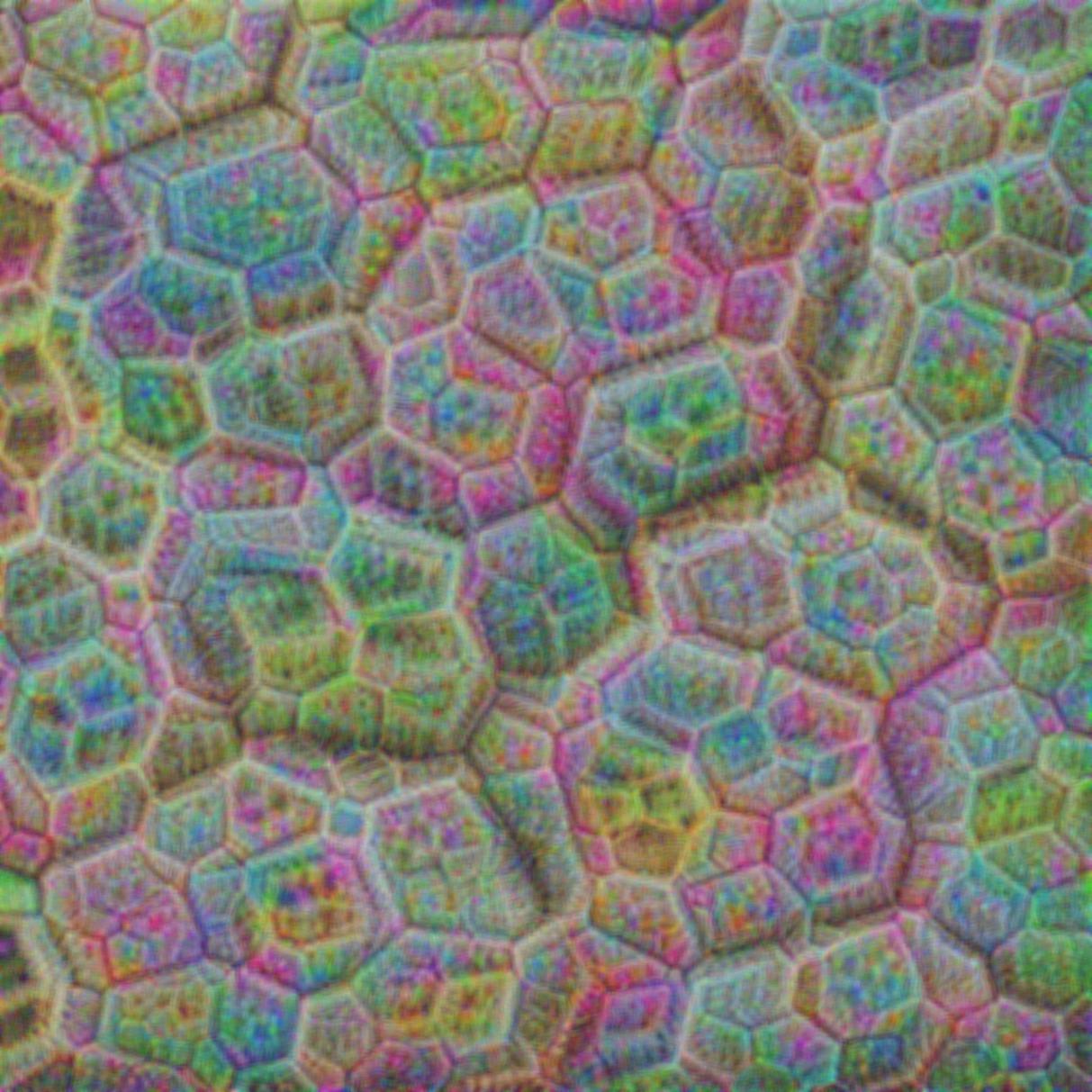} &
         \includegraphics[height=2.3cm]{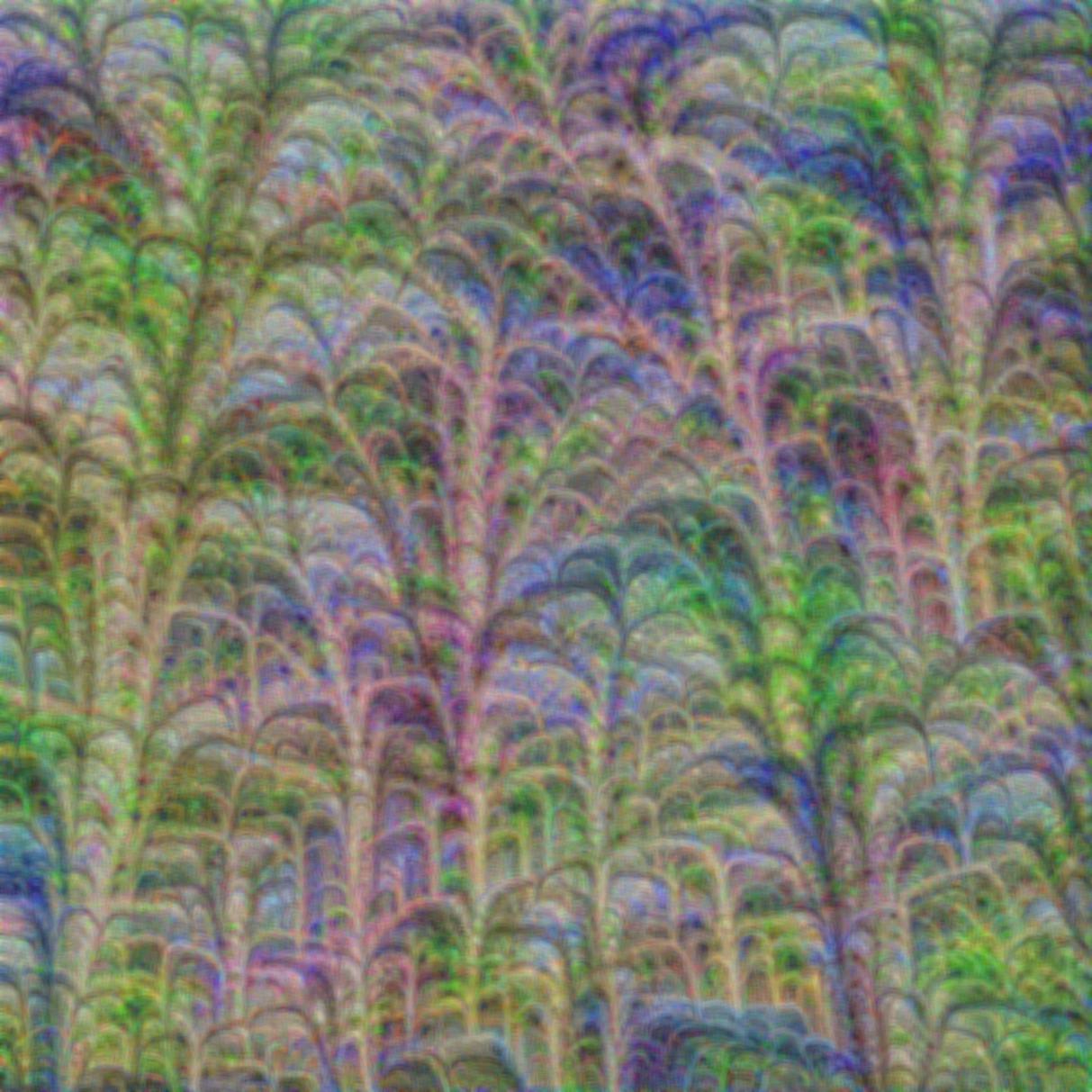} &
         \includegraphics[height=2.3cm]{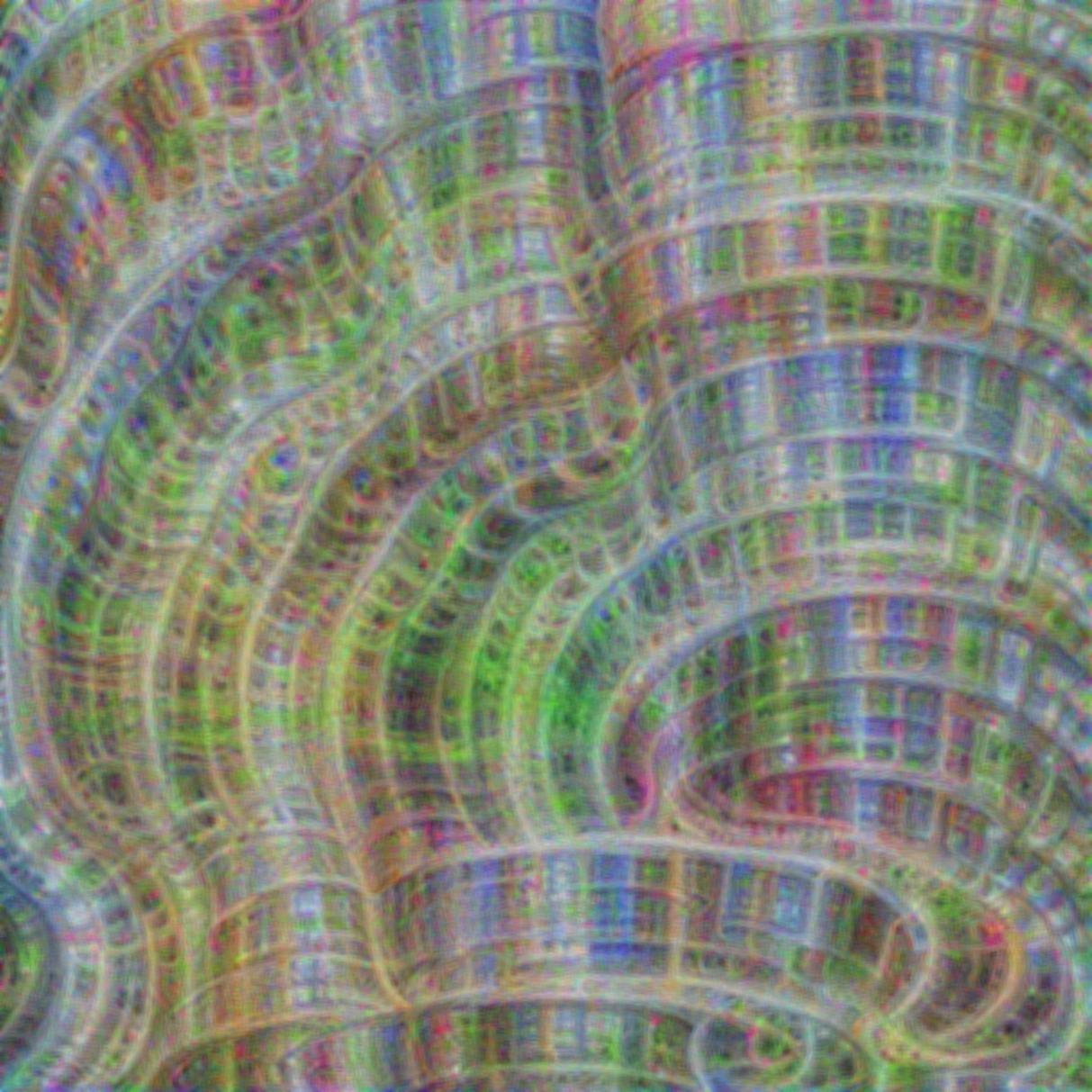} &
         \includegraphics[height=2.3cm]{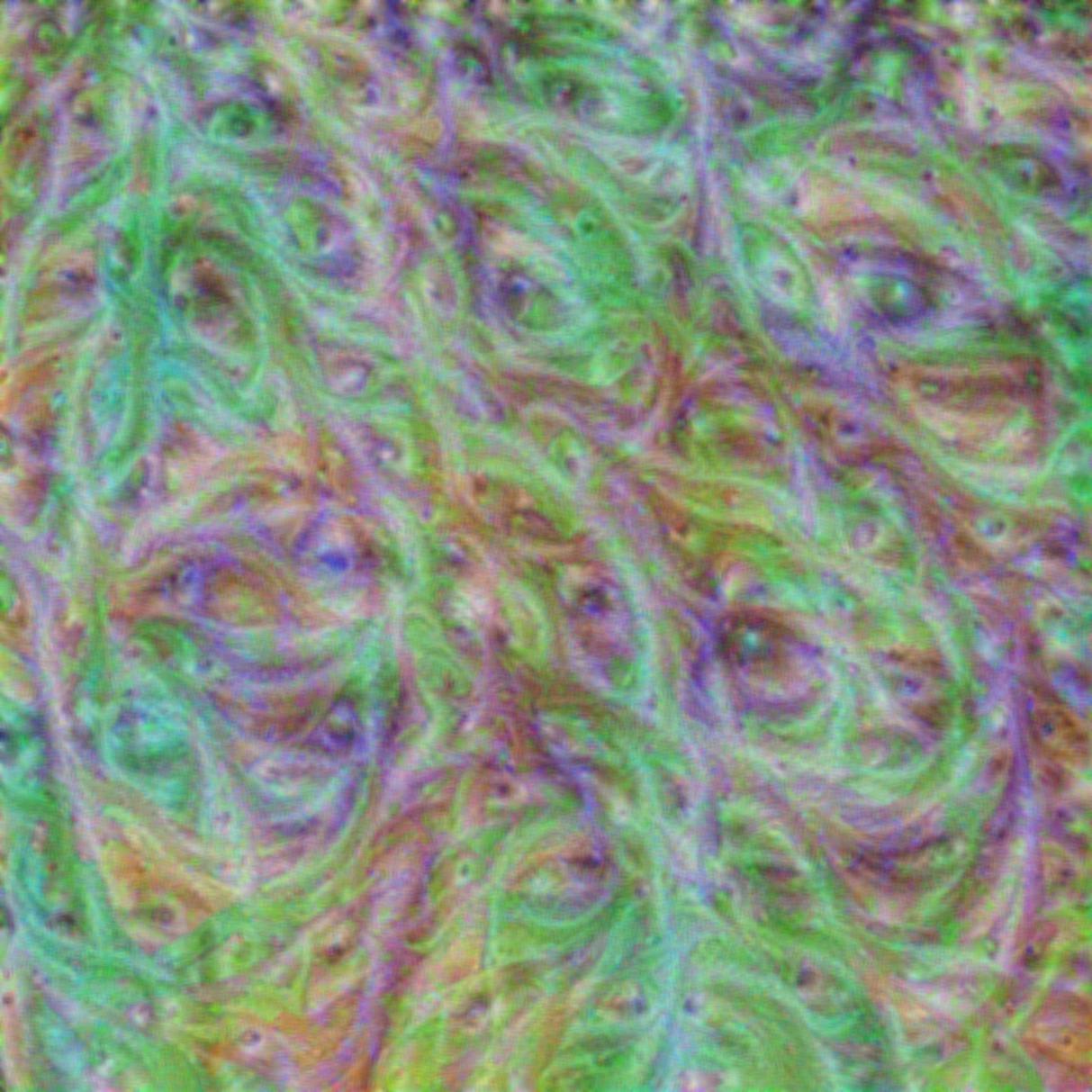} \\
         Layer 139 & 
         \includegraphics[height=2.3cm]{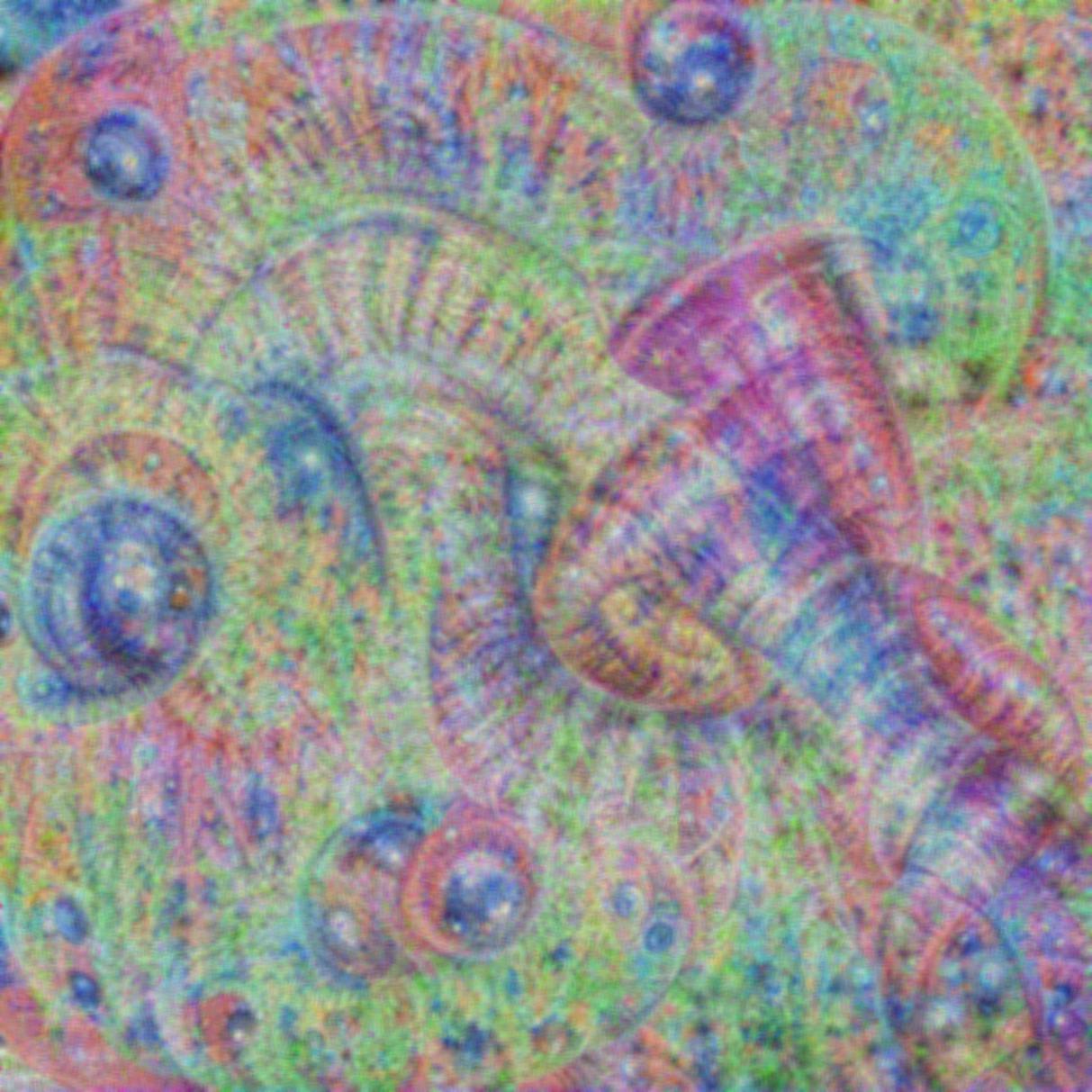} &
         \includegraphics[height=2.3cm]{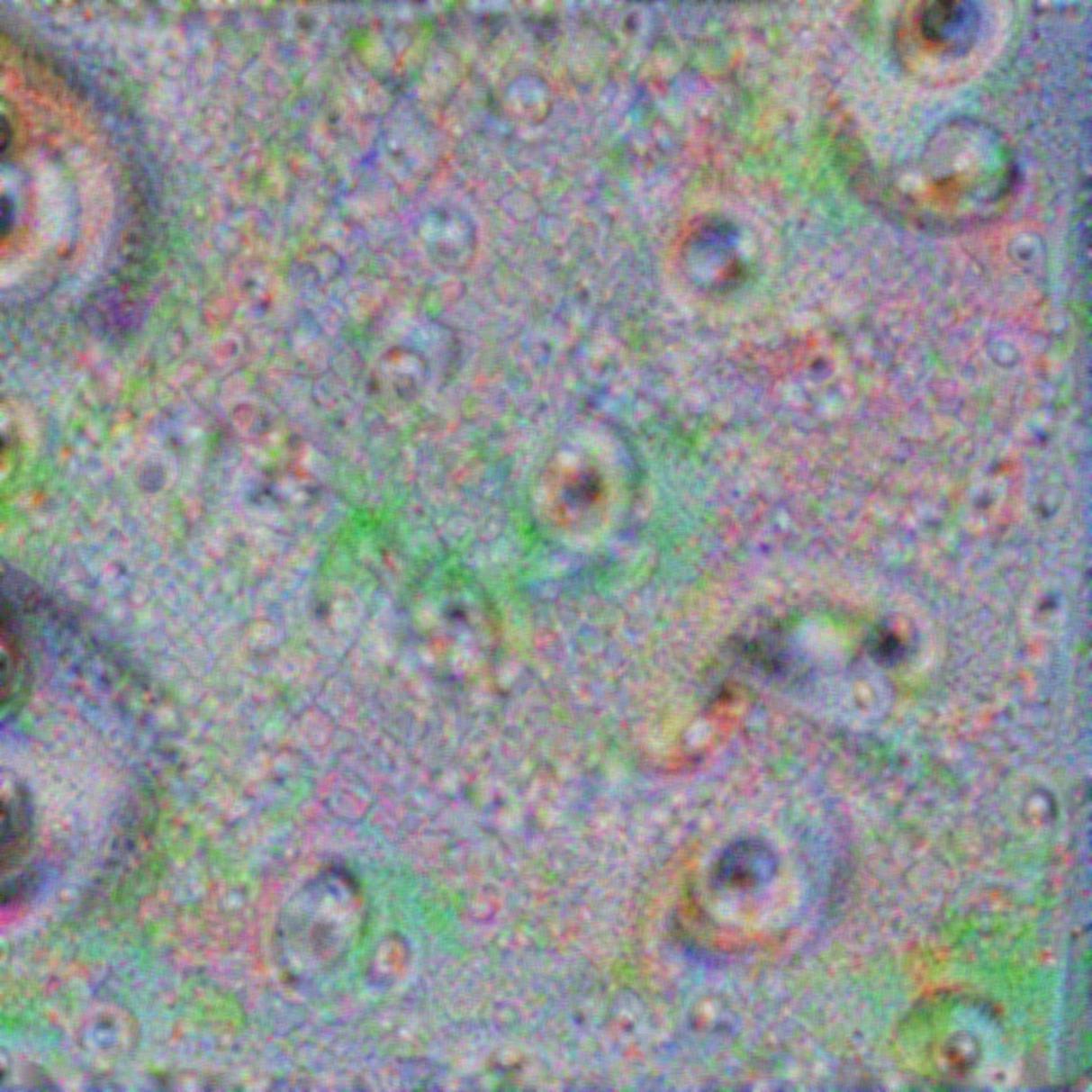} &
         \includegraphics[height=2.3cm]{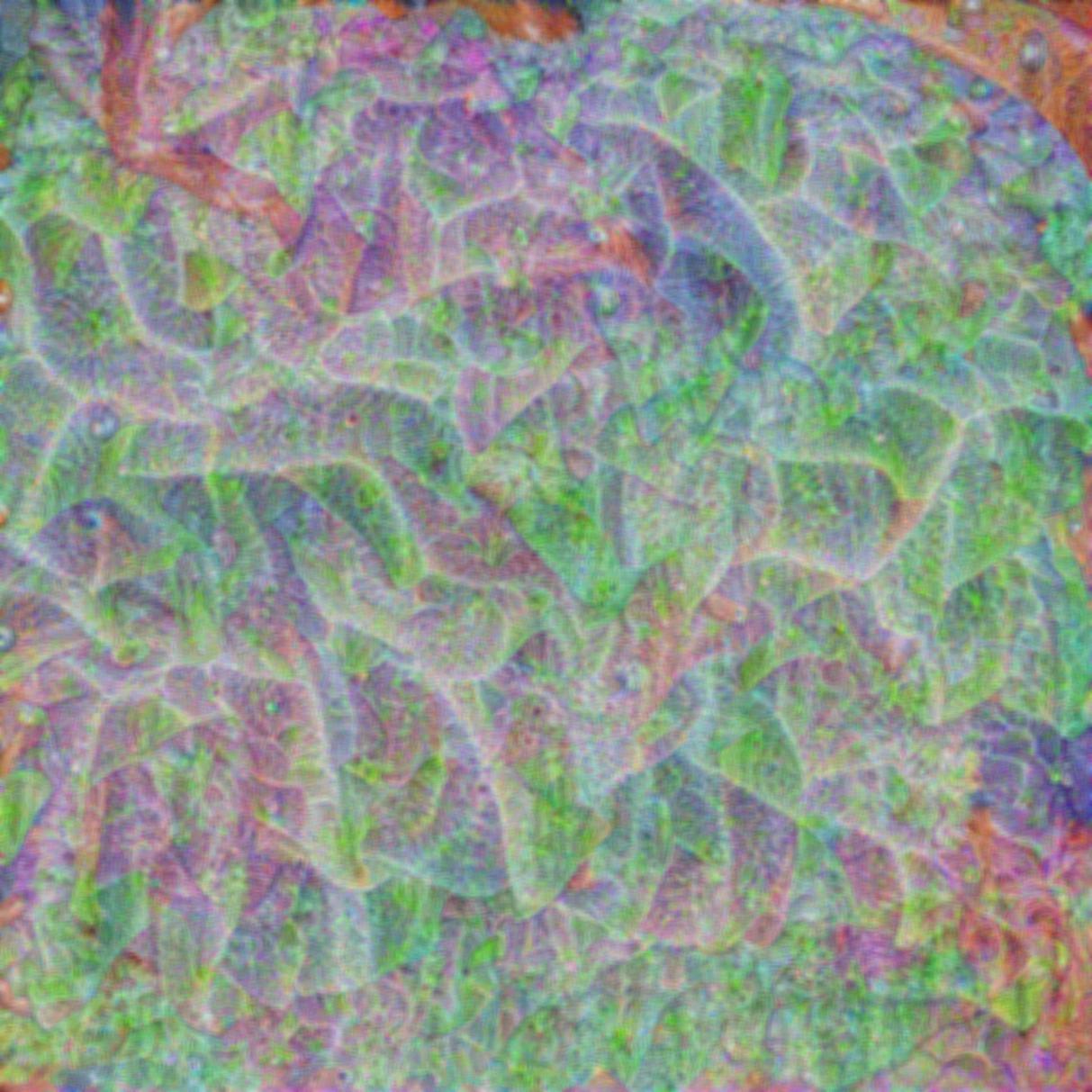} &
         \includegraphics[height=2.3cm]{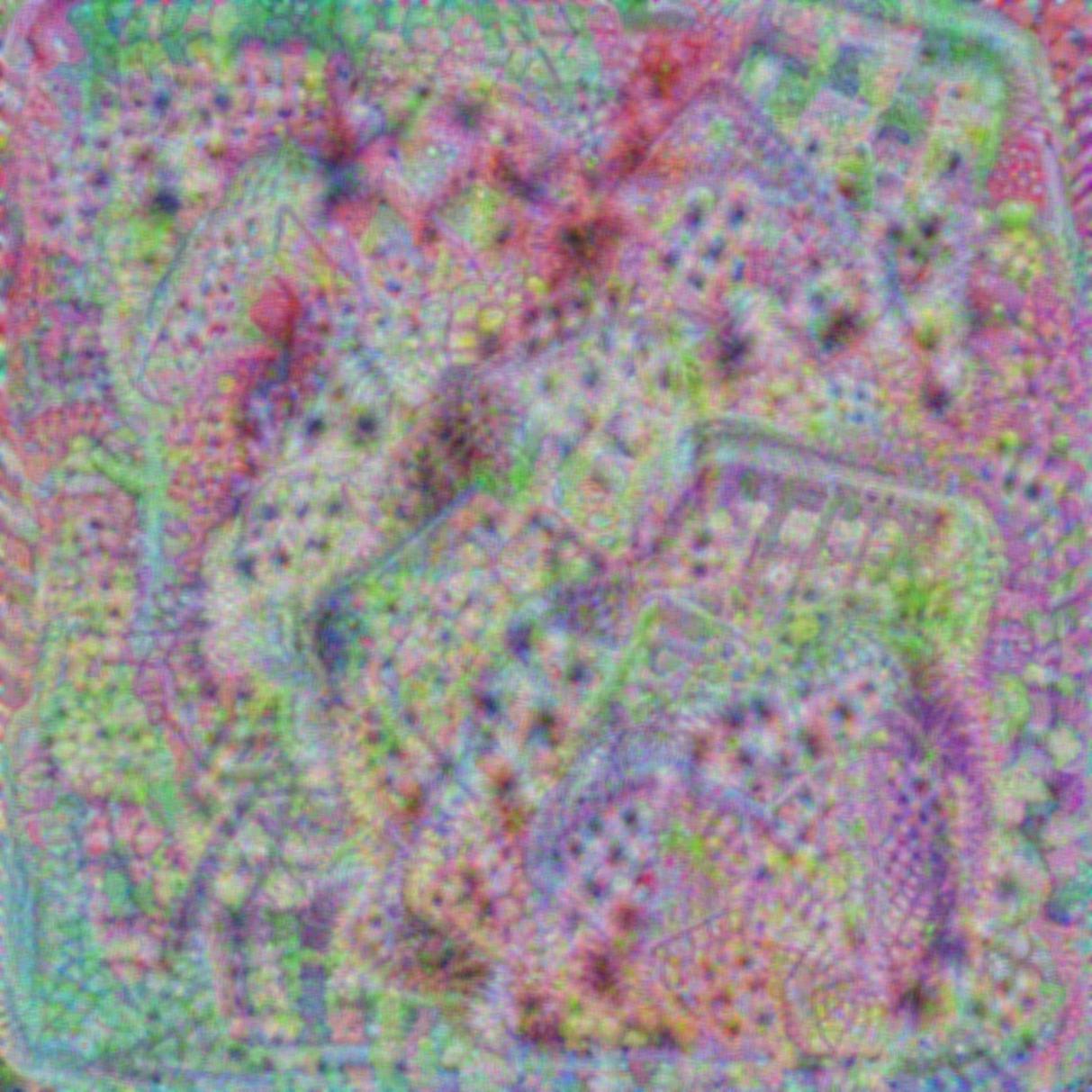} &
         \includegraphics[height=2.3cm]{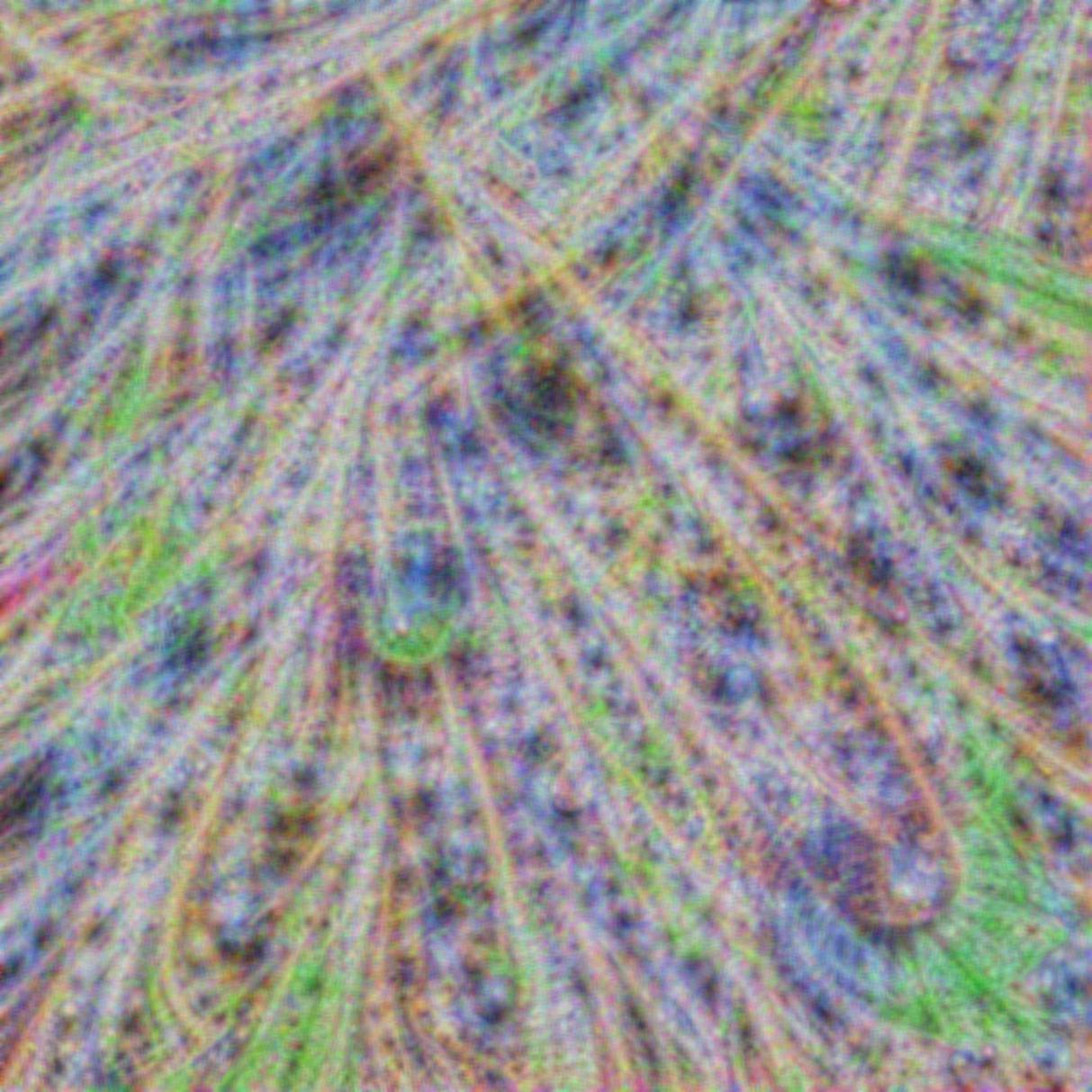} \\
    \end{array}
    \]
    \caption{Feature visualizations from varying layers of ResNet-152. For the fourth row, the first image is from layer 96, the middle 3 are from layer 105, and the fifth image is from layer 126.}
    \label{fig:resnetfeats}
\end{figure}

After running the activation optimization method on the ResNet component of ResMem (fig. \ref{fig:resnetfeats}), we visualized examples of features that drive some of the tensors in the different layers. In earlier layers, we observe more basic features, like edges and orientations (layer 13). In the middle-to-early layers, we see curves and lines, branching patterns, and embossments, as opposed to the earlier layers that only show shapes arranged in
lines (Layer 35).
In the middle-to-late layers, we see more
obvious objects, like faces, eyes, and architectural features. Finally,
the later layers filter for more complicated higher level conceptual features that are difficult to interpret; we begin to see ``part'' filters \citep{olahFeatureVisualization2017}. We can see
what appear to be facial features in the image from layer 96, geometric structures, archways, and architectural glass
ceilings in the images from layer 105. We can see eye-like patterns in the image from layer 126. The repeated convolutions
can make the features more difficult to optimize for, so by the 139th layer, these images become more difficult to produce. In
addition, the high level conceptual features are more complex, and thus
are more difficult to represent in a single image (layer 139).

These feature visualizations help us understand what shapes a filter is selecting for,
but it does not describe what those shapes are.
Instead of optimizing filter activation over the entire space of images, we can also optimize filter activation over just the images in our dataset.
Then, if we compare our feature visualizations with the maximally activating images (fig. \ref{fig:maxact}), we can more easily identify these features. 
We can see that the earlier layers (a-b) are maximally activated by images that are dominated by lines and simple shapes, and the later layers (c-d) are maximally activated by more complex forms, like the lower half of human faces, and the branching arch-like patterns that appear in plants and architecture.

\begin{pandoccrossrefsubfigures}

  \subfloat[]{
    \includegraphics[width=\textwidth,height=2.75cm]{layer_5-0-2_filter_8.jpg}
    \includegraphics[width=\textwidth,height=2.75cm]{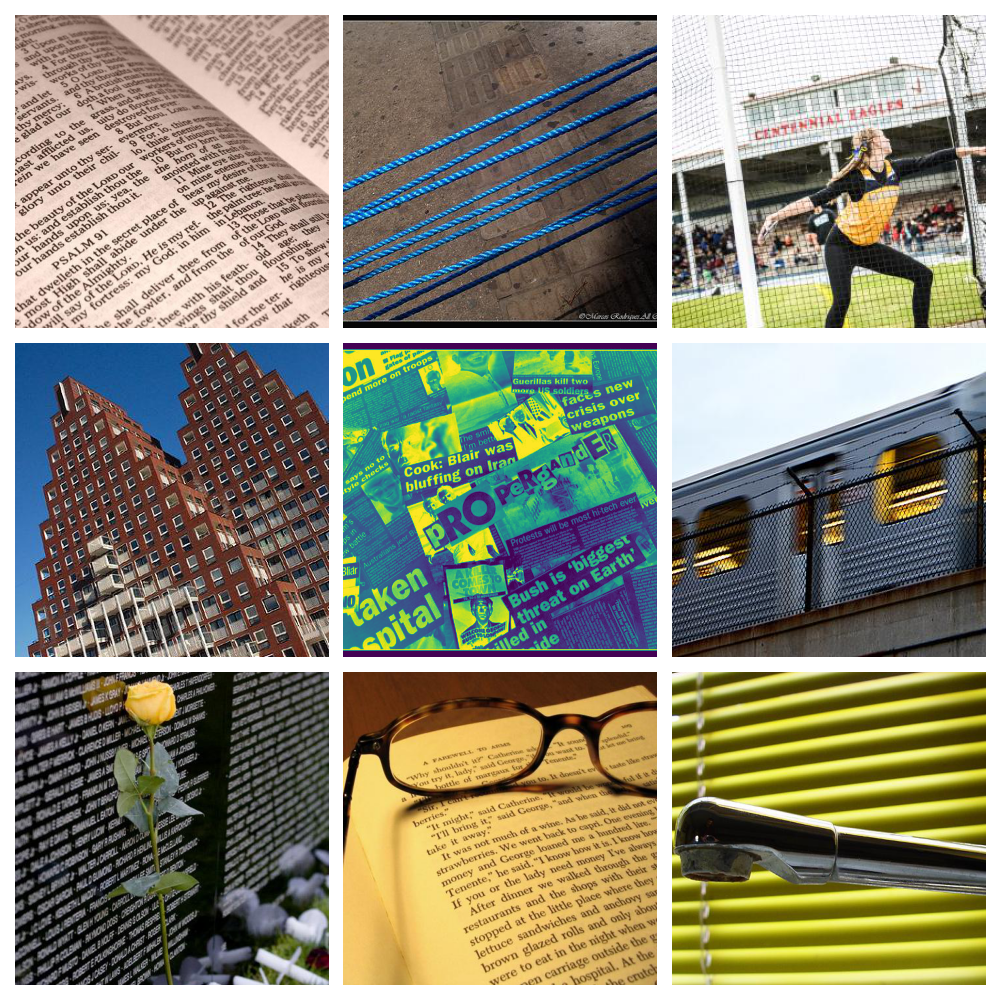}
    }
  \subfloat[]{
    \includegraphics[width=\textwidth,height=2.75cm]{layer_6-0-0_filter_24.jpg}
    \includegraphics[width=\textwidth,height=2.75cm]{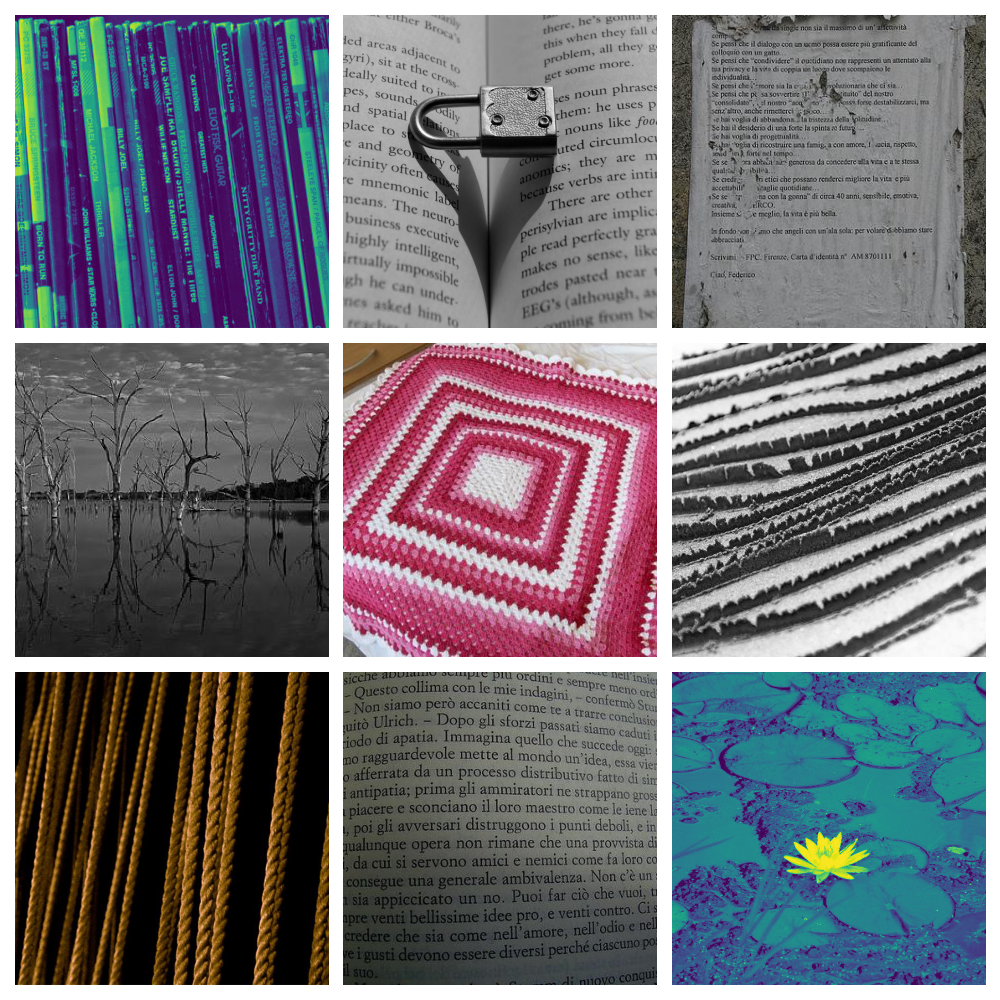}
    }

  \subfloat[]{
      \includegraphics[width=\textwidth,height=2.75cm]{layer_6-20-0_filter_7.jpg}
      \includegraphics[width=\textwidth,height=2.75cm]{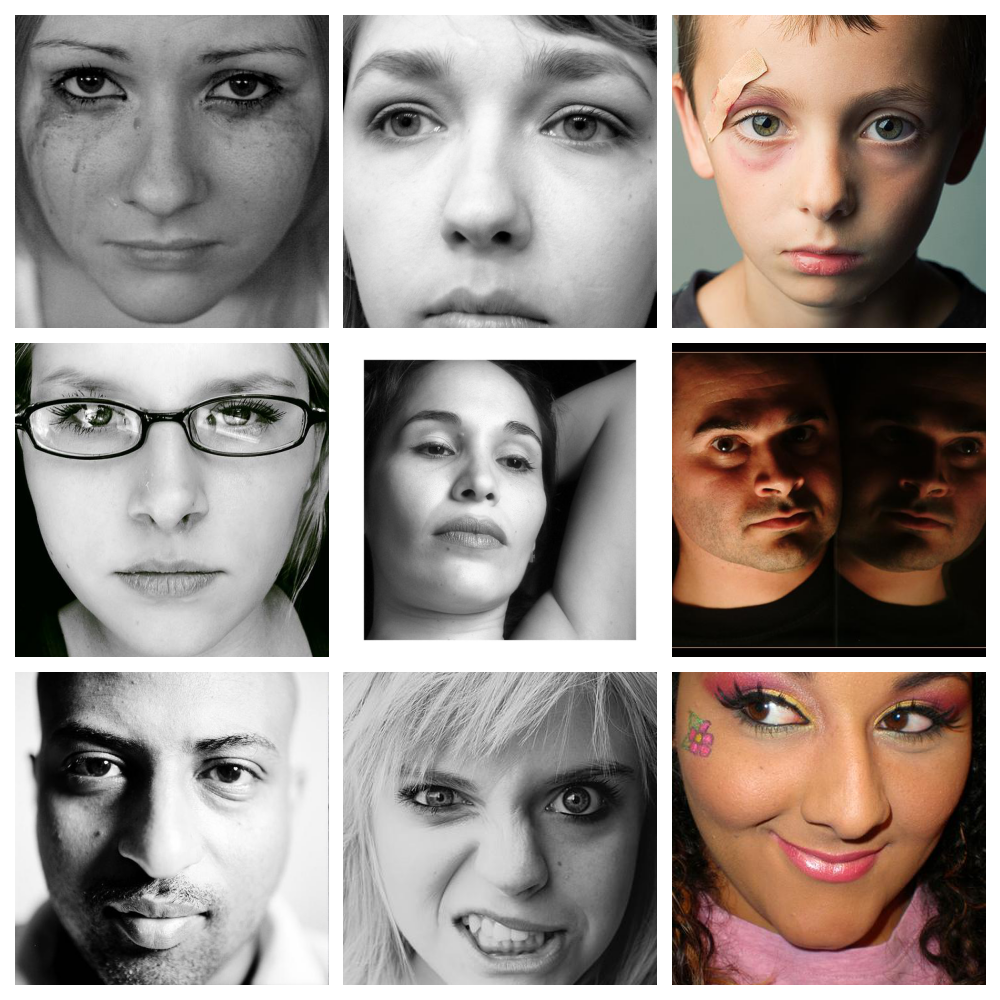}
    }
  \subfloat[]{
    \includegraphics[width=\textwidth,height=2.75cm]{layer_6-23-2_filter_79.jpg}
    \includegraphics[width=\textwidth,height=2.75cm]{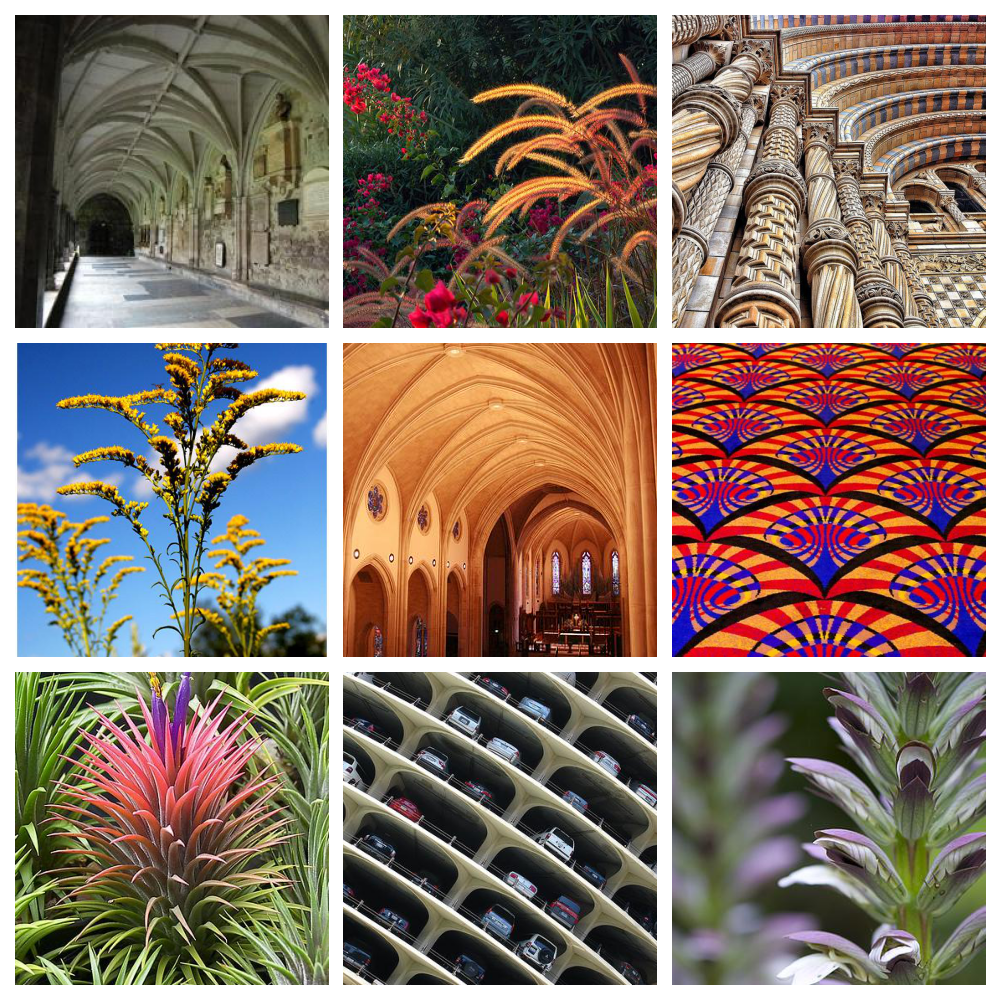}
    }

\caption[]{Feature Visualizations compared to the images that maximally activate the filter.
This analysis helps us understand for what each filter selects.
These examples show this analysis as done on both early layers (a-b), and late layers (c-d).
}
\label{fig:maxact}

\end{pandoccrossrefsubfigures}

\begin{figure}
    \centering
    \[\arraycolsep=2pt\def\arraystretch{2.2}
    \newcolumntype{K}{>{\centering\arraybackslash} m{3cm} }
    \begin{array}{K K K}
    \includegraphics[width=\textwidth,height=3cm]{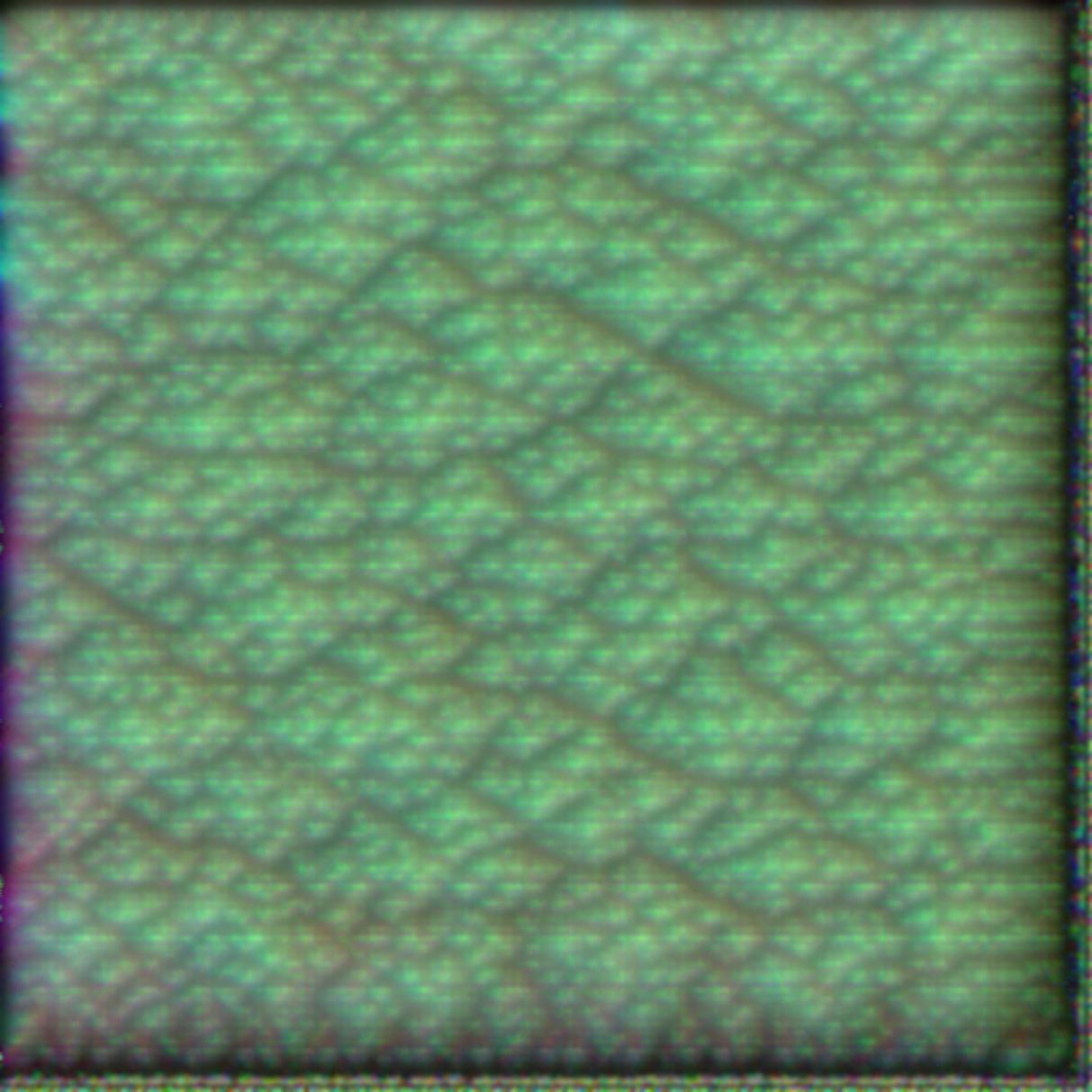} &
    \includegraphics[width=\textwidth,height=3cm]{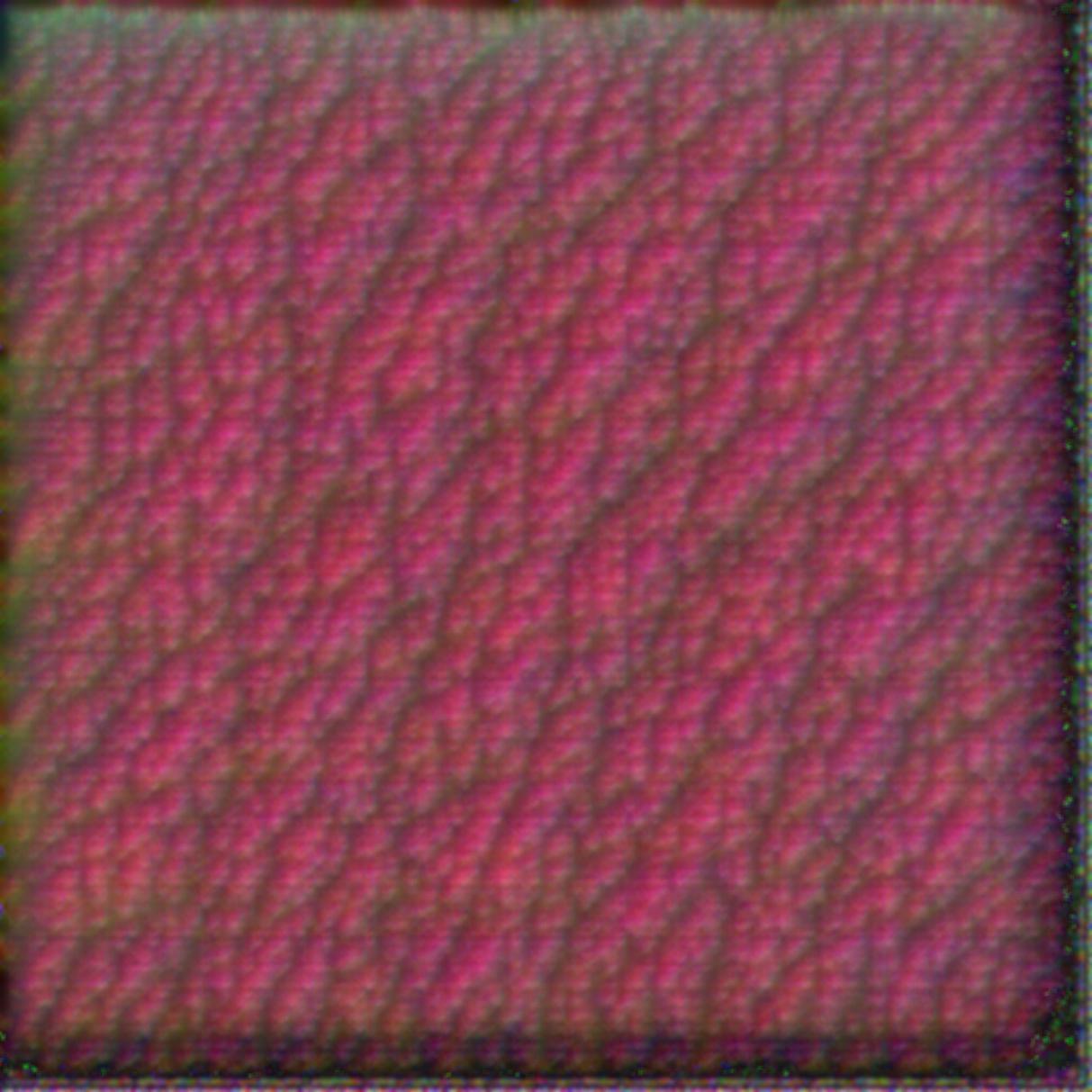} &
    \includegraphics[width=\textwidth,height=3cm]{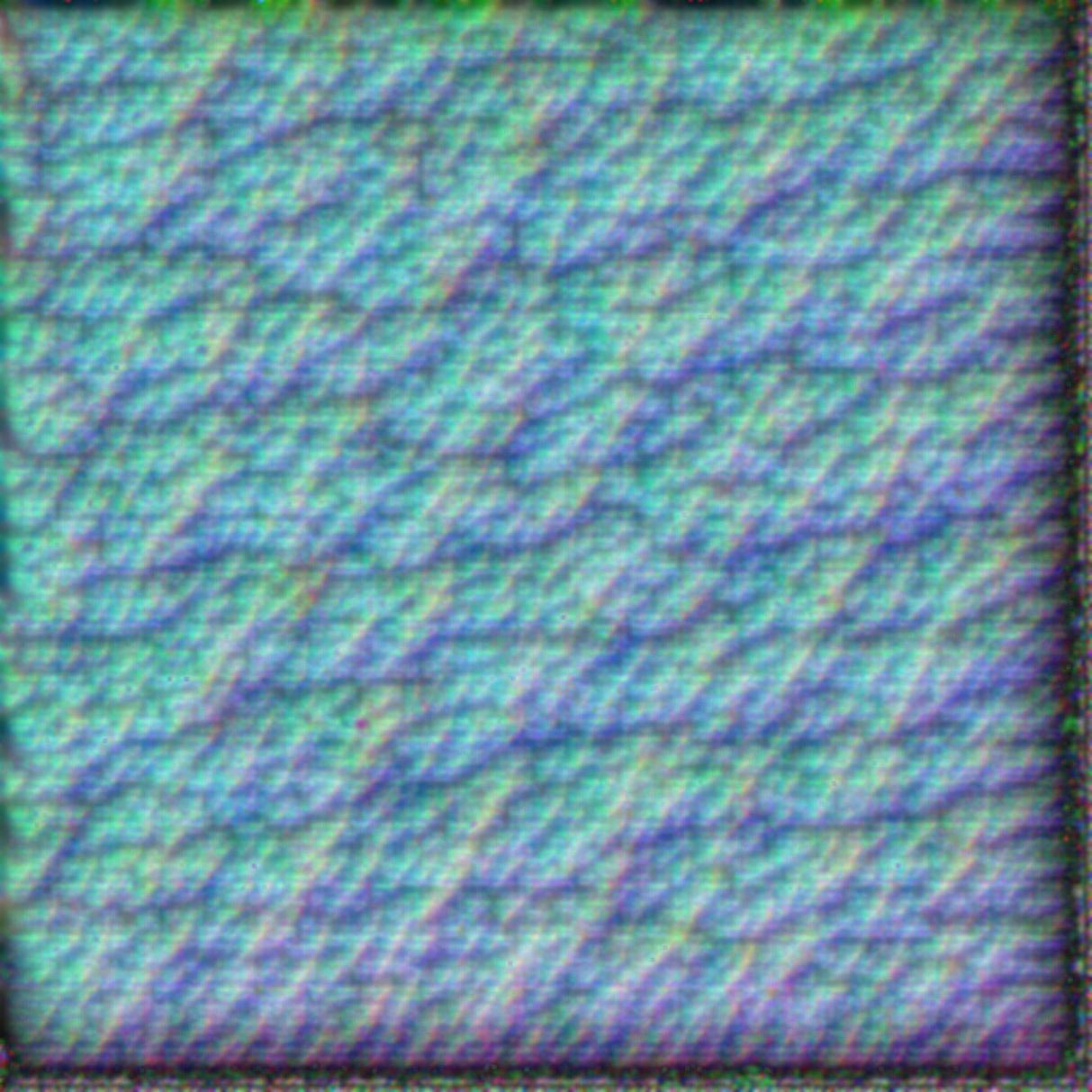} \\
    \includegraphics[width=\textwidth,height=3cm]{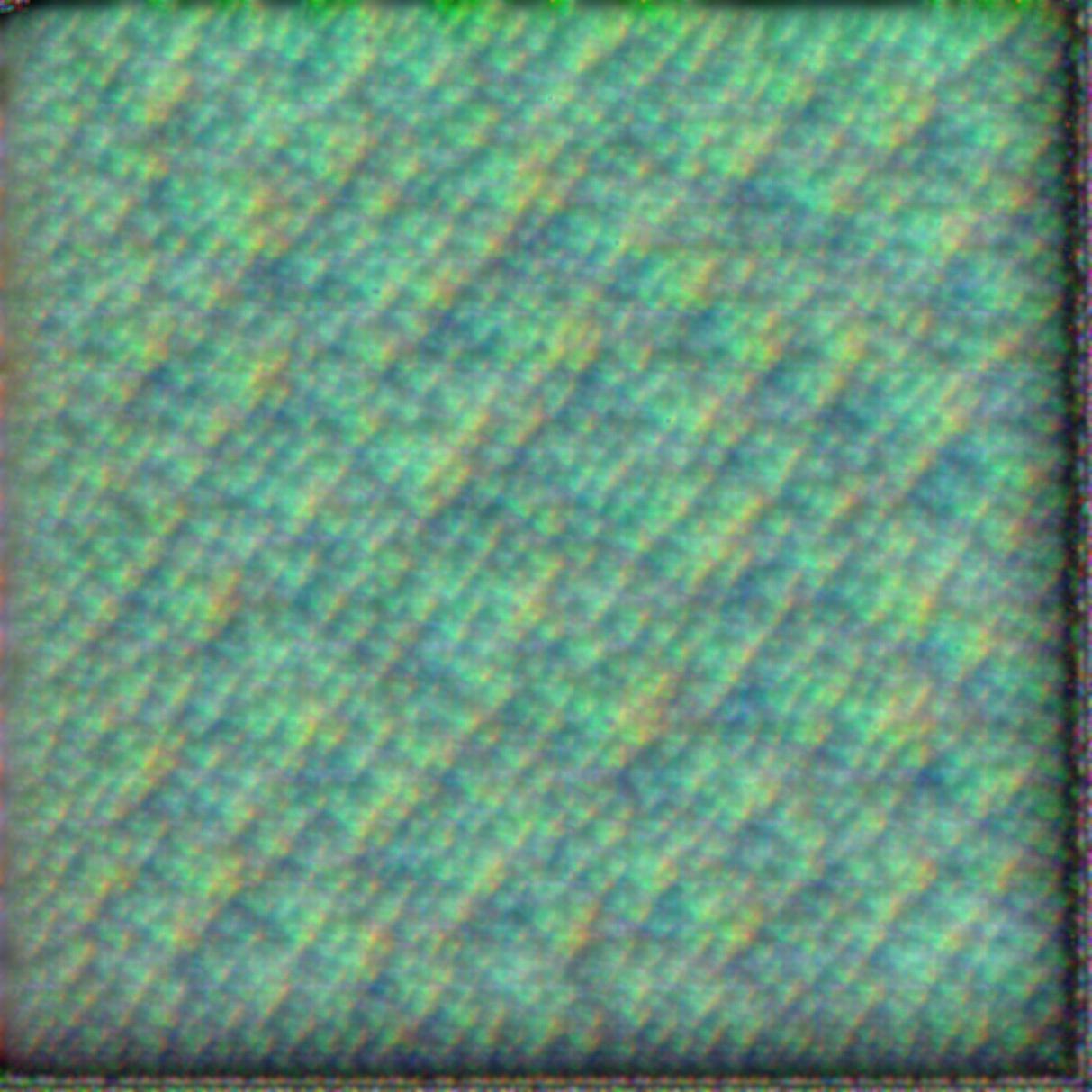} &
    \includegraphics[width=\textwidth,height=3cm]{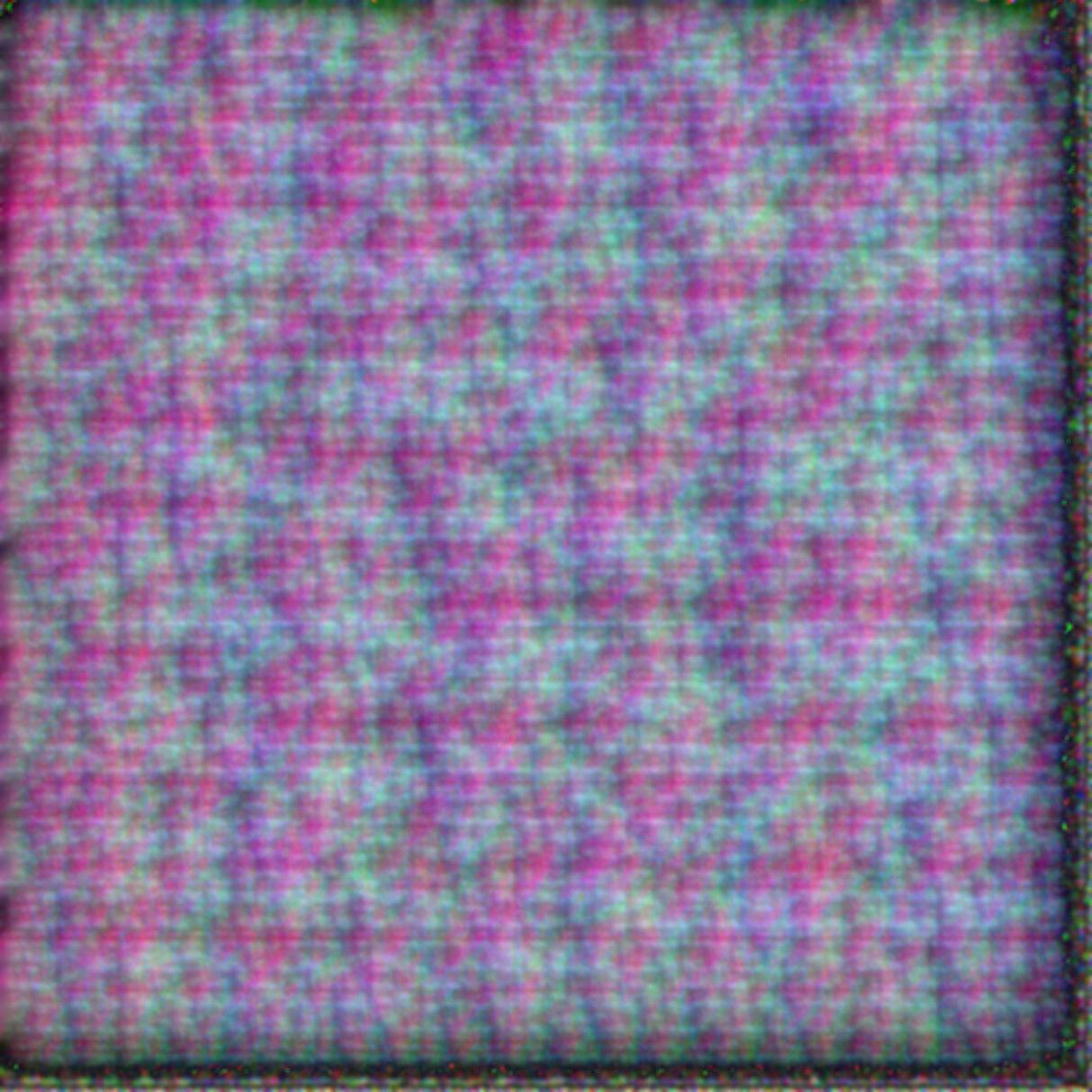} &
    \includegraphics[width=\textwidth,height=3cm]{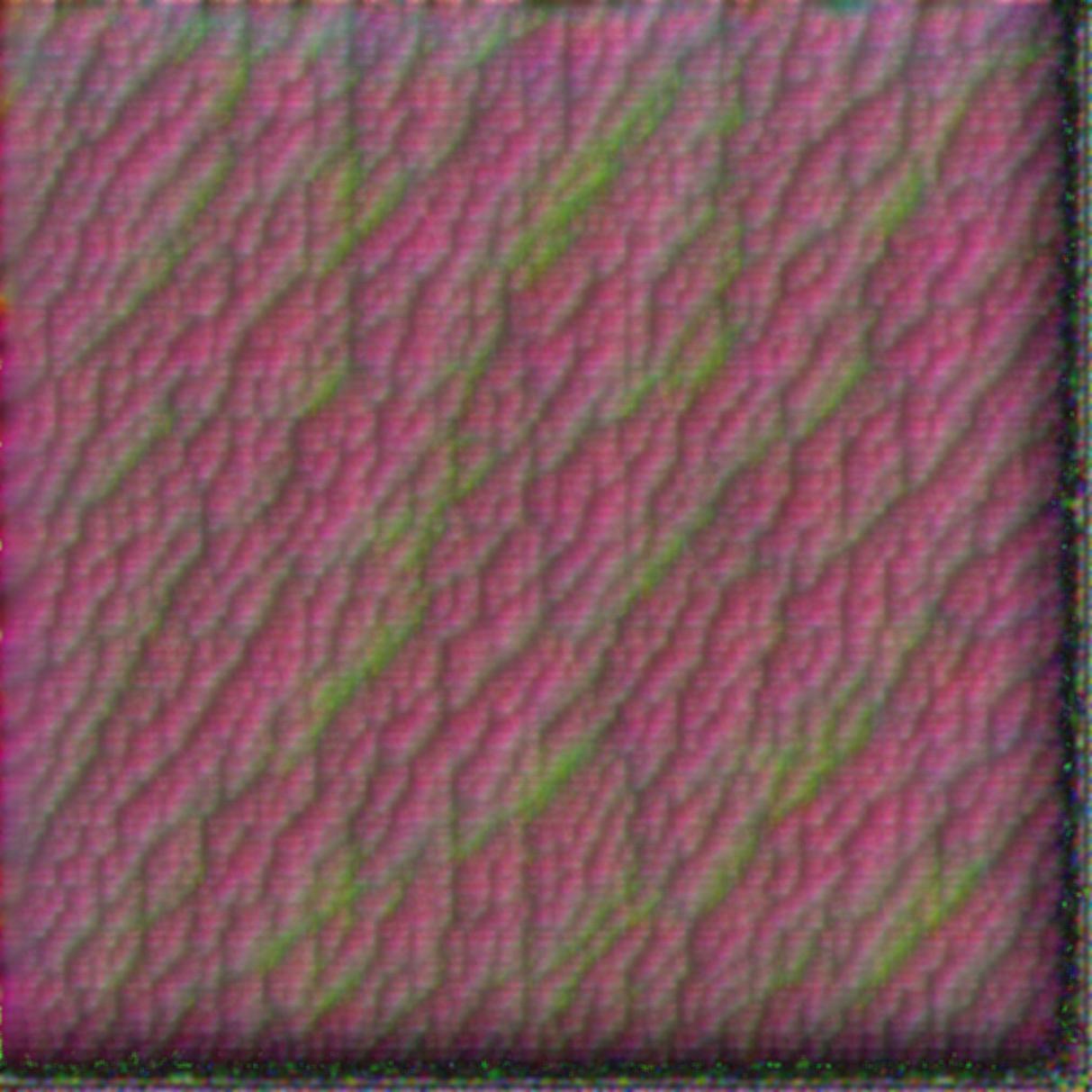} \\
    \includegraphics[width=\textwidth,height=3cm]{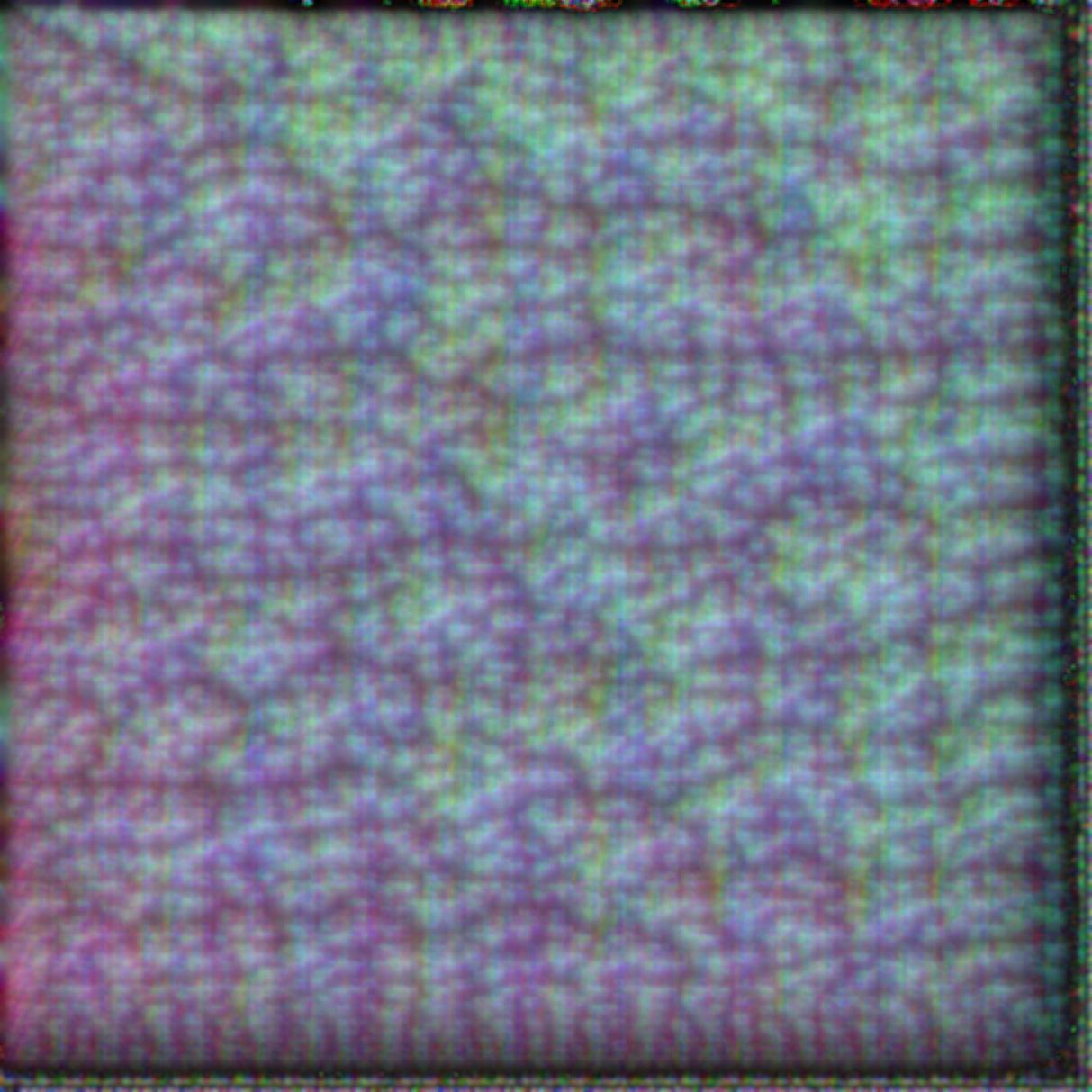} &
    \includegraphics[width=\textwidth,height=3cm]{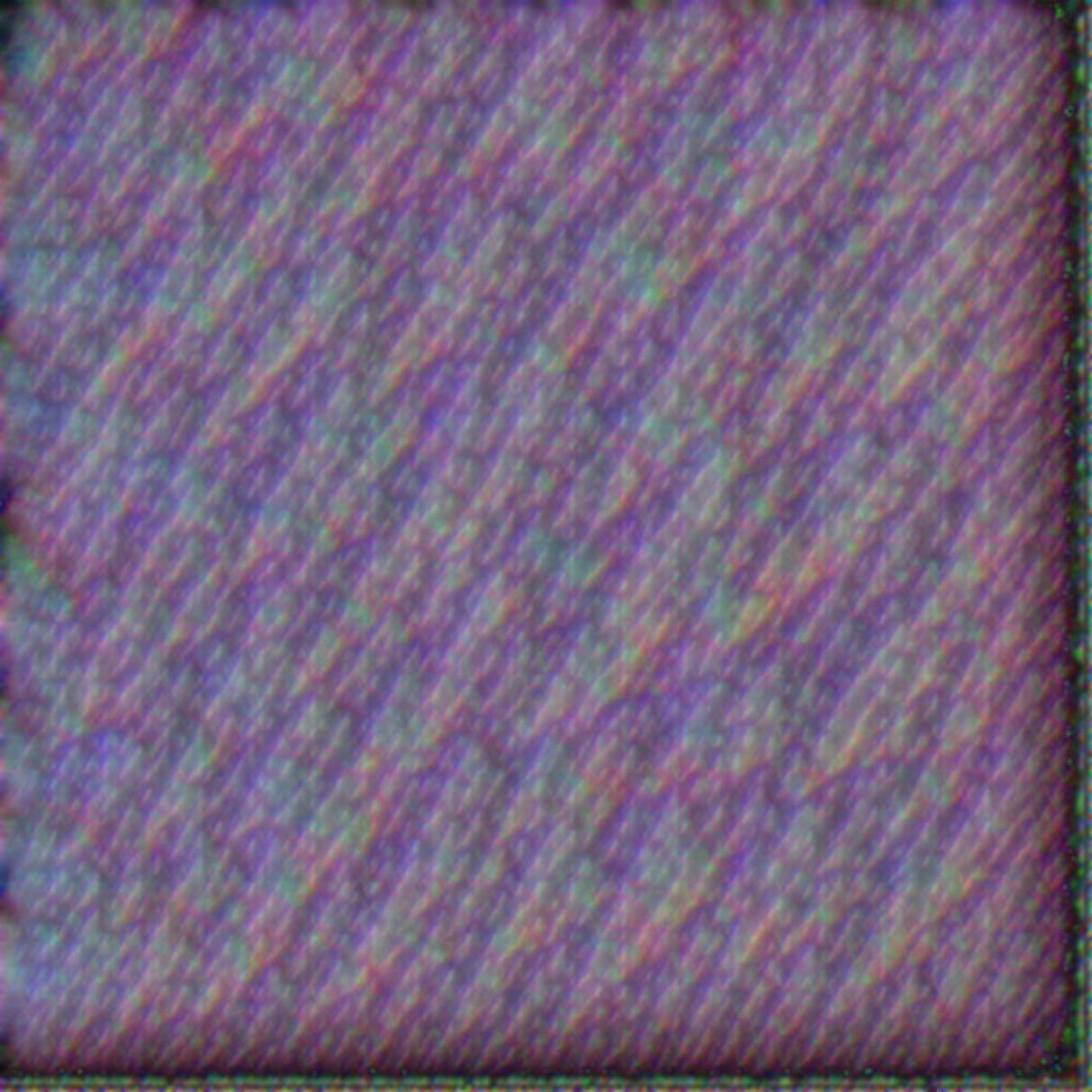} &
    \includegraphics[width=\textwidth,height=3cm]{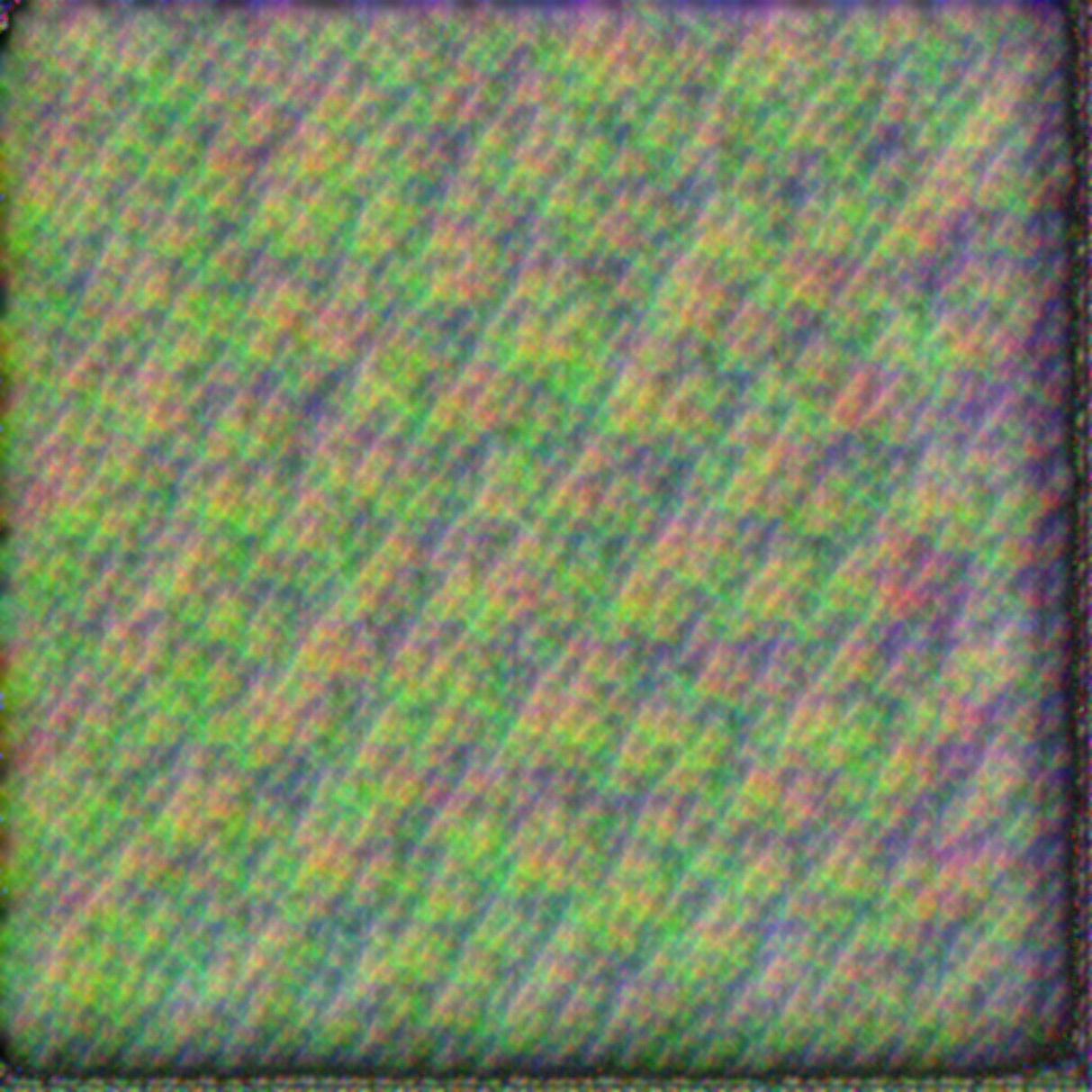} \\
    
    \end{array}
    \]
    \caption{AlexNet feature visualizations. Each of these features is taken from layer 4 out of 5. The features are all very texture-driven, perceptual features.}
    \label{fig:alexfeat}
\end{figure}

Finally, we can run the feature visualization analysis on the AlexNet features
(fig.~\ref{fig:alexfeat}). All of the images are taken from the fourth out
of five convolutional layers. While, in general, this technique is more
difficult to perform on convolutional networks without residual
connections, the images are clear enough to see what is occurring in each
feature. Some filters select for orientation and edge
information, and all of them show
evidence of selecting for textural information.

Even with the limitations on examining AlexNet features, when comparing
them to the ResNet features there is considerable evidence that AlexNet
is filtering for low level perceptual features, like textures and edges
and in some cases simple patterns, while the ResNet is able to filter for more
conceptual objects.

There are technical limitations to analyzing the ResNet features any further.
Convolutional neural networks are considered "black box models", meaning that it can be difficult to understand why a prediction is made in a particular case.
While great progress in analyzing deep neural networks has been made by the community, it is still difficult to compile human-readable descriptions of the inference process. 
As such, it is difficult to know which features of the model and aspects of the images are associated with memorability. However, we make these computed network features publicly available on the Open Science Framework (OSF, \href{https://osf.io/qf5ry/}{https://osf.io/qf5ry/}) to allow further investigations of these features that drive predictions of memorability.

\hypertarget{data-sharing}{%
\section{Data Sharing}\label{data-sharing}}

One impetus for this project is to make a network that is easily usable by psychologists, regardless of familiarity with programming or deep learning. Furthermore, the prior leading model, MemNet, is no longer available, with full code and methodological information inaccessible. Making our model easily available allows researchers across the vision and memory communities to control and select stimuli for experiments based on memorability. Further, this model may serve as a jumping point for developing more biologically inspired models of the visual memory system. We have taken several steps in order to promote the spread of ResMem.

First, ResMem is available as a Python package that can be easily installed using the command:

\begin{verbatim}
pip install resmem
\end{verbatim}

It requires the standard Python packages Numpy \citep{harris2020array},
Torch \citep{teamTorchTensorsDynamic}, and
Torchvision \citep{teamTorchvisionImageVideo} to run, which \texttt{pip}
will download automatically, and it requires minimal Python experience to
operate. Numpy is considered a standard package for working with data of any kind, and Torch and Torchvision are the most commonly used packages for deep learning research. Second, we have created a web application at
The Brain Bridge Lab Website (\href{https://brainbridgelab.uchicago.edu/resmem/}{https://brainbridgelab.uchicago.edu/resmem/}), where anyone can upload an image and quickly get back its estimated
memorability score without installing any software packages. Third, the code for the network as well as the code used to create the network feature visualizations is available on GitHub at \href{https://github.com/Brain-Bridge-Lab/resmem-analysis}{https://github.com/Brain-Bridge-Lab/resmem-analysis}, and a large number of pre-calculated visualizations are available on OSF at \href{https://osf.io/qf5ry/}{https://osf.io/qf5ry/}. This should allow researchers to view the specific architecture, weights, and biases in the network, while also allowing for explorations of how different features emerge from various network layers.

\subsection{Durability}

In the interest of making these resources available to the community as long as possible, we have taken measures to ensure that ResMem will be accessible even if one way to access it stops working.
The model consists of two files, a python file which encodes the architecture of the model, and a PyTorch file, which contains all of the weights and biases needed to make inference using the model.
One advantage of using PyTorch is that the PyTorch file is human-readable, unlike the weights and biases for a Caffe model, which is stored in a binary file.
This means that even when PyTorch is eventually outmoded, it will be easier to port the model to a new framework.
In addition, while the files that define the architecture for a Caffe model are human-readable, a number of parameters are defined implicitly, so it can be difficult to fully understand the architecture without advanced deep learning experience.
PyTorch largely follows the philosophy: ``explicit is better than implicit'', and as such PyTorch modules are easier for non-deep-learning experts to understand.

In the process of development, we used GitHub to track changes to the package, and thus the most up-to-date version of both files is stored at \href{https://github.com/Brain-Bridge-Lab/resmem}{https://github.com/Brain-Bridge-Lab/resmem}.
For easy installation and distribution, ResMem is also stored on PyPI, the official Python Package Index. 
Normally they do not distribute pre-trained models, but we have received special dispensation to use PyPI for ResMem, and thus, the package can be easily installed with one command.

For additional resilience, we have also uploaded all of the files needed to build ResMem to an OSF repository at \href{https://osf.io/qf5ry/}{https://osf.io/qf5ry/}. 
These three file distribution services (GitHub, PyPI, and OSF) are considered integral to the programming, python, and scientific communities respectively, and by distributing ResMem on all three, we expect it to be available to researchers until well after it is outmoded by future research. We have also included a step-by-step tutorial on how to start using ResMem in the Supplemental Information.

\subsection{Validations of ResMem}

One concern of the prior model MemNet was that the model could be overfit on the distinctive image set used (LaMem). There was also the concern that the broad range of images represented in LaMem prevented MemNet from making successful within-category predictions. In order to test the generalizability of ResMem, we successfully validated its predictions with human memory performance recorded from three other studies using entirely different image sets in different experimental paradigms.

First, we compared ResMem predictions with memory performance for black and white scene images by individuals at different stages of Alzheimer's dementia \citep{bainbridgeMemorabilityPhotographsSubjective2019}. In contrast with the continuous recognition tasks used in the training sets for ResMem (LaMem and MemCat), here, participants performed a perceptual indoor/outdoor scene categorization task, and then after a 70 minute delay, had to indicate memory with old/new judgments for the images intermixed with matched foils. ResMem was significantly able to predict memory performance for all participant groups, including healthy controls (Spearman rank correlation: $\rho = 0.30,\; p = \num{2.82e-19}$), those with a subjective cognitive decline ($\rho = 0.25,\; p = \num{2.63e-13}$), and those with a diagnosed mild cognitive impairment, MCI ($\rho = 0.14,\; p = \num{4.03e-5}$). ResMem outperforms the model reported in the original paper, which was unable to predict memory for those with MCI \citep{bainbridgeMemorabilityPhotographsSubjective2019}. ResMem's success has tremendous implications for its use in making predictions about memory in patient groups; for example, ResMem could be used to design highly memorable environments for those developing dementia.

Second, we tested ResMem with a highly constrained image set, specifically the Food Folio dataset \citep{lloydEatingDisorder}, a set of overhead photographs of food items on a white plate against a black background. ResMem predictions were significantly correlated with human continuous recognition task performance, $\rho = 0.24,\; p = 0.004$ \citep{XinyueFoods}. This indicates that ResMem's predictions are even successful for images that are highly visually similar and within the same semantic category (food). This also shows how ResMem could be used to select items within a specific set for real-world applications--for example, choosing the most memorable food item to use in a nutritional poster.

Finally, we tested whether ResMem would replicate human memory performance scores made publicly available in prior work from another research group \citep{dubeyWhatMakesObject}. In the original study, Dubey and colleagues (2015) studied the relationship of the memorability of objects to scenes containing those objects. ResMem was significantly able to predict memory performance for the scene images ($\rho = 0.36,\; p \approx 0$). Importantly, ResMem could also predict memory for cropped images of single objects within the scenes ($\rho = 0.14,\; p = \num{8.34e-14}$). ResMem thus is able to successfully replicate prior work, and could be used to test key questions about the relationships of memory across different stimulus categories (e.g., scenes and objects).

We envision more validations of ResMem will emerge over time, and have created a living resource of validations of the network (\href{https://brainbridgelab.uchicago.edu/resmem/resmem-validation/}{https://brainbridgelab.uchicago.edu/resmem/resmem-validation/}). This will allow researchers to see which types of images they can use with ResMem to predict memorability (and which are not successfully predictable).

\hypertarget{discussion}{%
\section{Discussion}\label{discussion}}

\subsection{Deep Learning}

Using modern machine learning techniques can vastly improve our ability
to estimate memorability, and it also opens up new avenues of analysis.
In addition to making memorability analysis simpler, models like ResMem
can be leveraged to do analysis of high-level visual features, and how
those features relate to memorability. As hardware technology continues
to progress, more complicated models, such as graph-based neural
networks, M3M without the deconvolution layers, and model-based
algorithms can be employed to help understand memorability in those
contexts. 

In addition to showing the ability of modern techniques to estimate memorability, we have also shown that computational psychologists need not be constrained to off-the-shelf general-purpose architectures like AlexNet or even ResNet.
By constructing a model specifically with memorability in mind, and taking into account known features of memorability, we have shown that when domain knowledge is incorporated into the architecture design, models will perform better than their general-purpose counterparts.

\subsection{Psychology}

An important goal of this research is to make ResMem available to any scientist who wants to study
what makes something memorable. 
This model will allow researchers to easily and inexpensively control for memorability in other studies. In many cases, it is helpful to know in advance what participants will remember or forget; researchers can select stimuli that are highly powered for driving differences in memory, or stimuli that are well-controlled for memorability effects. Researchers can also analyze data post-hoc to see if the intrinsic memorability of their images accounted for any of their effects. In fact, research has shown that sensitivity to stimulus memorability during the perception of an image can account for some patterns in the brain previously attributed to memory encoding \citep{bainbridgeDissociatingNeuralMarkers2018}. It is also possible that other work has examined stimulus category effects that could be confounded with memorability; for example, tests of face familiarity comparing celebrities versus non-celebrities could instead be examining memorable versus non-memorable faces, rather than prior visual experience. Thus, an easy method for quantifying the memorability of stimuli should be an indispensable resource for psychologists and neuroscientists across different fields.

More broadly, this model can serve as a resource for the general public -- allowing educators to identify memorable materials, helping the average person choose a memorable photo to post online, and aiding clinicians in creating memorable environments for those experiencing memory loss \citep{bainbridgeMemorabilityPhotographsSubjective2019}. In validations of ResMem, we found that it could significantly predict memory for patient groups experiencing the onset of dementia. Future work could use ResMem to identify the images that are most predictive of a developing deficit. We also found in our validations that ResMem can successfully predict memory for highly specific image sets, such as food images. ResMem would thus likely be able to identify which image in a set of similar images would best drive memory, for materials such as textbooks or infographics. Because ResMem is easy to use, it can be easily adapted for a wide range of these applications.
To support its easy use, ResMem is downloadable through the python package
\texttt{resmem}, and a demonstration is available on the Brain Bridge
Lab website (\href{https://brainbridgelab.uchicago.edu/resmem}{https://brainbridgelab.uchicago.edu/resmem}).

Importantly, the features that emerge from ResMem give an initial glimpse into the properties that make an image memorable. Thus far, it has been an open mystery of what features drive the memorability of an image. A large body of prior work has shown that memorability is not synonymous with other low-level or high-level image features, such as color, edges, aesthetics, interest, or saliency \citep{isolaWhatMakesPhotograph2014, bainbridgeMemorabilityStimulusdrivenPerceptual2017, Bainbridge_2017}. In fact, there is even no correlation between what people think they will remember and what is in fact memorable \citep{isolaWhatMakesPhotograph2014}. The current study has begun to give a glimpse into the features that drive memorability: showing that including high-level conceptual features in
the memorability estimation process improves the accuracy of those
estimations considerably. In comparison with MemNet which only
decomposes images into perceptual features, ResMem achieves better
predictive power in terms of generalizability, rank correlation, and
estimation accuracy. This, when taken in concert with previous research,
implies that memorability is related intimately to both the perceptual
and conceptual features of an image, and recent converging evidence in the psychological domain has also shown a dominance of semantic information in determining the memorability of an image \citep{kramerCharacterizingMemorabilityRepresentational2021}. Importantly, the fact that ResMem can successfully predict the memorability of images demonstrates that it is an image-computable feature that may be automatically processed during perception. In fact, image classification DNNs such as AlexNet show strong correspondences in representation with the human visual system \citep{cichyComparisonDeepNeural2016, Yamins_Hong_Cadieu_Solomon_Seibert_DiCarlo_2014}, and dissimilarity in early layers and similarity in late layers has shown correspondences with scene memorability \citep{Koch834796}. The current network may better capture  higher-level visual and semantic processes; even without a "memory" built into the model, ResMem is able to predict memory performance using the processing of visual features alone. As a whole, this suggests that a visual system can be sensitive to what we are likely to remember, even in the absence of top-down processes or processes of memory encoding and retrieval. Such findings could support current hypotheses that memorability reflects how visual information is prioritized for later memory, as a type of sorting metric for perceptual inputs \citep{xieMemorabilityWordsArbitrary2020}. 
Future work can examine the relationship between this model and the visual and memory systems in the brain, as prior work has done for visual areas \citep{cichyComparisonDeepNeural2016,jaeglePopulationResponseMagnitude2019}.

Carefully applying deep learning in the memorability field can
help us better understand the phenomenon for years to come.

\section{Acknowledgements}

We would like to acknowledge Lore Goetschalckx for providing us with details on MemNet's implementation.
We would also like to acknowledge Deepasri Prasad and Max Kramer for information sharing and general feedback.

\section{Declarations}

\subsection{Funding}

Not applicable

\subsection{Conflicts of Interest}

Not applicable

\subsection{Availability of data and material}

The model is available from the python packaging authority (\textcolor{BBBlue}{ \href{https://pypi.org/project/resmem/}{https://pypi.org/project/resmem/}}), and an online demo is available on the Brain Bridge Lab website ({ \color{BBBlue} \underline{\href{https://brainbridgelab.uchicago.edu/resmem}{https://brainbridgelab.uchicago.edu/resmem}}}).
Miscellaneous data, including feature analyses, prediction performance within all subcategories of MemCat, and an archival copy of the pre-trained model are hosted on OSF at ({ \color{BBBlue} \underline{\href{https://osf.io/qf5ry/}{https://osf.io/qf5ry/}}}).
The data used to train ResMem came from two sources. LaMem is hosted by MIT ({\color{BBBlue} \underline{\href{http://memorability.csail.mit.edu/download.html}{http://memorability.csail.mit.edu/download.html}}}). MemCat is hosted by the Flemish government ({\color{BBBlue} \underline{\href{https://gestaltrevision.be/projects/memcat/}{https://gestaltrevision.be/projects/memcat/}}}).

\subsection{Code Availability}

The code for the ResMem package as published is hosted on GitHub at ({\color{BBBlue} \underline{\href{https://github.com/Brain-Bridge-Lab/resmem}{https://github.com/Brain-Bridge-Lab/resmem}}}). The code used to generate figures and run analyses is split across two repositories, {\color{BBBlue} \underline{\href{https://github.com/Brain-Bridge-Lab/BrainBridge-MemNet}{https://github.com/Brain-Bridge-Lab/BrainBridge-MemNet}}}) and {\color{BBBlue} \underline{\href{https://github.com/Brain-Bridge-Lab/resmem-analysis}{https://github.com/Brain-Bridge-Lab/resmem-analysis}}}).

\subsection{Author Contribution}

W.A. Bainbridge conceived of the presented idea. C.D. Needell designed, programmed, and tuned the model. C.D. Needell and W.A. Bainbridge wrote the manuscript. W.A. Bainbridge supervised the project.

\pagebreak

\bibliography{resmem}


\begin{thebibliography}{45}
\ifx \bisbn   \undefined \def \bisbn  #1{ISBN #1}\fi
\ifx \binits  \undefined \def \binits#1{#1}\fi
\ifx \bauthor  \undefined \def \bauthor#1{#1}\fi
\ifx \batitle  \undefined \def \batitle#1{#1}\fi
\ifx \bjtitle  \undefined \def \bjtitle#1{#1}\fi
\ifx \bvolume  \undefined \def \bvolume#1{\textbf{#1}}\fi
\ifx \byear  \undefined \def \byear#1{#1}\fi
\ifx \bissue  \undefined \def \bissue#1{#1}\fi
\ifx \bfpage  \undefined \def \bfpage#1{#1}\fi
\ifx \blpage  \undefined \def \blpage #1{#1}\fi
\ifx \burl  \undefined \def \burl#1{\textsf{#1}}\fi
\ifx \doiurl  \undefined \def \doiurl#1{\url{https://doi.org/#1}}\fi
\ifx \betal  \undefined \def \betal{\textit{et al.}}\fi
\ifx \binstitute  \undefined \def \binstitute#1{#1}\fi
\ifx \binstitutionaled  \undefined \def \binstitutionaled#1{#1}\fi
\ifx \bctitle  \undefined \def \bctitle#1{#1}\fi
\ifx \beditor  \undefined \def \beditor#1{#1}\fi
\ifx \bpublisher  \undefined \def \bpublisher#1{#1}\fi
\ifx \bbtitle  \undefined \def \bbtitle#1{#1}\fi
\ifx \bedition  \undefined \def \bedition#1{#1}\fi
\ifx \bseriesno  \undefined \def \bseriesno#1{#1}\fi
\ifx \blocation  \undefined \def \blocation#1{#1}\fi
\ifx \bsertitle  \undefined \def \bsertitle#1{#1}\fi
\ifx \bsnm \undefined \def \bsnm#1{#1}\fi
\ifx \bsuffix \undefined \def \bsuffix#1{#1}\fi
\ifx \bparticle \undefined \def \bparticle#1{#1}\fi
\ifx \barticle \undefined \def \barticle#1{#1}\fi
\bibcommenthead
\ifx \bconfdate \undefined \def \bconfdate #1{#1}\fi
\ifx \botherref \undefined \def \botherref #1{#1}\fi
\ifx \url \undefined \def \url#1{\textsf{#1}}\fi
\ifx \bchapter \undefined \def \bchapter#1{#1}\fi
\ifx \bbook \undefined \def \bbook#1{#1}\fi
\ifx \bcomment \undefined \def \bcomment#1{#1}\fi
\ifx \oauthor \undefined \def \oauthor#1{#1}\fi
\ifx \citeauthoryear \undefined \def \citeauthoryear#1{#1}\fi
\ifx \endbibitem  \undefined \def \endbibitem {}\fi
\ifx \bconflocation  \undefined \def \bconflocation#1{#1}\fi
\ifx \arxivurl  \undefined \def \arxivurl#1{\textsf{#1}}\fi
\csname PreBibitemsHook\endcsname

\bibitem{bainbridgeMemorabilityHowWhat2019}
\begin{bbook}
\bauthor{\bsnm{Bainbridge}, \binits{W.A.}}:
\bbtitle{Memorability: {{How}} What We See Influences What We Remember},
p. \bfpage{27}
(\byear{2019})
\end{bbook}
\endbibitem

\bibitem{IsolaWhatMakes2011}
\begin{botherref}
\oauthor{\bsnm{Isola}, \binits{P.}},
\oauthor{\bsnm{Xiao}, \binits{J.}},
\oauthor{\bsnm{Torralba}, \binits{A.}},
\oauthor{\bsnm{Oliva}, \binits{A.}}:
What makes an image memorable?,
145--152
(2011).
\doiurl{10.1109/CVPR.2011.5995721}
\end{botherref}
\endbibitem

\bibitem{Bainbridge_Isola_Oliva_2013}
\begin{barticle}
\bauthor{\bsnm{Bainbridge}, \binits{W.A.}},
\bauthor{\bsnm{Isola}, \binits{P.}},
\bauthor{\bsnm{Oliva}, \binits{A.}}:
\batitle{The intrinsic memorability of face photographs.}
\bjtitle{Journal of Experimental Psychology: General}
\bvolume{142}(\bissue{4}),
\bfpage{1323}--\blpage{1334}
(\byear{2013}).
\doiurl{10.1037/a0033872}
\end{barticle}
\endbibitem

\bibitem{IsolaParikhTorralbaOliva2011}
\begin{bchapter}
\bauthor{\bsnm{Isola}, \binits{P.}},
\bauthor{\bsnm{Parikh}, \binits{D.}},
\bauthor{\bsnm{Torralba}, \binits{A.}},
\bauthor{\bsnm{Oliva}, \binits{A.}}:
\bctitle{Understanding the intrinsic memorability of images}.
In: \bbtitle{Advances in Neural Information Processing Systems}
(\byear{2011})
\end{bchapter}
\endbibitem

\bibitem{bainbridgeDissociatingNeuralMarkers2018}
\begin{barticle}
\bauthor{\bsnm{Bainbridge}, \binits{W.A.}},
\bauthor{\bsnm{Rissman}, \binits{J.}}:
\batitle{Dissociating neural markers of stimulus memorability and subjective
  recognition during episodic retrieval}.
\bjtitle{Scientific Reports}
\bvolume{8}(\bissue{1}),
\bfpage{8679}
(\byear{2018}).
\doiurl{10.1038/s41598-018-26467-5}
\end{barticle}
\endbibitem

\bibitem{bainbridgeMemorabilityStimulusdrivenPerceptual2017}
\begin{barticle}
\bauthor{\bsnm{Bainbridge}, \binits{W.A.}},
\bauthor{\bsnm{Dilks}, \binits{D.D.}},
\bauthor{\bsnm{Oliva}, \binits{A.}}:
\batitle{Memorability: {{A}} stimulus-driven perceptual neural signature
  distinctive from memory}.
\bjtitle{NeuroImage}
\bvolume{149},
\bfpage{141}--\blpage{152}
(\byear{2017}).
\doiurl{10.1016/j.neuroimage.2017.01.063}
\end{barticle}
\endbibitem

\bibitem{Mohsenzadeh2019}
\begin{barticle}
\bauthor{\bsnm{Mohsenzadeh}, \binits{Y.}},
\bauthor{\bsnm{Mullin}, \binits{C.}},
\bauthor{\bsnm{Oliva}, \binits{A.}},
\bauthor{\bsnm{Pantazis}, \binits{D.}}:
\batitle{The perceptual neural trace of memorable unseen scenes}.
\bjtitle{Scientific Reports}
\bvolume{9}(\bissue{1}),
\bfpage{6033}
(\byear{2019}).
\doiurl{10.1038/s41598-019-42429-x}
\end{barticle}
\endbibitem

\bibitem{jaeglePopulationResponseMagnitude2019}
\begin{barticle}
\bauthor{\bsnm{Jaegle}, \binits{A.}},
\bauthor{\bsnm{Mehrpour}, \binits{V.}},
\bauthor{\bsnm{Mohsenzadeh}, \binits{Y.}},
\bauthor{\bsnm{Meyer}, \binits{T.}},
\bauthor{\bsnm{Oliva}, \binits{A.}},
\bauthor{\bsnm{Rust}, \binits{N.}}:
\batitle{Population response magnitude variation in inferotemporal cortex
  predicts image memorability}.
\bjtitle{eLife}
\bvolume{8},
\bfpage{47596}
(\byear{2019}).
\doiurl{10.7554/eLife.47596}
\end{barticle}
\endbibitem

\bibitem{xieMemorabilityWordsArbitrary2020}
\begin{barticle}
\bauthor{\bsnm{Xie}, \binits{W.}},
\bauthor{\bsnm{Bainbridge}, \binits{W.A.}},
\bauthor{\bsnm{Inati}, \binits{S.K.}},
\bauthor{\bsnm{Baker}, \binits{C.I.}},
\bauthor{\bsnm{Zaghloul}, \binits{K.A.}}:
\batitle{Memorability of words in arbitrary verbal associations modulates
  memory retrieval in the anterior temporal lobe}.
\bjtitle{Nature Human Behaviour}
\bvolume{4}(\bissue{9}),
\bfpage{937}--\blpage{948}
(\byear{2020}).
\doiurl{10.1038/s41562-020-0901-2}
\end{barticle}
\endbibitem

\bibitem{isolaWhatMakesPhotograph2014}
\begin{barticle}
\bauthor{\bsnm{Isola}, \binits{P.}},
\bauthor{\bsnm{Xiao}, \binits{J.}},
\bauthor{\bsnm{Parikh}, \binits{D.}},
\bauthor{\bsnm{Torralba}, \binits{A.}},
\bauthor{\bsnm{Oliva}, \binits{A.}}:
\batitle{What {{Makes}} a {{Photograph Memorable}}?}
\bjtitle{IEEE Transactions on Pattern Analysis and Machine Intelligence}
\bvolume{36}(\bissue{7}),
\bfpage{1469}--\blpage{1482}
(\byear{2014}).
\doiurl{10.1109/TPAMI.2013.200}
\end{barticle}
\endbibitem

\bibitem{khoslaModifyingMemorabilityFace2013}
\begin{bchapter}
\bauthor{\bsnm{Khosla}, \binits{A.}},
\bauthor{\bsnm{Bainbridge}, \binits{W.A.}},
\bauthor{\bsnm{Torralba}, \binits{A.}},
\bauthor{\bsnm{Oliva}, \binits{A.}}:
\bctitle{Modifying the {{Memorability}} of {{Face Photographs}}}.
In: \bbtitle{2013 {{IEEE International Conference}} on {{Computer Vision}}},
pp. \bfpage{3200}--\blpage{3207}.
\bpublisher{{IEEE}},
\blocation{{Sydney, Australia}}
(\byear{2013}).
\doiurl{10.1109/ICCV.2013.397}
\end{bchapter}
\endbibitem

\bibitem{khoslaUnderstandingPredictingImage2015}
\begin{bchapter}
\bauthor{\bsnm{Khosla}, \binits{A.}},
\bauthor{\bsnm{Raju}, \binits{A.S.}},
\bauthor{\bsnm{Torralba}, \binits{A.}},
\bauthor{\bsnm{Oliva}, \binits{A.}}:
\bctitle{Understanding and {{Predicting Image Memorability}} at a {{Large
  Scale}}}.
In: \bbtitle{2015 {{IEEE International Conference}} on {{Computer Vision}}
  ({{ICCV}})},
pp. \bfpage{2390}--\blpage{2398}.
\bpublisher{{IEEE}},
\blocation{{Santiago, Chile}}
(\byear{2015}).
\doiurl{10.1109/ICCV.2015.275}
\end{bchapter}
\endbibitem

\bibitem{fajtlAMNetMemorabilityEstimation2018}
\begin{bchapter}
\bauthor{\bsnm{Fajtl}, \binits{J.}},
\bauthor{\bsnm{Argyriou}, \binits{V.}},
\bauthor{\bsnm{Monekosso}, \binits{D.}},
\bauthor{\bsnm{Remagnino}, \binits{P.}}:
\bctitle{{{AMNet}}: {{Memorability Estimation}} with {{Attention}}}.
In: \bbtitle{2018 {{IEEE}}/{{CVF Conference}} on {{Computer Vision}} and
  {{Pattern Recognition}}},
pp. \bfpage{6363}--\blpage{6372}.
\bpublisher{{IEEE}},
\blocation{{Salt Lake City, UT}}
(\byear{2018}).
\doiurl{10.1109/CVPR.2018.00666}
\end{bchapter}
\endbibitem

\bibitem{squalli-houssainiDeepLearningPredicting2018}
\begin{bchapter}
\bauthor{\bsnm{{Squalli-Houssaini}}, \binits{H.}},
\bauthor{\bsnm{Duong}, \binits{N.Q.K.}},
\bauthor{\bsnm{Gwenaelle}, \binits{M.}},
\bauthor{\bsnm{Demarty}, \binits{C.-H.}}:
\bctitle{Deep {{Learning}} for {{Predicting Image Memorability}}}.
In: \bbtitle{2018 {{IEEE International Conference}} on {{Acoustics}},
  {{Speech}} and {{Signal Processing}} ({{ICASSP}})},
pp. \bfpage{2371}--\blpage{2375}.
\bpublisher{{IEEE}},
\blocation{{Calgary, AB}}
(\byear{2018}).
\doiurl{10.1109/ICASSP.2018.8462292}
\end{bchapter}
\endbibitem

\bibitem{basavarajuObjectMemorabilityPrediction2019}
\begin{barticle}
\bauthor{\bsnm{Basavaraju}, \binits{S.}},
\bauthor{\bsnm{Gaj}, \binits{S.}},
\bauthor{\bsnm{Sur}, \binits{A.}}:
\batitle{Object {{Memorability Prediction}} using {{Deep Learning}}:
  {{Location}} and {{Size Bias}}}.
\bjtitle{Journal of Visual Communication and Image Representation}
\bvolume{59},
\bfpage{117}--\blpage{127}
(\byear{2019}).
\doiurl{10.1016/j.jvcir.2019.01.008}
\end{barticle}
\endbibitem

\bibitem{huiskesMIRFlickrRetrieval2008}
\begin{bchapter}
\bauthor{\bsnm{Huiskes}, \binits{M.J.}},
\bauthor{\bsnm{Lew}, \binits{M.S.}}:
\bctitle{The {{MIR}} flickr retrieval evaluation}.
In: \bbtitle{Proceedings of the 1st {{ACM}} International Conference on
  Multimedia Information Retrieval}.
\bsertitle{{{MIR}} '08},
pp. \bfpage{39}--\blpage{43}.
\bpublisher{{Association for Computing Machinery}},
\blocation{{New York, NY, USA}}
(\byear{2008}).
\doiurl{10.1145/1460096.1460104}
\end{bchapter}
\endbibitem

\bibitem{murrayAVALargescaleDatabase2012}
\begin{bchapter}
\bauthor{\bsnm{Murray}, \binits{N.}},
\bauthor{\bsnm{Marchesotti}, \binits{L.}},
\bauthor{\bsnm{Perronnin}, \binits{F.}}:
\bctitle{{{AVA}}: {{A}} large-scale database for aesthetic visual analysis}.
In: \bbtitle{2012 {{IEEE Conference}} on {{Computer Vision}} and {{Pattern
  Recognition}}},
pp. \bfpage{2408}--\blpage{2415}.
\bpublisher{{IEEE}},
\blocation{{Providence, RI}}
(\byear{2012}).
\doiurl{10.1109/CVPR.2012.6247954}
\end{bchapter}
\endbibitem

\bibitem{machajdikAffectiveImageClassification2010}
\begin{bchapter}
\bauthor{\bsnm{Machajdik}, \binits{J.}},
\bauthor{\bsnm{Hanbury}, \binits{A.}}:
\bctitle{Affective image classification using features inspired by psychology
  and art theory}.
In: \bbtitle{Proceedings of the 18th {{ACM}} International Conference on
  Multimedia}.
\bsertitle{{{MM}} '10},
pp. \bfpage{83}--\blpage{92}.
\bpublisher{{Association for Computing Machinery}},
\blocation{{New York, NY, USA}}
(\byear{2010}).
\doiurl{10.1145/1873951.1873965}
\end{bchapter}
\endbibitem

\bibitem{juddLearningPredictWhere2009}
\begin{bchapter}
\bauthor{\bsnm{Judd}, \binits{T.}},
\bauthor{\bsnm{Ehinger}, \binits{K.}},
\bauthor{\bsnm{Durand}, \binits{F.}},
\bauthor{\bsnm{Torralba}, \binits{A.}}:
\bctitle{Learning to predict where humans look}.
In: \bbtitle{{{IEEE}} International Conference on Computer Vision ({{ICCV}})}
(\byear{2009})
\end{bchapter}
\endbibitem

\bibitem{ramanathanEyeFixationDatabase2010}
\begin{bchapter}
\bauthor{\bsnm{Ramanathan}, \binits{S.}},
\bauthor{\bsnm{Hutchison}, \binits{D.}},
\bauthor{\bsnm{Kanade}, \binits{T.}},
\bauthor{\bsnm{Kittler}, \binits{J.}},
\bauthor{\bsnm{Kleinberg}, \binits{J.M.}},
\bauthor{\bsnm{Mattern}, \binits{F.}},
\bauthor{\bsnm{Mitchell}, \binits{J.C.}},
\bauthor{\bsnm{Naor}, \binits{M.}},
\bauthor{\bsnm{Nierstrasz}, \binits{O.}},
\bauthor{\bsnm{Pandu~Rangan}, \binits{C.}},
\bauthor{\bsnm{Steffen}, \binits{B.}},
\bauthor{\bsnm{Sudan}, \binits{M.}},
\bauthor{\bsnm{Terzopoulos}, \binits{D.}},
\bauthor{\bsnm{Tygar}, \binits{D.}},
\bauthor{\bsnm{Vardi}, \binits{M.Y.}},
\bauthor{\bsnm{Weikum}, \binits{G.}},
\bauthor{\bsnm{Katti}, \binits{H.}},
\bauthor{\bsnm{Sebe}, \binits{N.}},
\bauthor{\bsnm{Kankanhalli}, \binits{M.}},
\bauthor{\bsnm{Chua}, \binits{T.-S.}}:
\bctitle{An {{Eye Fixation Database}} for {{Saliency Detection}} in
  {{Images}}}.
In: \beditor{\bsnm{Daniilidis}, \binits{K.}},
\beditor{\bsnm{Maragos}, \binits{P.}},
\beditor{\bsnm{Paragios}, \binits{N.}} (eds.)
\bbtitle{Computer {{Vision}} \textendash{} {{ECCV}} 2010}
vol. \bseriesno{6314},
pp. \bfpage{30}--\blpage{43}.
\bpublisher{{Springer Berlin Heidelberg}},
\blocation{{Berlin, Heidelberg}}
(\byear{2010}).
\doiurl{10.1007/978-3-642-15561-1_3}
\end{bchapter}
\endbibitem

\bibitem{xiaoSUNDatabaseLargescale2010}
\begin{bchapter}
\bauthor{\bsnm{Xiao}, \binits{J.}},
\bauthor{\bsnm{Hays}, \binits{J.}},
\bauthor{\bsnm{Ehinger}, \binits{K.A.}},
\bauthor{\bsnm{Oliva}, \binits{A.}},
\bauthor{\bsnm{Torralba}, \binits{A.}}:
\bctitle{{{SUN}} database: {{Large}}-scale scene recognition from abbey to
  zoo}.
In: \bbtitle{2010 {{IEEE Computer Society Conference}} on {{Computer Vision}}
  and {{Pattern Recognition}}},
pp. \bfpage{3485}--\blpage{3492}.
\bpublisher{{IEEE}},
\blocation{{San Francisco, CA, USA}}
(\byear{2010}).
\doiurl{10.1109/CVPR.2010.5539970}
\end{bchapter}
\endbibitem

\bibitem{khoslaWhatMakesImage2014}
\begin{bchapter}
\bauthor{\bsnm{Khosla}, \binits{A.}},
\bauthor{\bsnm{Das~Sarma}, \binits{A.}},
\bauthor{\bsnm{Hamid}, \binits{R.}}:
\bctitle{What makes an image popular?}
In: \bbtitle{Proceedings of the 23rd International Conference on {{World}} Wide
  Web - {{WWW}} '14},
pp. \bfpage{867}--\blpage{876}.
\bpublisher{{ACM Press}},
\blocation{{Seoul, Korea}}
(\byear{2014}).
\doiurl{10.1145/2566486.2567996}
\end{bchapter}
\endbibitem

\bibitem{salehObjectCentricAnomalyDetection2013}
\begin{bchapter}
\bauthor{\bsnm{Saleh}, \binits{B.}},
\bauthor{\bsnm{Farhadi}, \binits{A.}},
\bauthor{\bsnm{Elgammal}, \binits{A.}}:
\bctitle{Object-{{Centric Anomaly Detection}} by {{Attribute}}-{{Based
  Reasoning}}}.
In: \bbtitle{2013 {{IEEE Conference}} on {{Computer Vision}} and {{Pattern
  Recognition}}},
pp. \bfpage{787}--\blpage{794}.
\bpublisher{{IEEE}},
\blocation{{Portland, OR, USA}}
(\byear{2013}).
\doiurl{10.1109/CVPR.2013.107}
\end{bchapter}
\endbibitem

\bibitem{farhadiDescribingObjectsTheir2009}
\begin{bchapter}
\bauthor{\bsnm{Farhadi}, \binits{A.}},
\bauthor{\bsnm{Endres}, \binits{I.}},
\bauthor{\bsnm{Hoiem}, \binits{D.}},
\bauthor{\bsnm{Forsyth}, \binits{D.}}:
\bctitle{Describing objects by their attributes}.
In: \bbtitle{2009 {{IEEE Conference}} on {{Computer Vision}} and {{Pattern
  Recognition}}},
pp. \bfpage{1778}--\blpage{1785}.
\bpublisher{{IEEE}},
\blocation{{Miami, FL}}
(\byear{2009}).
\doiurl{10.1109/CVPR.2009.5206772}
\end{bchapter}
\endbibitem

\bibitem{goetschalckxMemCatNewCategorybased2019}
\begin{barticle}
\bauthor{\bsnm{Goetschalckx}, \binits{L.}},
\bauthor{\bsnm{Wagemans}, \binits{J.}}:
\batitle{{{MemCat}}: A new category-based image set quantified on
  memorability}.
\bjtitle{PeerJ}
\bvolume{7},
\bfpage{8169}
(\byear{2019}).
\doiurl{10.7717/peerj.8169}
\end{barticle}
\endbibitem

\bibitem{teamTorchTensorsDynamic}
\begin{botherref}
\oauthor{\bsnm{Team}, \binits{P.}}:
Torch: {{Tensors}} and {{Dynamic}} Neural Networks in {{Python}} with Strong
  {{GPU}} Acceleration
\end{botherref}
\endbibitem

\bibitem{fukushimaNeocognitronSelforganizingNeural1980}
\begin{barticle}
\bauthor{\bsnm{Fukushima}, \binits{K.}}:
\batitle{Neocognitron: {{A}} self-organizing neural network model for a
  mechanism of pattern recognition unaffected by shift in position}.
\bjtitle{Biological Cybernetics}
\bvolume{36}(\bissue{4}),
\bfpage{193}--\blpage{202}
(\byear{1980}).
\doiurl{10.1007/BF00344251}
\end{barticle}
\endbibitem

\bibitem{chellapillaHighPerformanceConvolutional}
\begin{bchapter}
\bauthor{\bsnm{Chellapilla}, \binits{K.}},
\bauthor{\bsnm{Puri}, \binits{S.}},
\bauthor{\bsnm{Simard}, \binits{P.}}:
\bctitle{{High Performance Convolutional Neural Networks for Document
  Processing}}.
In: \beditor{\bsnm{Lorette}, \binits{G.}} (ed.)
\bbtitle{{Tenth International Workshop on Frontiers in Handwriting
  Recognition}}.
\bpublisher{{Suvisoft}},
\blocation{La Baule (France)}
(\byear{2006}).
\bcomment{{Universit{\'e} de Rennes 1}. http://www.suvisoft.com}.
\burl{https://hal.inria.fr/inria-00112631}
\end{bchapter}
\endbibitem

\bibitem{ciresanDeepBigSimple2010}
\begin{barticle}
\bauthor{\bsnm{Cire{\c s}an}, \binits{D.C.}},
\bauthor{\bsnm{Meier}, \binits{U.}},
\bauthor{\bsnm{Gambardella}, \binits{L.M.}},
\bauthor{\bsnm{Schmidhuber}, \binits{J.}}:
\batitle{Deep, {{Big}}, {{Simple Neural Nets}} for {{Handwritten Digit
  Recognition}}}.
\bjtitle{Neural Computation}
\bvolume{22}(\bissue{12}),
\bfpage{3207}--\blpage{3220}
(\byear{2010})
\end{barticle}
\endbibitem

\bibitem{krizhevskyImageNetClassificationDeep2012}
\begin{barticle}
\bauthor{\bsnm{Krizhevsky}, \binits{A.}},
\bauthor{\bsnm{Sutskever}, \binits{I.}},
\bauthor{\bsnm{Hinton}, \binits{G.E.}}:
\batitle{{{ImageNet}} classification with deep convolutional neural networks}.
\bjtitle{Communications of the ACM}
\bvolume{60}(\bissue{6}),
\bfpage{84}--\blpage{90}
(\byear{2012}).
\doiurl{10.1145/3065386}
\end{barticle}
\endbibitem

\bibitem{heDeepResidualLearning2015}
\begin{botherref}
\oauthor{\bsnm{He}, \binits{K.}},
\oauthor{\bsnm{Zhang}, \binits{X.}},
\oauthor{\bsnm{Ren}, \binits{S.}},
\oauthor{\bsnm{Sun}, \binits{J.}}:
Deep {{Residual Learning}} for {{Image Recognition}}.
arXiv:1512.03385 [cs]
(2015)
{\href{https://arxiv.org/abs/1512.03385}{{arXiv:1512.03385}}}
{[cs]}
\end{botherref}
\endbibitem

\bibitem{deng2009imagenet}
\begin{bchapter}
\bauthor{\bsnm{Deng}, \binits{J.}},
\bauthor{\bsnm{Dong}, \binits{W.}},
\bauthor{\bsnm{Socher}, \binits{R.}},
\bauthor{\bsnm{Li}, \binits{L.-J.}},
\bauthor{\bsnm{Li}, \binits{K.}},
\bauthor{\bsnm{{Fei-Fei}}, \binits{L.}}:
\bctitle{Imagenet: {{A}} large-scale hierarchical image database}.
In: \bbtitle{2009 {{IEEE}} Conference on Computer Vision and Pattern
  Recognition},
pp. \bfpage{248}--\blpage{255}
(\byear{2009}).
\bcomment{{Ieee}}
\end{bchapter}
\endbibitem

\bibitem{teamTorchvisionImageVideo}
\begin{botherref}
\oauthor{\bsnm{Team}, \binits{P.C.}}:
Torchvision: Image and Video Datasets and Models for Torch Deep Learning
\end{botherref}
\endbibitem

\bibitem{jozwikDeepConvolutionalNeural2018}
\begin{bchapter}
\bauthor{\bsnm{Jozwik}, \binits{K.M.}},
\bauthor{\bsnm{Kriegeskorte}, \binits{N.}},
\bauthor{\bsnm{Cichy}, \binits{R.M.}},
\bauthor{\bsnm{Mur}, \binits{M.}}:
\bctitle{Deep convolutional neural networks, features, and categories perform
  similarly at explaining primate high-level visual representations}.
In: \bbtitle{2018 {{Conference}} on {{Cognitive Computational Neuroscience}}}.
\bpublisher{{Cognitive Computational Neuroscience}},
\blocation{{Philadelphia, Pennsylvania, USA}}
(\byear{2018}).
\doiurl{10.32470/CCN.2018.1232-0}
\end{bchapter}
\endbibitem

\bibitem{olahFeatureVisualization2017}
\begin{barticle}
\bauthor{\bsnm{Olah}, \binits{C.}},
\bauthor{\bsnm{Mordvintsev}, \binits{A.}},
\bauthor{\bsnm{Schubert}, \binits{L.}}:
\batitle{Feature {{Visualization}}}.
\bjtitle{Distill}
\bvolume{2}(\bissue{11}),
\bfpage{10}--\blpage{2391500007}
(\byear{2017}).
\doiurl{10.23915/distill.00007}
\end{barticle}
\endbibitem

\bibitem{harris2020array}
\begin{barticle}
\bauthor{\bsnm{Harris}, \binits{C.R.}},
\bauthor{\bsnm{Millman}, \binits{K.J.}},
\bauthor{\bsnm{{van der Walt}}, \binits{S.J.}},
\bauthor{\bsnm{Gommers}, \binits{R.}},
\bauthor{\bsnm{Virtanen}, \binits{P.}},
\bauthor{\bsnm{Cournapeau}, \binits{D.}},
\bauthor{\bsnm{Wieser}, \binits{E.}},
\bauthor{\bsnm{Taylor}, \binits{J.}},
\bauthor{\bsnm{Berg}, \binits{S.}},
\bauthor{\bsnm{Smith}, \binits{N.J.}},
\bauthor{\bsnm{Kern}, \binits{R.}},
\bauthor{\bsnm{Picus}, \binits{M.}},
\bauthor{\bsnm{Hoyer}, \binits{S.}},
\bauthor{\bsnm{{van Kerkwijk}}, \binits{M.H.}},
\bauthor{\bsnm{Brett}, \binits{M.}},
\bauthor{\bsnm{Haldane}, \binits{A.}},
\bauthor{\bsnm{{del R{\'i}o}}, \binits{J.F.}},
\bauthor{\bsnm{Wiebe}, \binits{M.}},
\bauthor{\bsnm{Peterson}, \binits{P.}},
\bauthor{\bsnm{{G{\'e}rard-Marchant}}, \binits{P.}},
\bauthor{\bsnm{Sheppard}, \binits{K.}},
\bauthor{\bsnm{Reddy}, \binits{T.}},
\bauthor{\bsnm{Weckesser}, \binits{W.}},
\bauthor{\bsnm{Abbasi}, \binits{H.}},
\bauthor{\bsnm{Gohlke}, \binits{C.}},
\bauthor{\bsnm{Oliphant}, \binits{T.E.}}:
\batitle{Array programming with {{NumPy}}}.
\bjtitle{Nature}
\bvolume{585}(\bissue{7825}),
\bfpage{357}--\blpage{362}
(\byear{2020}).
\doiurl{10.1038/s41586-020-2649-2}
\end{barticle}
\endbibitem

\bibitem{bainbridgeMemorabilityPhotographsSubjective2019}
\begin{barticle}
\bauthor{\bsnm{Bainbridge}, \binits{W.A.}},
\bauthor{\bsnm{Berron}, \binits{D.}},
\bauthor{\bsnm{Sch{\"u}tze}, \binits{H.}},
\bauthor{\bsnm{Cardenas-Blanco}, \binits{A.}},
\bauthor{\bsnm{Metzger}, \binits{C.}},
\bauthor{\bsnm{Dobisch}, \binits{L.}},
\bauthor{\bsnm{Bittner}, \binits{D.}},
\bauthor{\bsnm{Glanz}, \binits{W.}},
\bauthor{\bsnm{Spottke}, \binits{A.}},
\bauthor{\bsnm{Rudolph}, \binits{J.}},
\bauthor{\bsnm{Brosseron}, \binits{F.}},
\bauthor{\bsnm{Buerger}, \binits{K.}},
\bauthor{\bsnm{Janowitz}, \binits{D.}},
\bauthor{\bsnm{Fliessbach}, \binits{K.}},
\bauthor{\bsnm{Heneka}, \binits{M.}},
\bauthor{\bsnm{Laske}, \binits{C.}},
\bauthor{\bsnm{Buchmann}, \binits{M.}},
\bauthor{\bsnm{Peters}, \binits{O.}},
\bauthor{\bsnm{Diesing}, \binits{D.}},
\bauthor{\bsnm{Li}, \binits{S.}},
\bauthor{\bsnm{Priller}, \binits{J.}},
\bauthor{\bsnm{Spruth}, \binits{E.J.}},
\bauthor{\bsnm{Altenstein}, \binits{S.}},
\bauthor{\bsnm{Schneider}, \binits{A.}},
\bauthor{\bsnm{Kofler}, \binits{B.}},
\bauthor{\bsnm{Teipel}, \binits{S.}},
\bauthor{\bsnm{Kilimann}, \binits{I.}},
\bauthor{\bsnm{Wiltfang}, \binits{J.}},
\bauthor{\bsnm{Bartels}, \binits{C.}},
\bauthor{\bsnm{Wolfsgruber}, \binits{S.}},
\bauthor{\bsnm{Wagner}, \binits{M.}},
\bauthor{\bsnm{Jessen}, \binits{F.}},
\bauthor{\bsnm{Baker}, \binits{C.I.}},
\bauthor{\bsnm{D{\"u}zel}, \binits{E.}}:
\batitle{Memorability of photographs in subjective cognitive decline and mild
  cognitive impairment: {{Implications}} for cognitive assessment}.
\bjtitle{Alzheimer's \& Dementia: Diagnosis, Assessment \& Disease Monitoring}
\bvolume{11}(\bissue{1}),
\bfpage{610}--\blpage{618}
(\byear{2019}).
\doiurl{10.1016/j.dadm.2019.07.005}
\end{barticle}
\endbibitem

\bibitem{lloydEatingDisorder}
\begin{barticle}
\bauthor{\bsnm{Lloyd}, \binits{E.C.}},
\bauthor{\bsnm{Shehzad}, \binits{Z.}},
\bauthor{\bsnm{Schebendach}, \binits{J.}},
\bauthor{\bsnm{Bakkour}, \binits{A.}},
\bauthor{\bsnm{Xue}, \binits{A.M.}},
\bauthor{\bsnm{Assaf}, \binits{N.F.}},
\bauthor{\bsnm{Jilani}, \binits{R.}},
\bauthor{\bsnm{Walsh}, \binits{B.T.}},
\bauthor{\bsnm{Steinglass}, \binits{J.}},
\bauthor{\bsnm{Foerde}, \binits{K.}}:
\batitle{Food folio by columbia center for eating disorders: A freely available
  food image database}.
\bjtitle{Frontiers in Psychology}
\bvolume{11},
\bfpage{3556}
(\byear{2020}).
\doiurl{10.3389/fpsyg.2020.585044}
\end{barticle}
\endbibitem

\bibitem{XinyueFoods}
\begin{botherref}
\oauthor{\bsnm{Li}, \binits{X.}}:
The effect of memorability on food choice: Do people prefer more memorable
  foods?
(THESIS).
\doiurl{10.6082/uchicago.2912}
\end{botherref}
\endbibitem

\bibitem{dubeyWhatMakesObject}
\begin{botherref}
\oauthor{\bsnm{Dubey}, \binits{R.}},
\oauthor{\bsnm{Peterson}, \binits{J.}},
\oauthor{\bsnm{Khosla}, \binits{A.}},
\oauthor{\bsnm{Yang}, \binits{M.-H.}},
\oauthor{\bsnm{Ghanem}, \binits{B.}}:
What {{Makes}} an {{Object Memorable}}?,
9
\end{botherref}
\endbibitem

\bibitem{Bainbridge_2017}
\begin{barticle}
\bauthor{\bsnm{Bainbridge}, \binits{W.A.}}:
\batitle{The memorability of people: Intrinsic memorability across
  transformations of a person’s face.}
\bjtitle{Journal of Experimental Psychology: Learning, Memory, and Cognition}
\bvolume{43}(\bissue{5}),
\bfpage{706}--\blpage{716}
(\byear{2017}).
\doiurl{10.1037/xlm0000339}
\end{barticle}
\endbibitem

\bibitem{kramerCharacterizingMemorabilityRepresentational2021}
\begin{bchapter}
\bauthor{\bsnm{Kramer}, \binits{M.}},
\bauthor{\bsnm{Hebart}, \binits{M.H.}},
\bauthor{\bsnm{Baker}, \binits{C.I.}},
\bauthor{\bsnm{Bainbridge}, \binits{W.A.}}:
\bctitle{Characterizing memorability in representational space: {{Analyzing
  Relative Contributions}} of {{Perceptual}} and {{Conceptual Information}}}.
In: \bbtitle{Vision {{Sciences Society}}}
(\byear{2021})
\end{bchapter}
\endbibitem

\bibitem{cichyComparisonDeepNeural2016}
\begin{barticle}
\bauthor{\bsnm{Cichy}, \binits{R.M.}},
\bauthor{\bsnm{Khosla}, \binits{A.}},
\bauthor{\bsnm{Pantazis}, \binits{D.}},
\bauthor{\bsnm{Torralba}, \binits{A.}},
\bauthor{\bsnm{Oliva}, \binits{A.}}:
\batitle{Comparison of deep neural networks to spatio-temporal cortical
  dynamics of human visual object recognition reveals hierarchical
  correspondence}.
\bjtitle{Scientific Reports}
\bvolume{6}(\bissue{1}),
\bfpage{27755}
(\byear{2016}).
\doiurl{10.1038/srep27755}
\end{barticle}
\endbibitem

\bibitem{Yamins_Hong_Cadieu_Solomon_Seibert_DiCarlo_2014}
\begin{barticle}
\bauthor{\bsnm{Yamins}, \binits{D.L.K.}},
\bauthor{\bsnm{Hong}, \binits{H.}},
\bauthor{\bsnm{Cadieu}, \binits{C.F.}},
\bauthor{\bsnm{Solomon}, \binits{E.A.}},
\bauthor{\bsnm{Seibert}, \binits{D.}},
\bauthor{\bsnm{DiCarlo}, \binits{J.J.}}:
\batitle{Performance-optimized hierarchical models predict neural responses in
  higher visual cortex}.
\bjtitle{Proceedings of the National Academy of Sciences}
\bvolume{111}(\bissue{23}),
\bfpage{8619}--\blpage{8624}
(\byear{2014}).
\doiurl{10.1073/pnas.1403112111}
\end{barticle}
\endbibitem

\bibitem{Koch834796}
\begin{barticle}
\bauthor{\bsnm{Koch}, \binits{G.E.}},
\bauthor{\bsnm{Akpan}, \binits{E.}},
\bauthor{\bsnm{Coutanche}, \binits{M.N.}}:
\batitle{Image memorability is predicted at different stages of a convolutional
  neural network}.
\bjtitle{bioRxiv}
(\byear{2020})
{\href{https://arxiv.org/abs/https://www.biorxiv.org/content/early/2020/03/14/834796.full.pdf}{{https://www.biorxiv.org/content/early/2020/03/14/834796.full.pdf}}}.
\doiurl{10.1101/834796}
\end{barticle}
\endbibitem

\end{thebibliography}

\end{document}


\maketitle
\section{Package Tutorial}

While the \texttt{resmem} package may be used in any way that the user wishes, this section will outline the recommended steps for basic usage.

First, it is recommended to set up a Python\citep{vanrossumPythonReferenceManual2009} virtual environment. Then \texttt{resmem} can be installed using pip, the built-in Python package manager. Also install Pillow\citep{clarkPillowPythonImaging} to handle image importing.

\begin{lstlisting}[language=bash]
python -m venv venv
source venv/bin/activate
pip install resmem pillow
\end{lstlisting}

Then in a Python file, Jupyter notebook, or Python REPL:

\begin{lstlisting}[language=Python]
from resmem import ResMem, transformer
from PIL import Image

model = ResMem(pretrained=True)
\end{lstlisting}

To get an estimation for an image stored at \texttt{./path/to/image.jpg}.

\begin{lstlisting}[language=Python]
img = Image.open('./path/to/image.jpg')  
# Load the image into the computer's memory.
img = img.convert('RGB')  
#  The model only accepts images in RGB format.

model.eval()  
# Sets resmem into Evaluation/Inference mode so that the parameters do not change.
# It also turns off other behavior that is useful for training but leads to lower quality estimations.

image_x = transformer(img)  # Runs the preprocessing function on the image.

prediction = model(image_x.view(-1, 3, 227, 227))  
# ResMem expects a batch of images for optimization reasons, so if we give it only one image, it needs to be arranged as a batch of one. 
# This is what the .view() method does.
\end{lstlisting}

For large datasets, it is recommended to create a \texttt{DataSet} object using \texttt{torch} data utilities\citep{teamTorchTensorsDynamic}. Depending on how the dataset is organized, the following code may need to be adjusted. It is written with the assumption that there is a text file named \texttt{data.csv} which has image file names as the first column and their experimentally determined memorability scores as the second column, but the user may structure this however they want.
For managing data, the package \texttt{numpy} is often used. Both Numpy and Torch are included in the resmem installation. In the following code it is aliased as \texttt{np}.

\begin{lstlisting}[language=python]
import numpy as np
from torch.utils.data import Dataset

class ResMemDataset(Dataset):
    def __init__(self, loc='./Sources/dataset/', transform=transformer):  # In this case all of the images are stored in the folder Sources/dataset/images.
        self.frame = np.array(np.loadtxt(f'{loc}data.txt', delimiter=','))
        self.loc = loc
        self.transform = transform
        
    def __len__(self):
        return self.frame.shape[0] 
        # This tells the system how many datapoints are in the image set.
    
    def __getitem__(self, idx):
        # This function tells python how to get a 
        # particular image out of the object.
        if torch.istensor(idx):
            idx = idx.tolist()  
            # ensures compatibility with all torch operations.
        
        img_name = self.frame[idx, 0]
        image = Image.open(f'{self.loc}/images/{img_name}')
        image = image.convert('RGB')
        y = self.frame[idx, 1]
        y = torch.Tensor([float(y)])
        image_x = self.transform(image)
        return image_x, y, img_name
    
\end{lstlisting}

Then, it is recommended to use the \texttt{DataLoader} from \texttt{torch} to automatically create batches.

\begin{lstlisting}[language=Python]
from torch.utils.data import DataLoader

image_set = ResMemDataset()
data = DataLoader(image_set, batch_size=8, num_workers=4, pin_memory=True)
# A batch size of 8 will be appropriate for most GPUs. 
# Pinning the memory greatly increases performance on large datasets. (n >> batch_size)

model = ResMem(pretrained=True).cuda()  
# The .cuda() method will transfer the model onto a GPU. 
# If you do not have an NVIDIA GPU, omit this method.

model.eval()

rloss = 0
y_hats = []
ys = []
names = []
t = 1

for batch in data:
    x, y, name = batch
    ys += y.squeeze().tolist()  
    # Store the ground truths for later. 
    # If your dataset does not include ground truths, omit any references to y.
    bs, c, h, w = x.size()  
    # Get all of the size and shape information about the data.
    y_hat = model(x.cuda().view(-1, c, h, w)).view(bs, -1).mean(1)
    y_hats += y_hat.squeeze().tolist()
    names += name

rcorr = spearmanr(ys, preds)[0]
loss = ((np.array(ys) - np.array(preds)) ** 2).mean()
\end{lstlisting}

\section{Model Training}

\begin{figure}[p]
\includegraphics[width=\textwidth]{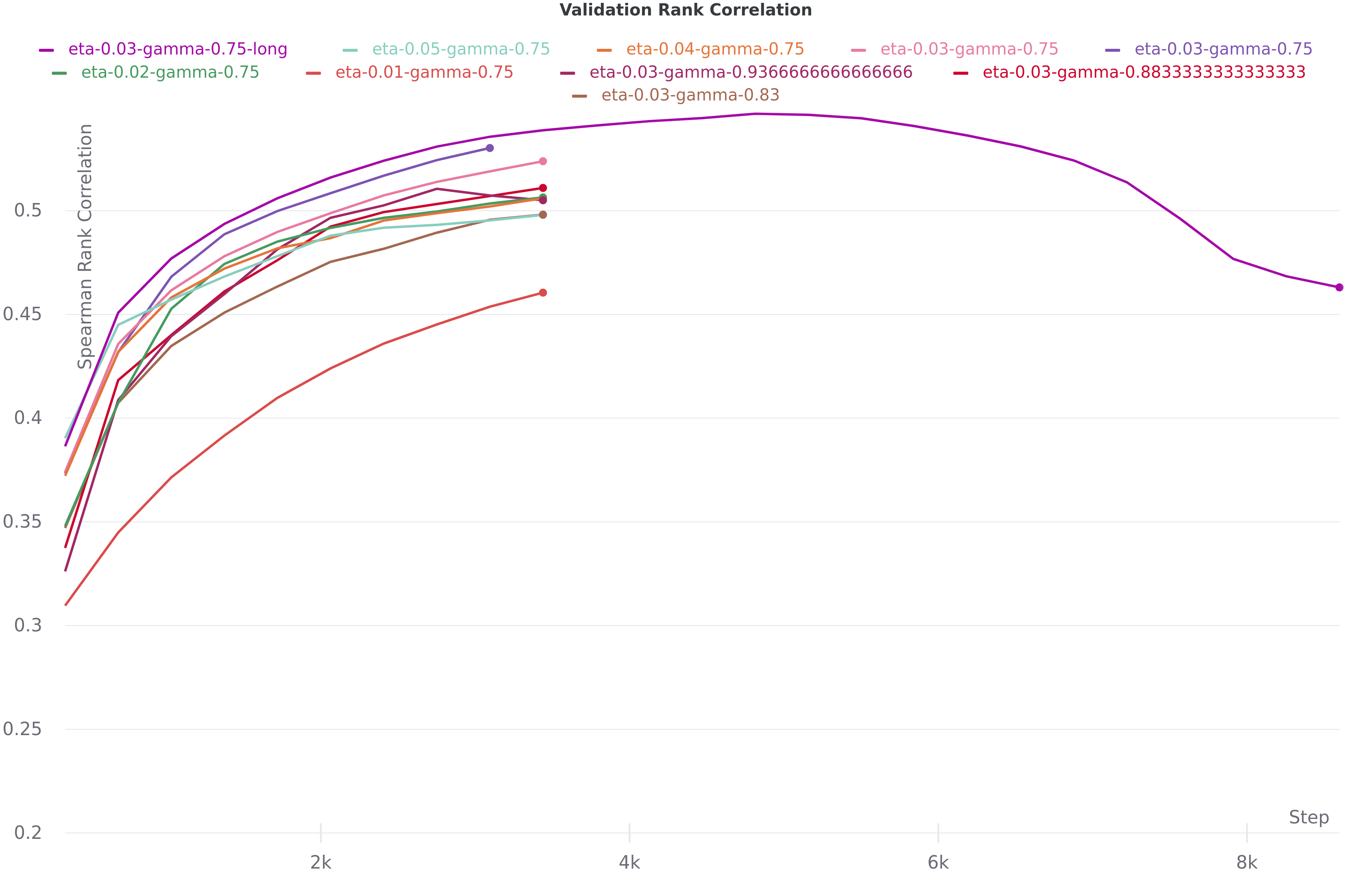}
\caption{The MemNet tuning process is shown in (a). In this figure, hyperparameters were generated by hand to explore the 
behavior of the model in training. One run is longer than the others to examine how the model
behaves in the overfitting regime. Eta($\eta$) represents the learning rate, and gamma($\gamma$) represents
the momentum of the optimization method.}
\label{fig:mnsweep}
\end{figure}

We conducted our  hyperparameter tuning using
\href{https://www.wandb.ai}{https://www.wandb.ai} \citep{biasesWandbCLILibrary} (fig.~\ref{fig:mnsweep}). Each of the lines in (fig.~\ref{fig:mnsweep}) represents a separate
training session of our reconstructed MemNet, recording the Spearman
rank correlation of the model's predictions on a validation dataset
after each epoch. Each attempt has different hyperparameters, or the
training/testing environment was changed in some minor way. Hundreds of training and testing 
attempts were run, and a selection are shown in fig.~\ref{fig:mnsweep}

\begin{figure}[p]
\includegraphics[width=\textwidth]{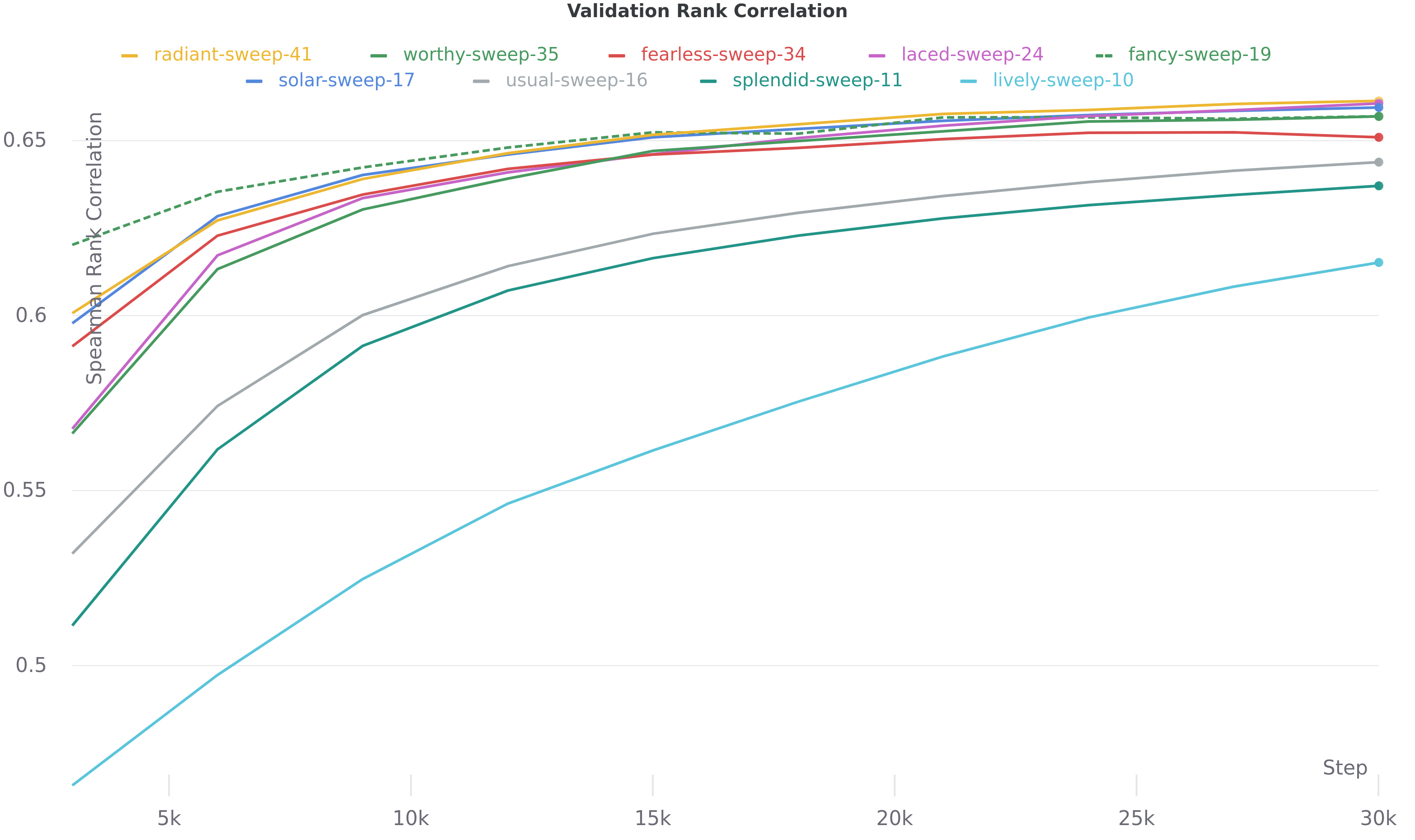}
\caption{ Validation performance from ResMem's hyperparameter tuning
phase. Each line represents a model with slightly different
hyperparameters (training rates, momenta). As the training phase goes
on, for the most part, performance improves on the validation
set. }

\label{fig:rnsweep}
\end{figure}

When examined in comparison with the validation performance of ResMem (fig.~\ref{fig:rnsweep}), it can be seen that ResMem trains much more slowly, in most cases not even reaching an overfitting regime by the 30,000th training step. 
This is a feature of ResMem being a much larger, more complex model. 

The hyperparameter tuning phase is done by training and testing hundreds of models with slightly tweaked hyperparameters. The hyperparameter set that is associated with the best performance in terms of Spearman rank correlation was then used as the hyperparameter set for the final training process. The model created by this process was used for all of the analyses in the main text, and in the case of ResMemRetrain, is published for use by future researchers.

\pagebreak
\bibliography{resmem}